\documentclass[acmtog]{acmart}
\usepackage{booktabs} 
\usepackage{multirow}
\usepackage{amsmath}
\usepackage{algorithm}
\usepackage{algpseudocode}
\usepackage{graphicx}
\makeatletter
\renewcommand{\ALG@name}{\small ALGORITHM}
\makeatother
\usepackage{caption}
\usepackage{subfigure}
\usepackage{xspace}

\setcopyright{acmcopyright}
\setcopyright{acmlicensed}
\setcopyright{rightsretained}

\algrenewcommand{\algorithmicindent}{1.1em}
\usepackage{cleveref}
\usepackage{dsfont}

\usepackage{amsmath,amsfonts,bm}

\def\eqref#1{equation~\ref{#1}}

\def\1{\bm{1}}

\def\rvs{{\mathbf{s}}}

\def\rvz{{\mathbf{z}}}

\DeclareMathAlphabet{\mathsfit}{\encodingdefault}{\sfdefault}{m}{sl}
\SetMathAlphabet{\mathsfit}{bold}{\encodingdefault}{\sfdefault}{bx}{n}

\usepackage[normalem]{ulem}
\usepackage{listings}
\usepackage{titletoc}
\usepackage{placeins}
\definecolor{codegreen}{rgb}{0,0.6,0}
\definecolor{codegray}{rgb}{0.5,0.5,0.5}
\definecolor{codepurple}{rgb}{0.58,0,0.82}
\definecolor{backcolour}{rgb}{0.95,0.95,0.92}

\copyrightyear{2026}
\acmYear{2026}
\setcopyright{cc}
\setcctype{by-nc-nd}
\acmConference[SIGGRAPH Conference Papers '26]{Special Interest Group on Computer Graphics and Interactive Techniques Conference Conference Papers}{July 19--23, 2026}{Los Angeles, CA, USA}
\acmBooktitle{Special Interest Group on Computer Graphics and Interactive Techniques Conference Conference Papers (SIGGRAPH Conference Papers '26), July 19--23, 2026, Los Angeles, CA, USA}
\acmDOI{10.1145/3799902.3811038}
\acmISBN{979-8-4007-2554-8/2026/07}

\begin{document}

\author{Yi Shi}
\affiliation{
  \institution{Simon Fraser University}
  \city{Burnaby}
  \state{British Columbia}
  \country{Canada}
}
\affiliation{
  \institution{NVIDIA}
  \city{Santa Clara}
  \state{California}
  \country{USA}
}
\email{yish@nvidia.com}

\author{Yifeng Jiang}
\affiliation{
  \institution{NVIDIA}
  \city{Santa Clara}
  \state{California}
  \country{USA}
}
\email{yifengj@nvidia.com}

\author{Chen Tessler}
\affiliation{
  \institution{NVIDIA}
  \city{Santa Clara}
  \state{California}
  \country{USA}
}
\email{ctessler@nvidia.com}

\author{Xue Bin Peng}
\affiliation{
  \institution{Simon Fraser University}
  \city{Burnaby}
  \state{British Columbia}
  \country{Canada}
}
\affiliation{
  \institution{NVIDIA}
  \city{Vancouver}
  \state{British Columbia}
  \country{Canada}
}
\email{japeng@nvidia.com}

\begin{teaserfigure}
  \centering
  \includegraphics[width=\textwidth]{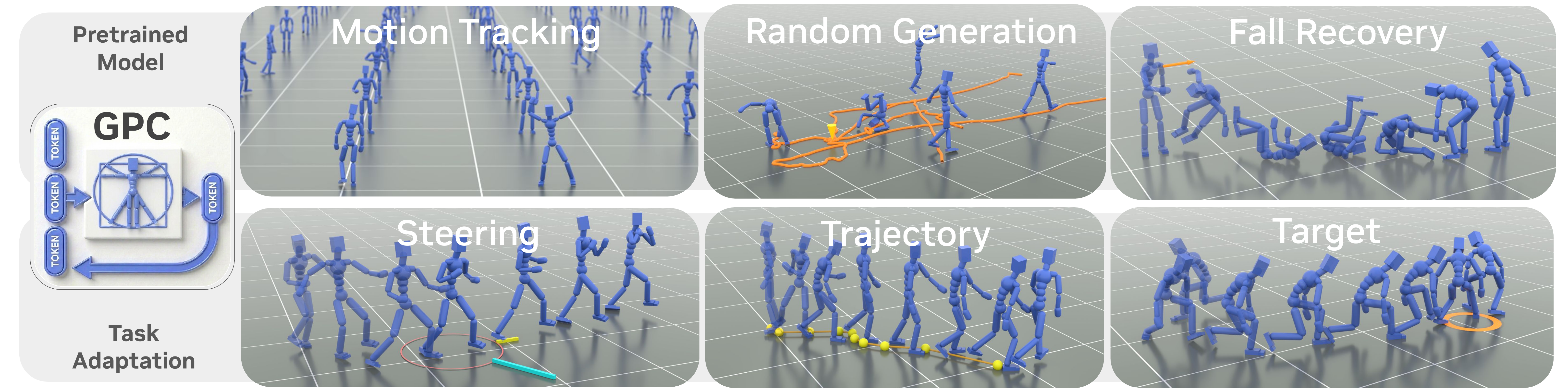}
  \label{fig:teaser}
  \caption{ We present Generative Pretrained Controllers (GPC), which models a wide range of motor skills through next-token prediction with a learned discrete representation of behaviors. GPC can be trained on large-scale motion datasets, and then adapted to downstream tasks, while retaining the naturalistic behaviors acquired by the pretrained model.}
\end{teaserfigure}

\title{GPC: Large-Scale Generative Pretraining for Transferable Motor Control}
\begin{abstract}
Developing controllers capable of completing a wide range of tasks in a natural and life-like manner is a key challenge in enabling practical applications of physics-based character animation. In this work, we introduce Generative Pretrained Controllers (GPC), which leverage tokenization and next-token modeling to create general-purpose, reusable \emph{generative} controllers from large-scale motion datasets. Our framework utilizes end-to-end reinforcement learning to jointly optimize a "motion vocabulary", modeled via Finite Scalar Quantization (FSQ), along with a corresponding control policy that can map the discrete codes to physics-based controls. After the "codebook" has been learned, the underlying structure of this large vocabulary is modeled by training a GPT-style autoregressive transformer, leading to a powerful generative controller that generates controls for a physically simulated character by performing next-token prediction. Once the generative controller has been trained, we propose a suite of adaptation techniques for finetuning the controller for new downstream tasks. Our proposed framework greatly simplifies the training process compared to previous tokenized methods, and achieves a 99.98\% success rate in reproducing a vast corpus of motion clips. The generative controller exhibits a variety of natural emergent behaviors, such as responsive behaviors to perturbations and recovery behaviors after falling. This results in highly robust general purpose controllers for a variety of downstream applications.

\end{abstract}

\maketitle

\section{Introduction}
\label{sec:intro}
Training physics-based controllers that can endow virtual characters with versatile, human-like behaviors is essential for applications such as film, video games, and extended reality (XR), as it enables characters to interact realistically with dynamic environments and generalize beyond manually authored animations. Achieving this goal requires controllers capable of capturing a broad spectrum of human motor skills and providing a convenient mechanism to reuse these skills across different tasks. Prior works have leveraged generative models, such as variational autoencoders (VAEs) and generative adversarial networks (GANs) \citep{kingma2022vae, goodfellow2014gan}, to learn latent representations of a wide range of motor skills from diverse motion datasets \citep{tessler2024maskedmimic, luo2023perpetual, peng2022ase}. Once such a generative model is trained, a high-level controller can be developed to select appropriate skills from the model's latent space to perform various downstream tasks. Constraining the high-level controller to operate within this structured latent space, instead of directly issuing low-level joint commands, provides an inductive bias toward behaviors that are more consistent with those in the dataset. However, most existing generative controllers adopt continuous latent spaces, which are prone to mode collapse and unnatural behaviors caused by drift and \emph{gaps} in the latent manifold \citep{peng2022ase, dou2022case, won2022physvae}.

To address these challenges, recent work has explored discrete latent models based on VQ-VAEs \citep{oord2018vqvae}. By mapping continuous data into discrete codes, these models sidestep the difficulty of modeling complex continuous distributions by encoding continuous features into discrete embeddings \citep{yao2024moconvq, zhu2023neural, plt2024bae, bae2025hybridlatent}. However, VQ-VAE–based tokenization can be prone to degeneracies, such as low code usage, which often requires complex training heuristics to mitigate \citep{oord2018vqvae}. These degeneracies can compromise the expressiveness of the learned representation, which can in turn increase the difficulty of capturing diverse behaviors from large motion datasets \citep{yao2024moconvq}. As a result, existing VQ-VAE–based tracking controllers have been trained primarily on datasets of modest scale, ranging from small curated datasets to subsets of larger datasets, such as AMASS \citep{mahmood2019amass}, which contain approximately 20 hours of motion data \citep{yao2024moconvq}.

In this work, we introduce generative pretrained controllers (GPC), which learn a discretized latent representation of motor skills using Finite Scalar Quantization (FSQ) \citep{mentzer2023fsq}. Unlike VQ-VAE–based methods, FSQ does not require an explicit codebook, which substantially simplifies training by eliminating the need for codebook embedding updates, auxiliary losses, and ad hoc heuristics such as dead-code re-initialization \citep{oord2018vqvae}. By training with a simple motion-tracking objective, our model is able to learn a discrete latent space that captures a wide range of behaviors, including highly dynamic skills such as vaulting, cartwheels, and flips. Once the discrete skill representation has been constructed, we train a GPT-style autoregressive transformer to model the distribution of skill tokens \citep{radford2018gpt, radford2019gpt2}, which serves as the generative controller over skills. The learned generative controller produces natural emergent behaviors, such as human-like responses to perturbations and recovery strategies. To adapt the pretrained generative controller to new tasks, we leverage parameter-efficient fine-tuning (PEFT), which inserts lightweight modulation layers into the pretrained generative controller. The PEFT layers can be updated with additional task objectives, while keeping the underlying generative model fixed. This enables GPC to leverage previously learned skills to perform new tasks while preserving the natural, life-like behaviors encoded in the pretrained model.

The core contribution of this work is a generative pretrained controller that uses a discrete latent space modeled with Finite Scalar Quantization (FSQ). The controller is trained end-to-end with reinforcement learning on a large-scale human motion dataset containing over 600 hours of diverse behaviors. The proposed framework supports parameter-efficient fine-tuning (PEFT), enabling efficient adaptation of the generative controller to new downstream tasks without retraining. The resulting FSQ-based tracking controller produces natural behaviors and achieves a 99.98\% tracking success rate on a large-scale dataset. Once trained, GPC can be effectively adapted and reused
across a diverse set of downstream tasks through parameter-efficient fine-tuning.

\begin{figure*}[t]
    \centering
    \begin{minipage}{0.48\textwidth}
        \centering
        \includegraphics[width=\linewidth]{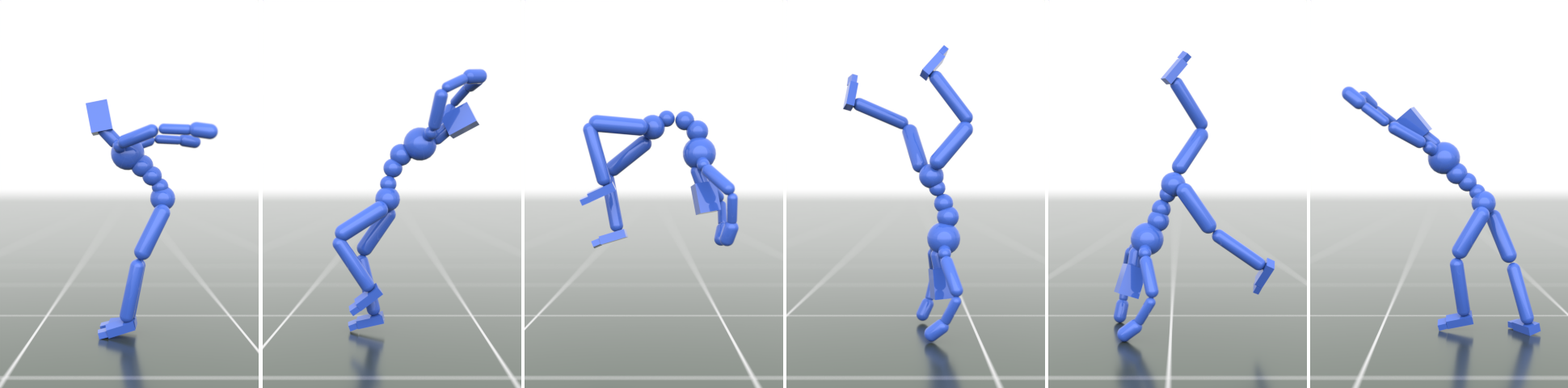}
        \small (a) Acrobat. 
    \end{minipage}\hfill
    \begin{minipage}{0.48\textwidth}
        \centering
        \includegraphics[width=\linewidth]{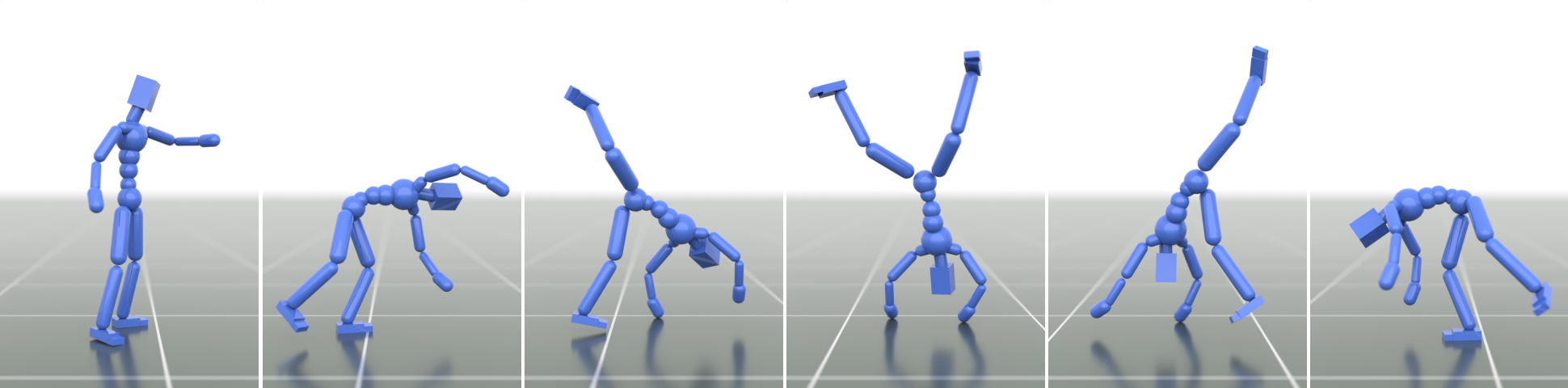}
        \small (b) Cartwheel.
    \end{minipage}

    \begin{minipage}{0.48\textwidth}
        \centering
        \includegraphics[width=\linewidth]{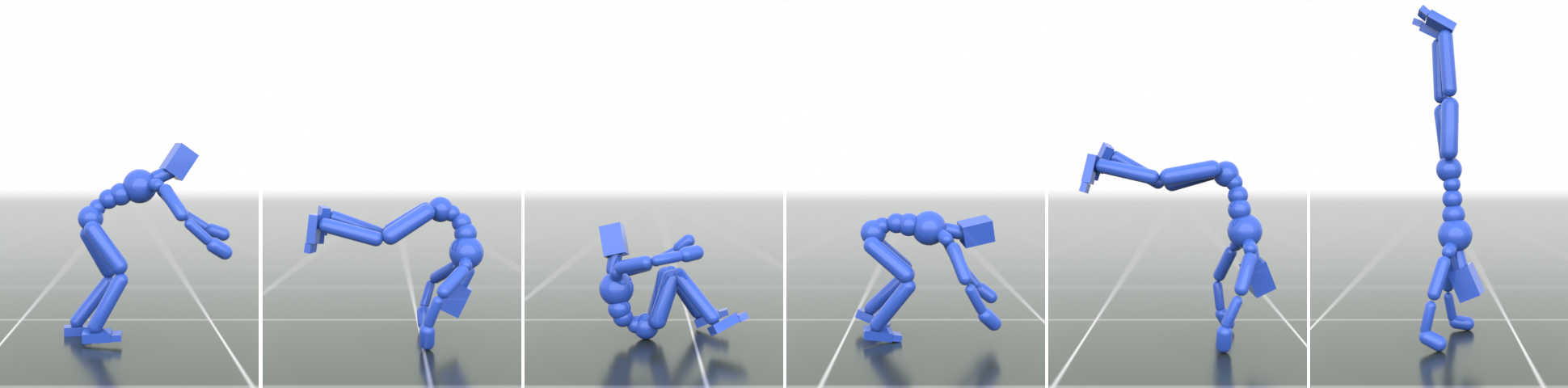}
        \small (c) Handstand.
    \end{minipage}\hfill
    \begin{minipage}{0.48\textwidth}
        \centering
        \includegraphics[width=\linewidth]{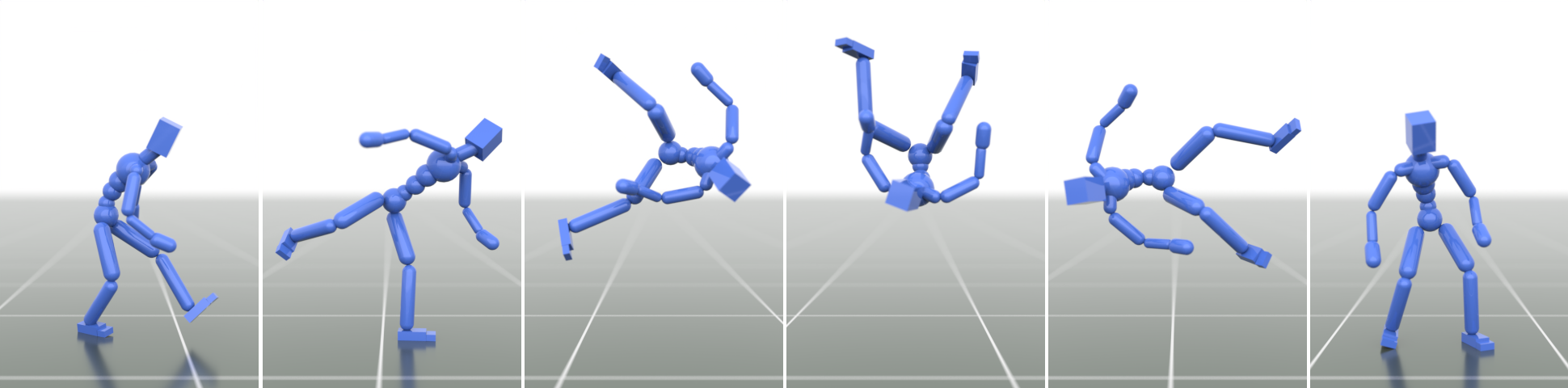}
        \small (d) Sideflip.
    \end{minipage}

    \caption{Keyframes of the results produced by the FSQ tracking controller on agile motions from Bones.}
    \label{fig:inhouse_track}
\end{figure*}

\section{Related Work}
\label{sec:related}
Our work proposes a framework for training and deploying generative controllers for physically simulated characters. The framework is centered on a generative controller that models a distribution over motor skills learned from large motion datasets. In the following, we review prior work most closely related to the key components.

\subsection{Physics-Based Character Controller}
\label{sec:related:physics}
Physics-based character animation leverages physics simulations to produce physically plausible motions. The motions of the physically simulated characters are driven by \emph{controllers}, which specify control signals for each joint in the character's body.
Early methods often rely on manually-crafted controller strategies that leverage human knowledge of a particular type of motor skill, such as walking, running, or balance recovery \citep{yin2007simbicon, coros2008stepping_walking, ye2010abstract, hodgins1995athletes, dasilva2008shorthorizon}. While effective for narrowly defined behaviors, these methods are often difficult to scale to more complex motions and require substantial manual effort to design for new skills or environments. 

To reduce this burden, subsequent work has explored example-guided reinforcement learning, in which controllers are trained using motion demonstrations, typically obtained from motion capture as dense reward signals that guide policy learning and substantially reduce the need for manually specified reward functions \citep{peng2018deepmimic, peng2017deeploco, chentanez2018physics}.
Subsequent studies further improve the generality of imitation-learning–based motion tracking by adapting controllers to different body shapes \citep{won2019learning}, reducing reliance on carefully tuned objective hyperparameters \citep{zhang2025ADD}, and scaling to large motion datasets with lengths of dozens of hours \citep{luo2023perpetual}.  

\subsection{Generative Controller}
\label{sec:related:generative}
Human motions are inherently multi-modal, with a diverse range of possible behaviors that a human may perform in a given scenario. To cope with the complexity of these behavioral distributions, some prior work adopts a hierarchical framework that separates motion generation from physical execution \citep{wang2024pacer2, xu2025parc, tevet2024closd, ye2023physdiff}. In these approaches, kinematic generative models are first trained to produce diverse motion trajectories, and then the generated trajectories are tracked by physics-based motion tracking controllers during simulation. While this decomposition allows the generative model and the physics-based controller to focus on complementary objectives, it introduces a mismatch between the two modules, undermining physical performance. Some methods instead explicitly learn skill representations that can be reused and composed for downstream tasks. For example, ASE learns reusable skill embeddings through adversarial imitation learning within a GAN framework \citep{peng2022ase}, allowing high-level policies to produce complex behaviors by composing learned primitives. Subsequent work further enhances the structure and interpretability of these representations by incorporating semantic motion labels \citep{dou2022case}, enabling more controllable and steerable behaviors in downstream tasks \citep{tessler2023calm}. \citet{tessler2024maskedmimic} learn latent skill representations using a masked-inpainting training procedure, which enables the model to be reused for new downstream tasks via flexible kinematic constraints. Training generative controllers end-to-end on large and diverse motion datasets, however, remains challenging. To address this, several methods adopt a two-stage training paradigm, in which an expert motion-tracking controller is first trained, and its behavior is then distilled into a generative model \citep{merel2019neuralprob, won2022physvae}. In this line of research, recent studies scale the expert training stage to larger datasets up to 40 hours of motion and employ more expressive generative models, such as conditional VAEs and diffusion models, to capture a broader range of behaviors \citep{luo2024pulse, truong2024pdp, huang2025diffusecloc}. Despite their successes, these methods rely on continuous latent spaces that are vulnerable to mode collapse and unstable towards off-manifold states, often leading to failures when producing highly dynamic motions.

\subsection{Skill Quantization}
\label{sec:related:quantization}
Discrete latent-variable models mitigate these issues by imposing an explicit structure on the representation space, helping constrain generation to remain closer to the data manifold. As a result, discrete representations have been widely adopted in image generation \citep{esser2021vqvgan, oord2018vqvae}, speech modeling \citep{baevski2020vqwav}, and motion synthesis \citep{starke2024categorical, guo2024momask, jiang2023motiongpt}. In physics-based character control, 
\citet{zhu2023neural} trains a generative controller that samples from a discrete latent space constructed by a VQ-VAE tracking controller. \citet{yao2024moconvq} learn a world model using a residual VQ-VAE and extend it to text-conditioned motion generation.  
\citet{plt2024bae} introduces body–part–specific quantization to achieve fine-grained control over different body parts of a character. Many prior methods rely on VQ-VAE–based discretization, which requires careful tuning and auxiliary heuristics. 
In contrast, we adopt Finite Scalar Quantization (FSQ) \citep{mentzer2023fsq}, which removes the need for a learned codebook and avoids common VQ-VAE failure modes, enabling more stable training of discrete motion representations on large-scale datasets. We then model the relationship between sequences of discrete latent codes using an autoregressive transformer, which captures the temporal structure of diverse behaviors in large motion datasets.

\subsection{Parameter-Efficient Fine-Tuning}
\label{sec:related:peft}

As the size of training datasets and models has drastically increased in recent years, efficient methods for fine-tuning large models on new tasks have become vital to the practical application of large pretrained models. PEFT methods freeze the pretrained backbone and introduce a small number of task-specific parameters to enable efficient adaptation. One class of PEFT approaches augments frozen models with auxiliary task-specific networks, such as adapters \citep{houlsby2019parameter, pfeiffer2021adapterfusion}. In diffusion models, ControlNet follows this paradigm by attaching trainable control branches to a frozen generator to introduce new conditioning signals, albeit with a relatively high parameter cost \citep{zhang2023contrlnet}.
A complementary class of PEFT methods focuses on modulating existing weights rather than introducing separate networks. LoRA injects trainable low-rank updates directly into pretrained weight matrices, significantly reducing memory and computation overhead \citep{hu2021lora}. Its successors improve on LoRA in terms of efficiency \citep{dettmers2023qlora}, and automatic rank selection \citep{zhang2023adalora}. DoRA further refines this formulation by decomposing weight updates into magnitude and directional components, improving optimization stability and expressiveness \citep{liu2024dora}. Our approach bridges these paradigms by introducing task-specific conditioning via a learned task token while employing low-rank adaptation to modulate a frozen generative prior, enabling efficient downstream adaptation without large auxiliary networks.

\section{Background}
\label{sec:background}
\subsection{Reinforcement Learning}
 We leverage Reinforcement learning (RL) in training tracking controllers and controllers for downstream tasks. RL studies how an agent can learn a policy for sequential decision-making, commonly modeled as a Markov Decision Process (MDP) \citep{sutton1998reinforcement}. At each time step $t$, the agent observes a state $\rvs_{t}$ and selects an action $a_t \sim \pi(a_t | s_t)$ according to a policy $\pi$. The environment then transitions to the next state $\rvs_{t+1}$ according to the transition dynamics $s_{t+1} \sim p(s_{t+1}| s_t, a_t)$, and the agent receives a scalar reward $r_t = r(s_t, a_t, s_{t+1})$.  The agent's objective is then to learn an optimal policy that maximizes the expected discounted return:

\begin{equation}
J(\pi) = \mathbb{E}_{\tau \sim p(\tau | \pi)} \left[ \sum_{t=0}^{T-1} \gamma^t r_t \right],
\end{equation}
where $\gamma\in[0,1)$ is a discount factor that balances short-term and long-term rewards.

\subsection{Vector Quantization}
Vector-Quantized Variational Autoencoder (VQ-VAE) introduces discrete latent representations by mapping continuous embeddings from an encoder to the nearest entry in a discrete codebook. The encoder maps an input $\mathbf{x}$ to a latent $\mathbf{z}=\mathcal{E}(\mathbf{x})$, which is then quantized via nearest-neighbor lookup in a learned codebook $\mathcal{C}=\{\mathbf{e}_k\}_{k=1}^K$:
\begin{equation}
k^* = \arg\min_k \lVert \mathbf{z} - \mathbf{e}_k \rVert_2.
\end{equation}
The decoder then reconstructs the data by taking the embeddings from the codebook $\hat{\mathbf{x}} = \mathcal{D}(\mathbf{e}_k)$. 
Training a VQ-VAE requires three loss components: 1) the reconstruction loss that encourages the decoder to reproduce the input from the quantized latent codes, 2) A codebook loss that updates the embedding vectors towards the encoder's outputs, ensuring that the discrete codes represent the data distribution, 3) a commitment loss that penalizes deviations of the encoder outputs from their assigned codebook entries, encouraging stable code usage by penalizing large deviations between the encoder output and its selected codebook vector, which discourages frequent switching between different codes for similar input or same input during successive updates during training.
The VQ-VAE training objective is then given by,
\begin{equation}
\mathcal{L_{VQ}}
= \lVert \mathbf{x} - \hat{\mathbf{x}} \rVert_2^2
+ \lVert \mathrm{sg}[\mathcal{E}(x)] - \mathbf{e}_{k^*}\rVert_2^2
+ \beta \lVert \mathbf{z} - \mathrm{sg}[\mathbf{e}_{k^*}] \rVert_2^2 \,.
\end{equation}

VQ-VAEs are difficult to train as the discrete codebook updates can suffer from poor utilization and codebook collapse. These issues are typically addressed using heuristics such as exponential moving average (EMA) updates for the codebook, and periodically reinitialization of unused codes to encourage high codebook utilization.

\begin{figure}[htbp]
    \centering
    \begin{minipage}{\columnwidth}
        \centering
        \includegraphics[width=\linewidth]{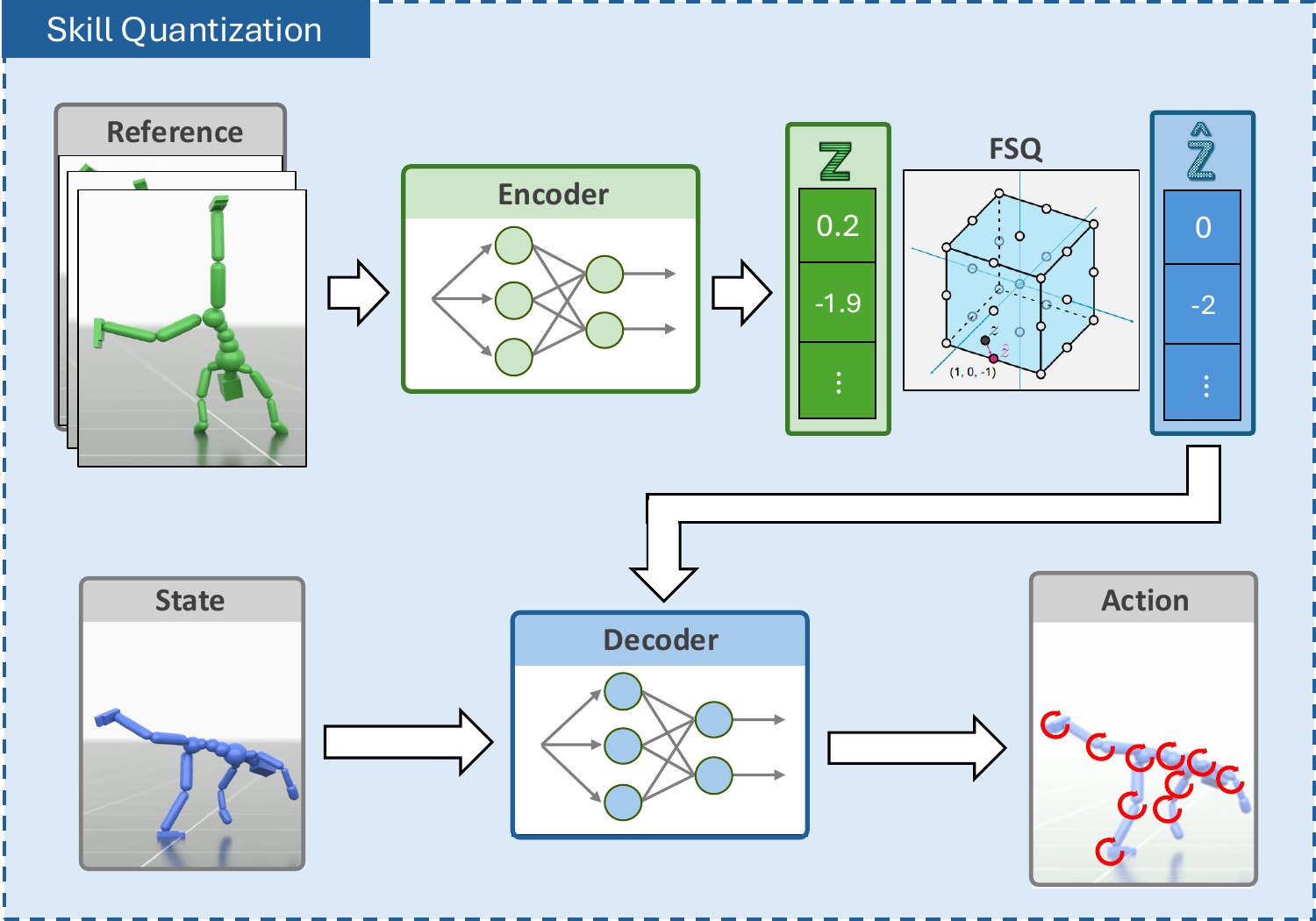}
        \includegraphics[width=\linewidth]{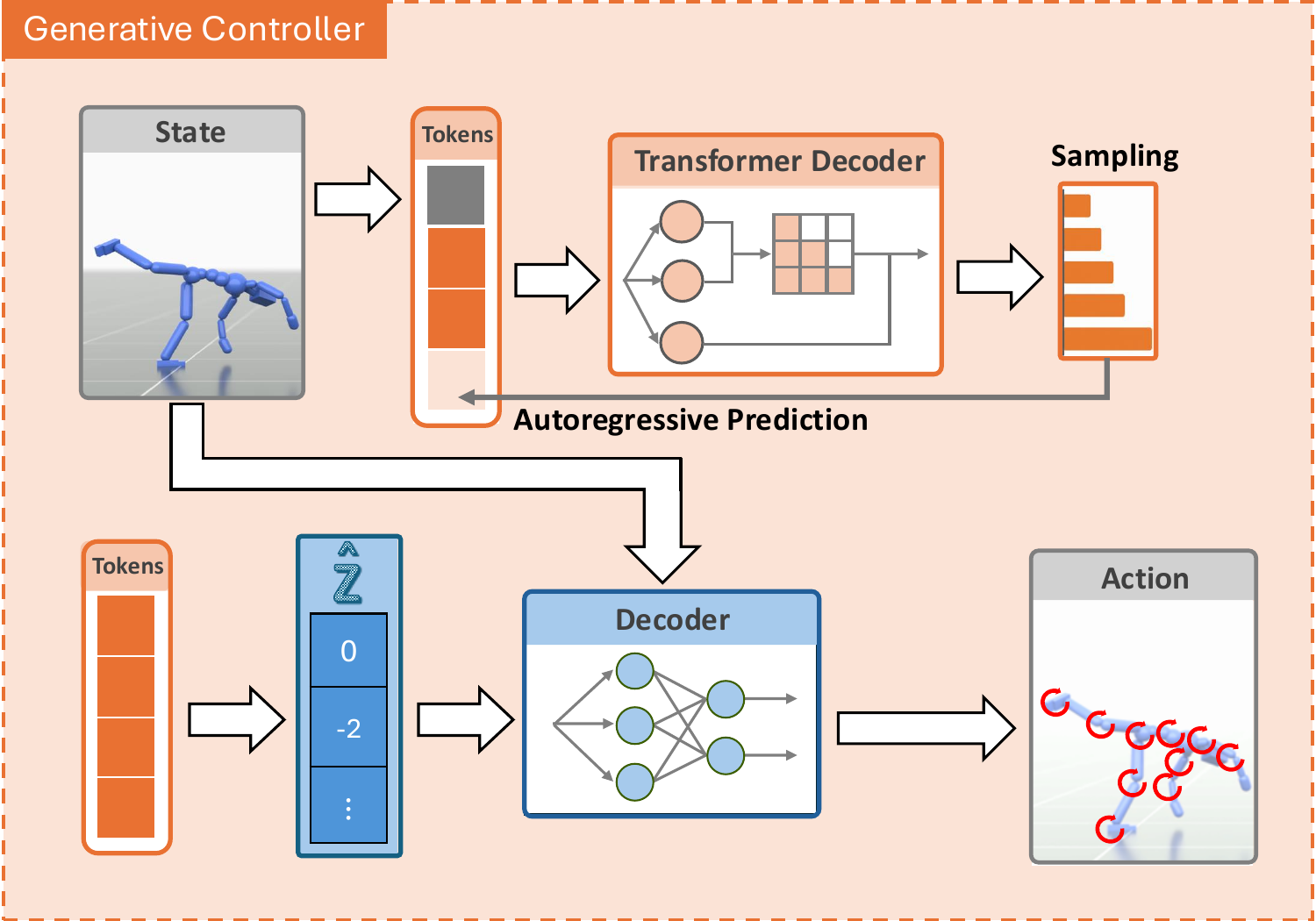}
        \includegraphics[width=\linewidth]{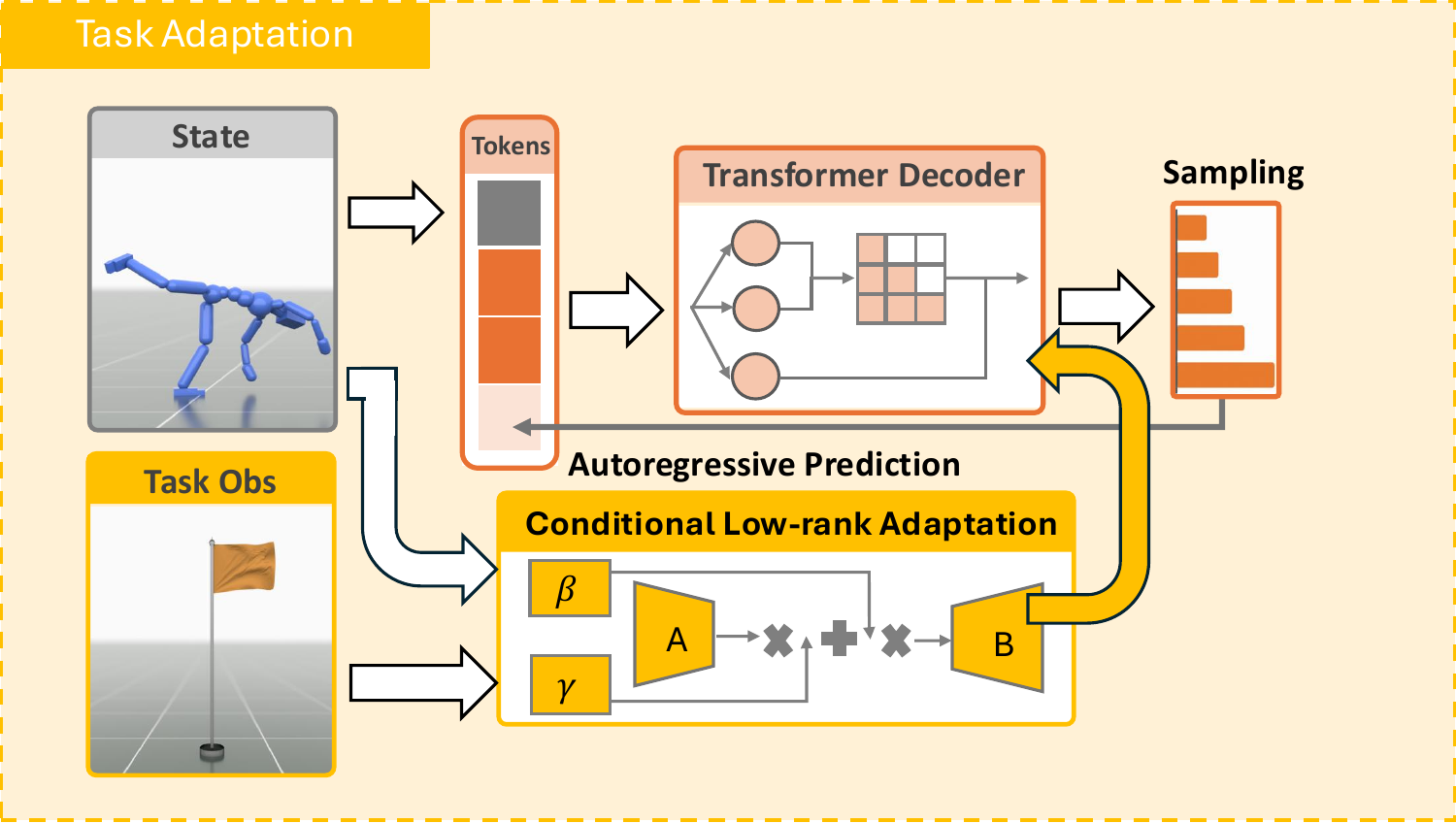}
    \end{minipage}

  \caption{ 
\textbf{Skill quantization}: An FSQ motion tracking controller maps reference motions to discrete latent tokens and decodes them to actions. The FSQ module is trained end-to-end with reinforcement learning using a motion-tracking objective (top). 
\textbf{Generative controller training}: a transformer decoder models the skill distribution via causal self-attention, enabling autoregressive sampling of discrete skill tokens conditioned on the character state $\mathbf{s}$. The generative controller is trained with teacher forcing and cross-entropy loss to predict the tokens produced by FSQ (middle).   
\textbf{Task adaptation}: Lightweight CoLA layers are added to adapt the frozen pretrained generative controller to complete downstream tasks (bottom). This adaptation is parameter-efficient, adding less than $1\%$ additional parameters to the model. }
\label{fig:pipeline2}
\end{figure}

\section{Framework Overview}
\label{sec:method}

\begin{figure*}[t]
    \centering
    \begin{minipage}{0.24\textwidth}
        \centering
        \includegraphics[width=\linewidth]{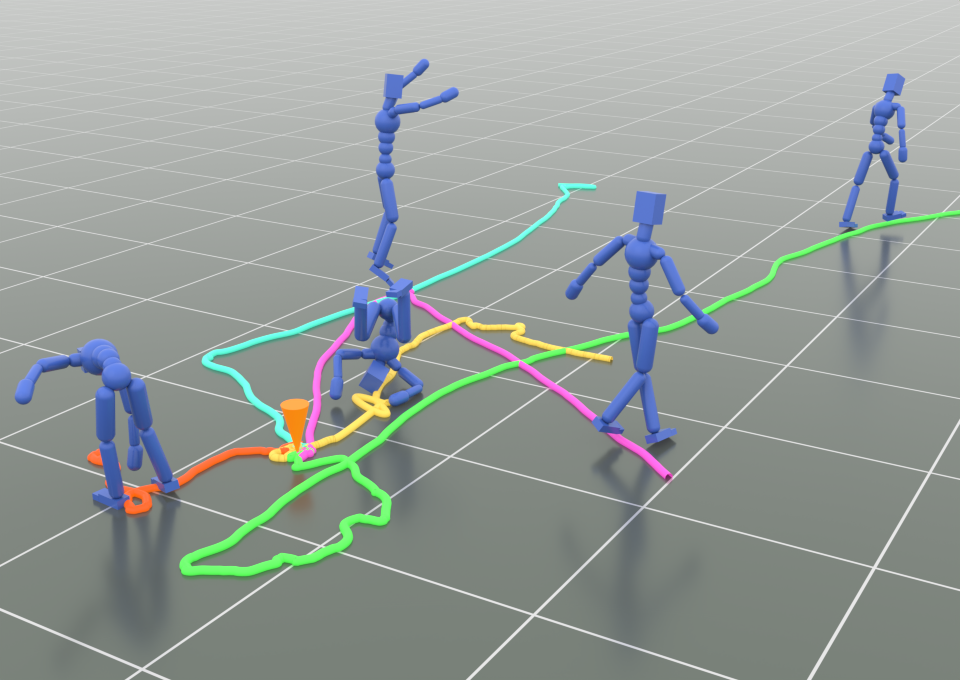}
        \small (a) Trajectories from unconditional sampling with GPC.
    \end{minipage}\hfill
    \begin{minipage}{0.24\textwidth}
        \centering
        \includegraphics[width=\linewidth]{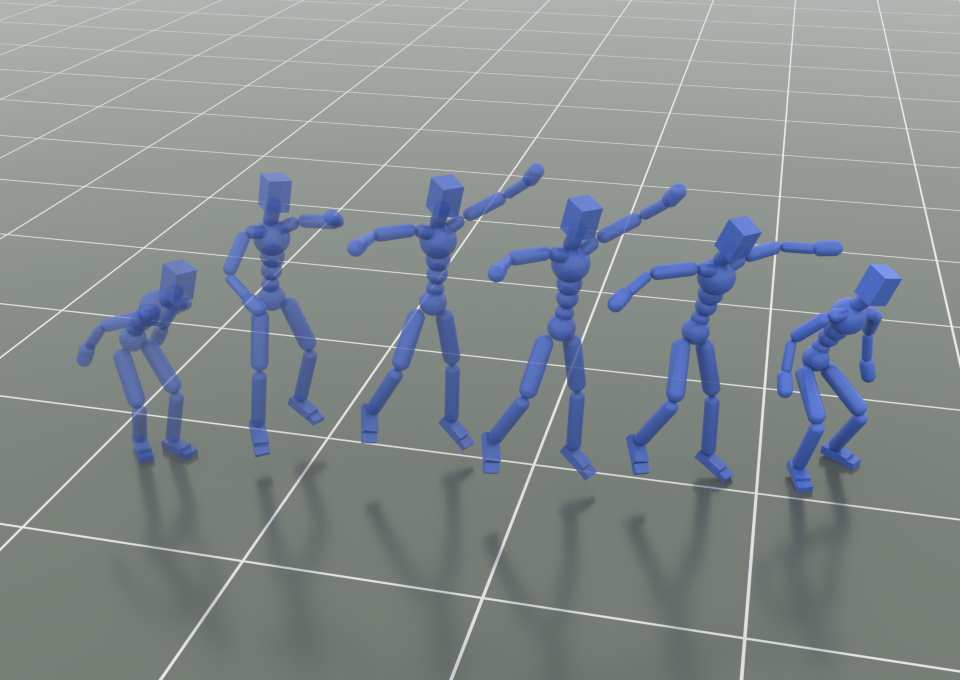}
        \small (b) Jumping skill synthesized by the generative controller.
    \end{minipage}\hfill
    \begin{minipage}{0.485\textwidth}
        \centering
        \includegraphics[width=\linewidth]{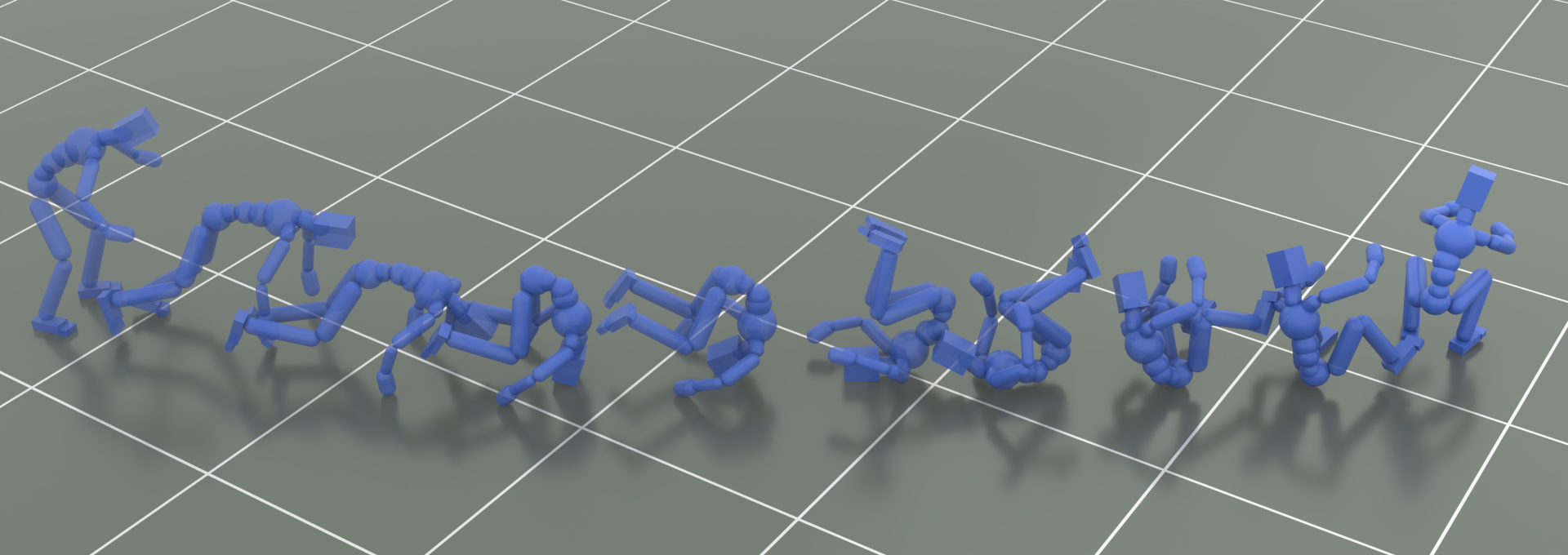}
        \small (c) A motion sequence featuring a forward leap, front roll, and transition to a squat.
    \end{minipage}
    \caption{Unconditional sampling of the generative controller. GPC produces a wide array of highly dynamic skills, such as jumping, leaping, and rolling.}
    \label{fig:uncond}
\end{figure*}

\begin{figure*}[t]
    \centering
    \begin{minipage}[t]{0.24\textwidth}
        \centering
        \includegraphics[width=\linewidth]{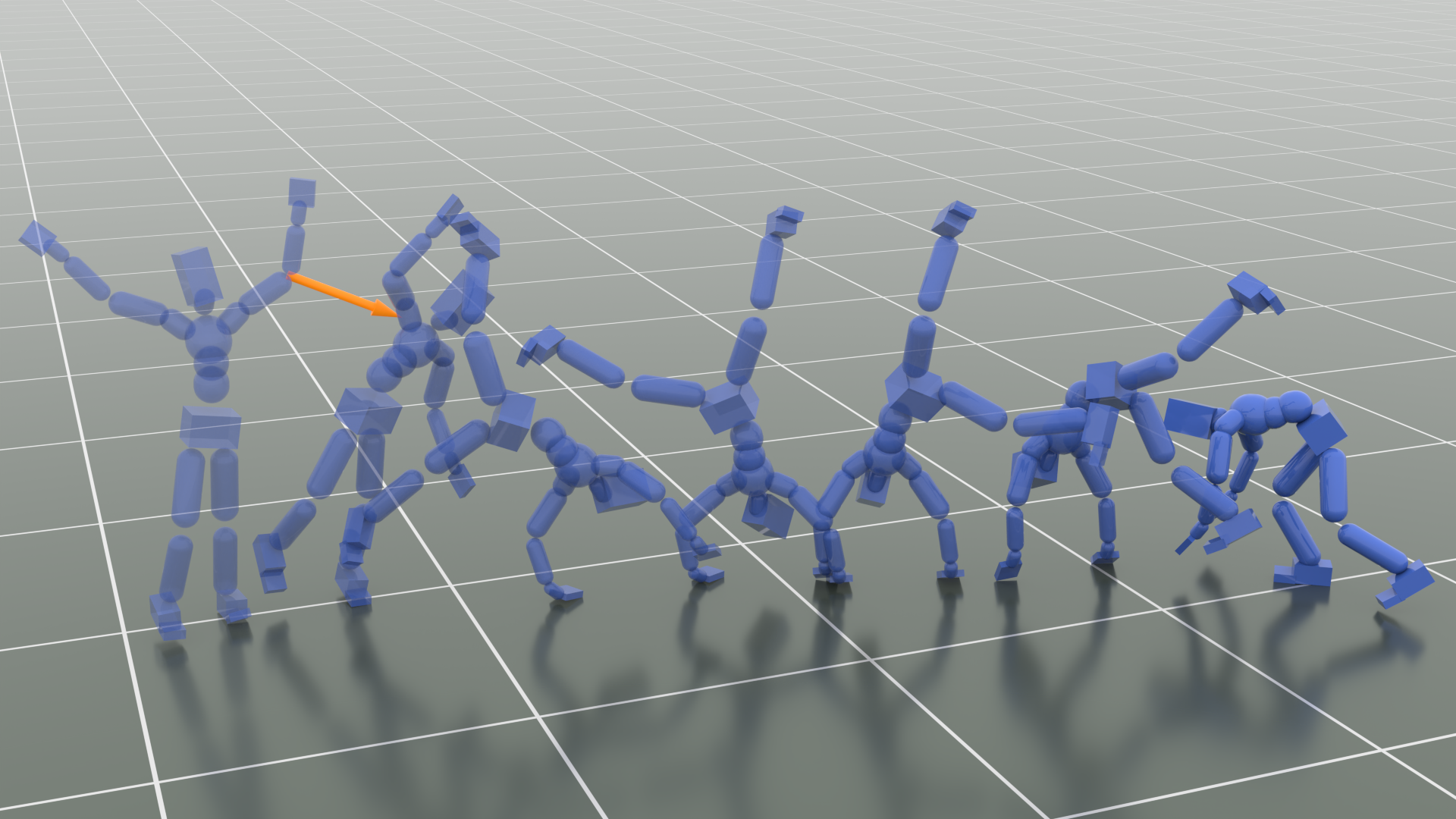}
        \small (a) A perturbation force on the arm triggers a cartwheel-like recovery. 
    \end{minipage}\hfill
    \begin{minipage}[t]{0.24\textwidth}
        \centering
        \includegraphics[width=\linewidth]{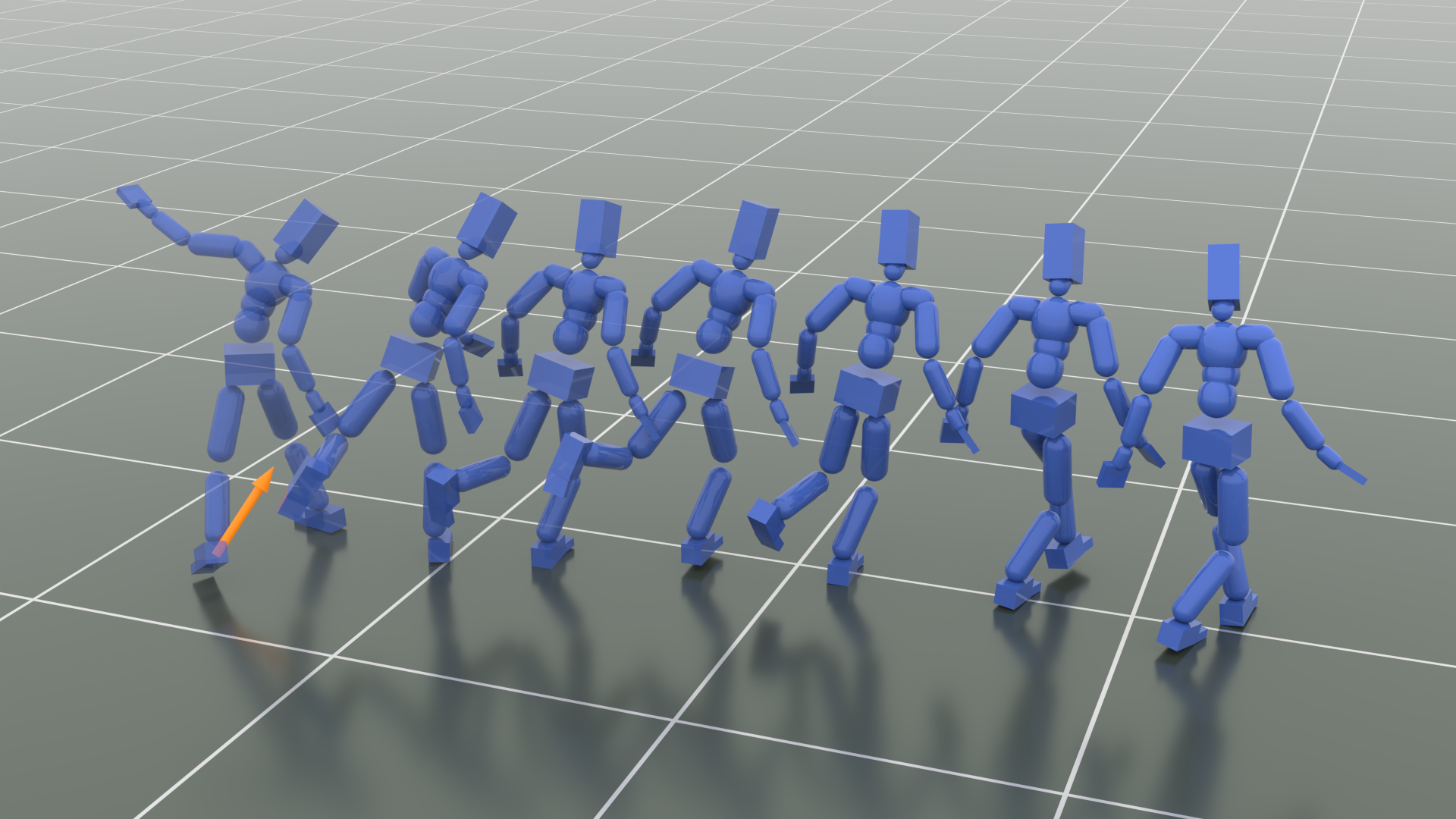}
        \small (b) After a perturbation force is applied to the character's leg, it automatically adjusts its step to regain balance.
    \end{minipage}\hfill
    \begin{minipage}[t]{0.24\textwidth}
        \centering
        \includegraphics[width=\linewidth]{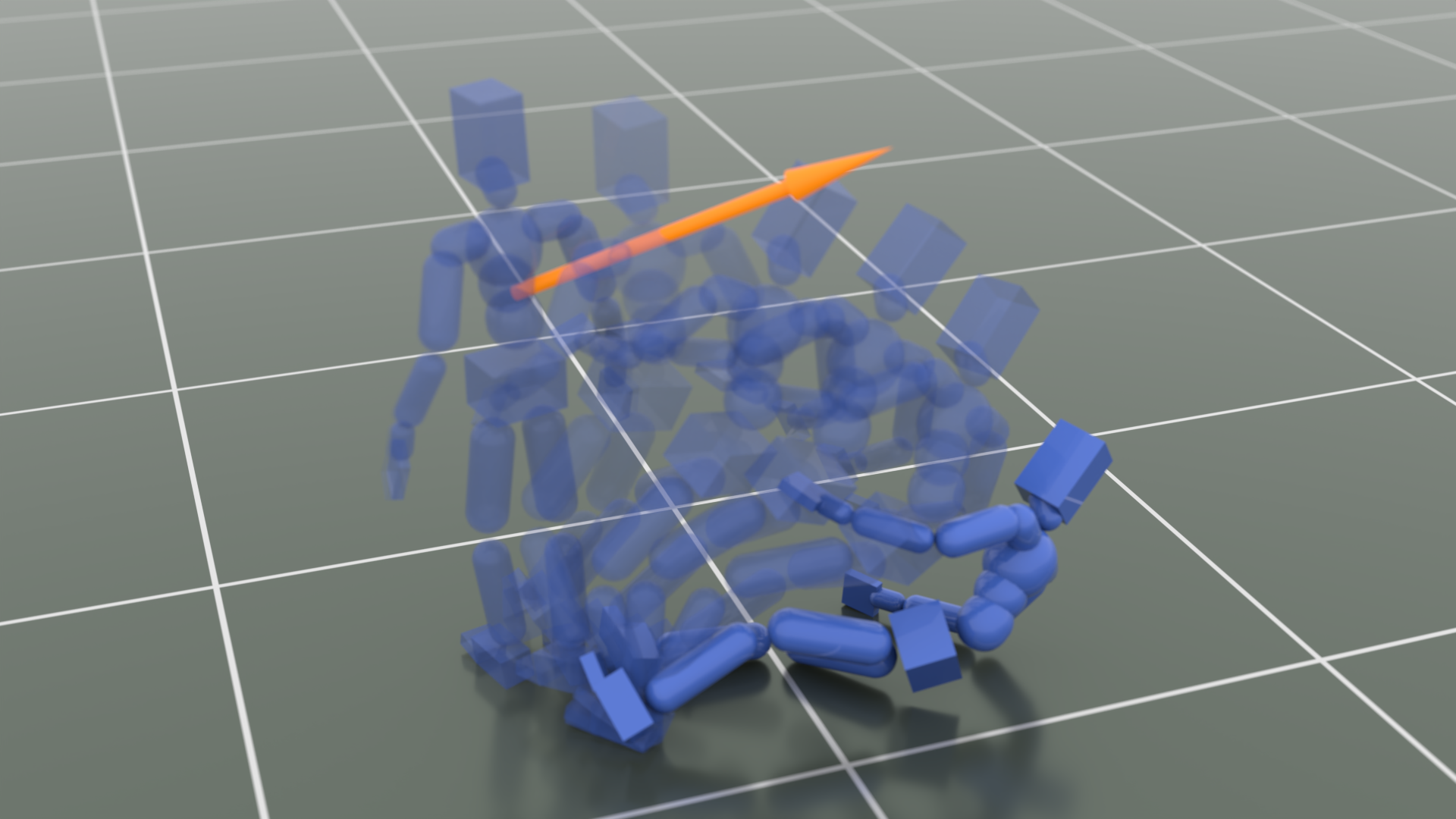}
        \small (c) A strong external force applied to the spine destabilizes the character and causes it to fall in a natural way, and is ready to recover.
         \end{minipage}\hfill
    \begin{minipage}[t]{0.24\textwidth}
        \centering
        \includegraphics[width=\linewidth]{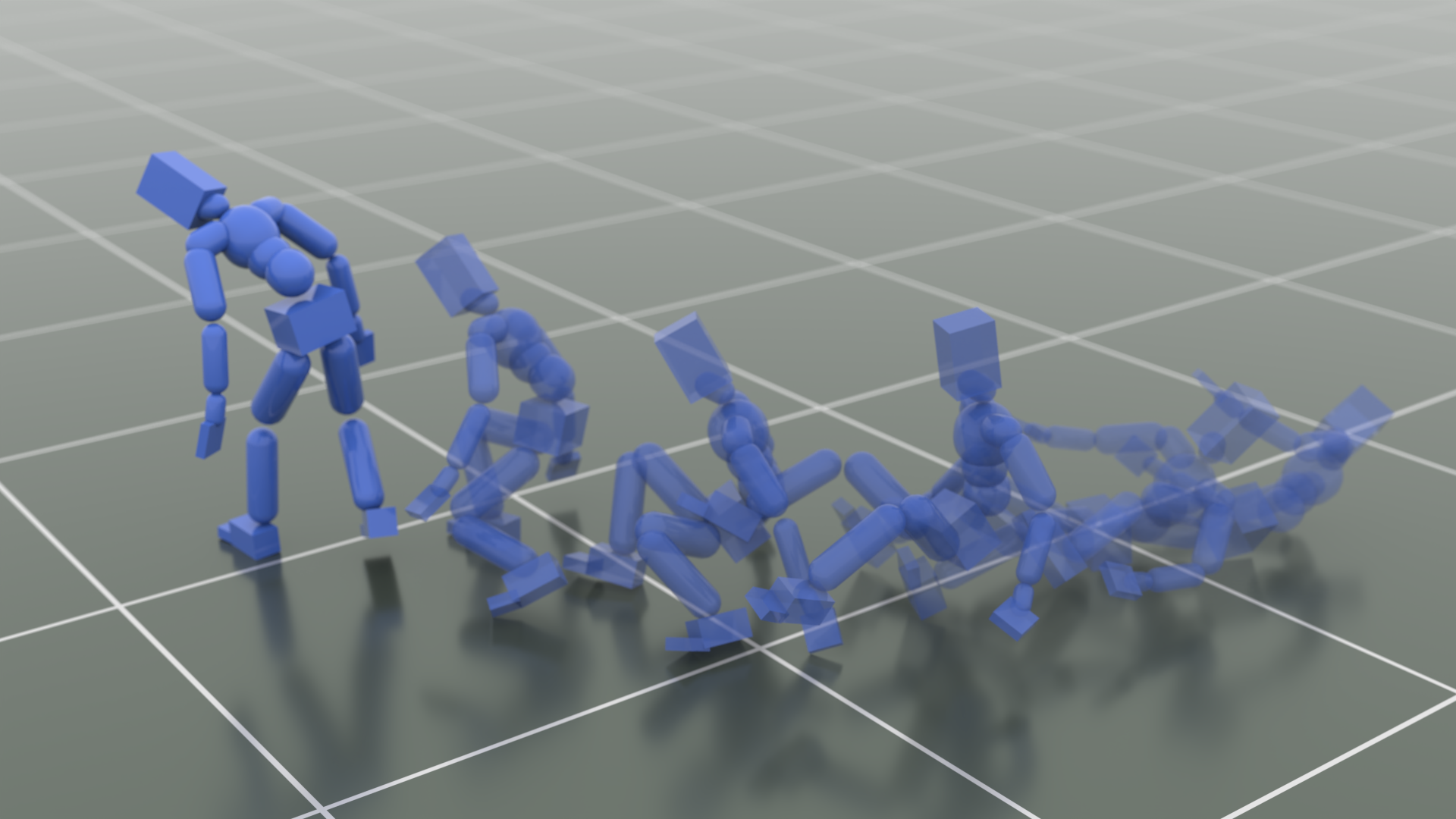}
        \small (d) The character executes a get-up skill and recovers to a standing pose from a fallen state shown in Figure~\ref{fig:amass_force}(c).
    \end{minipage}
    \caption{GPC produces robust and natural recovery behaviors when subjected to external perturbations.}
    \label{fig:amass_force}
\end{figure*}

\begin{figure*}[t]
    \centering
    \begin{minipage}{0.485\textwidth}
        \centering
        \includegraphics[width=\linewidth]{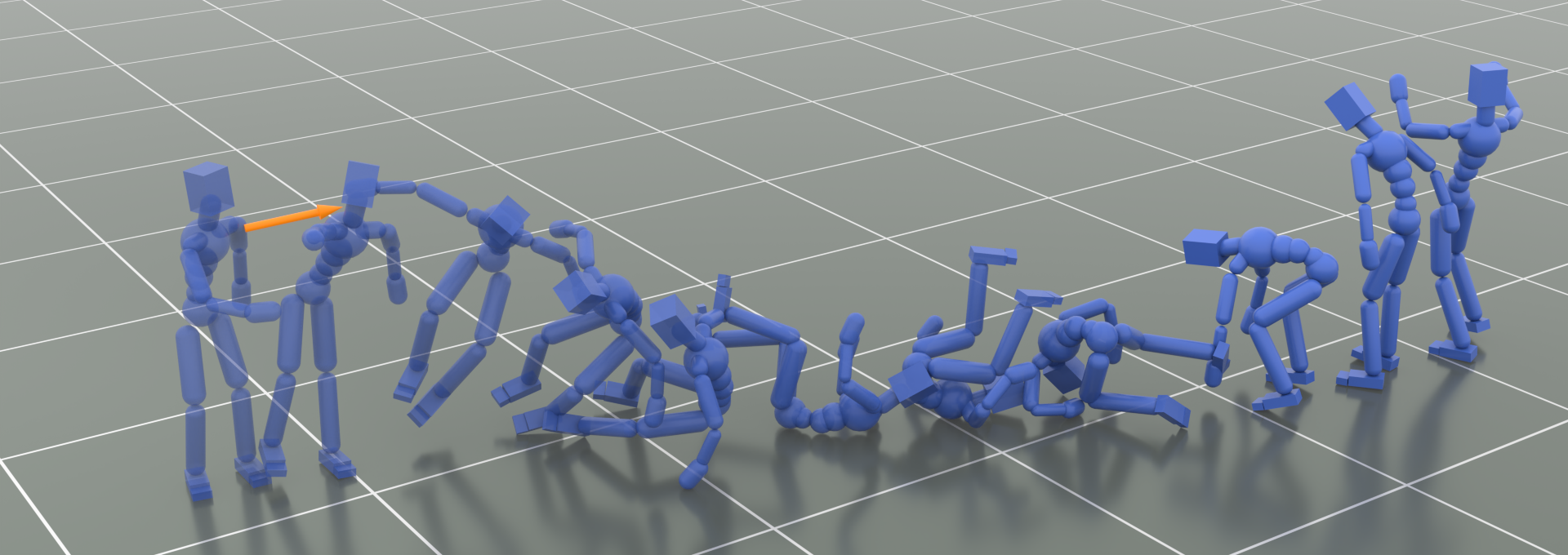}
        \small (a) When a character falls down due to external forces, it automatically transitions to rolling skills to recover and get back up.
    \end{minipage}\hfill
    \begin{minipage}{0.485\textwidth}
        \centering
        \includegraphics[width=\linewidth]{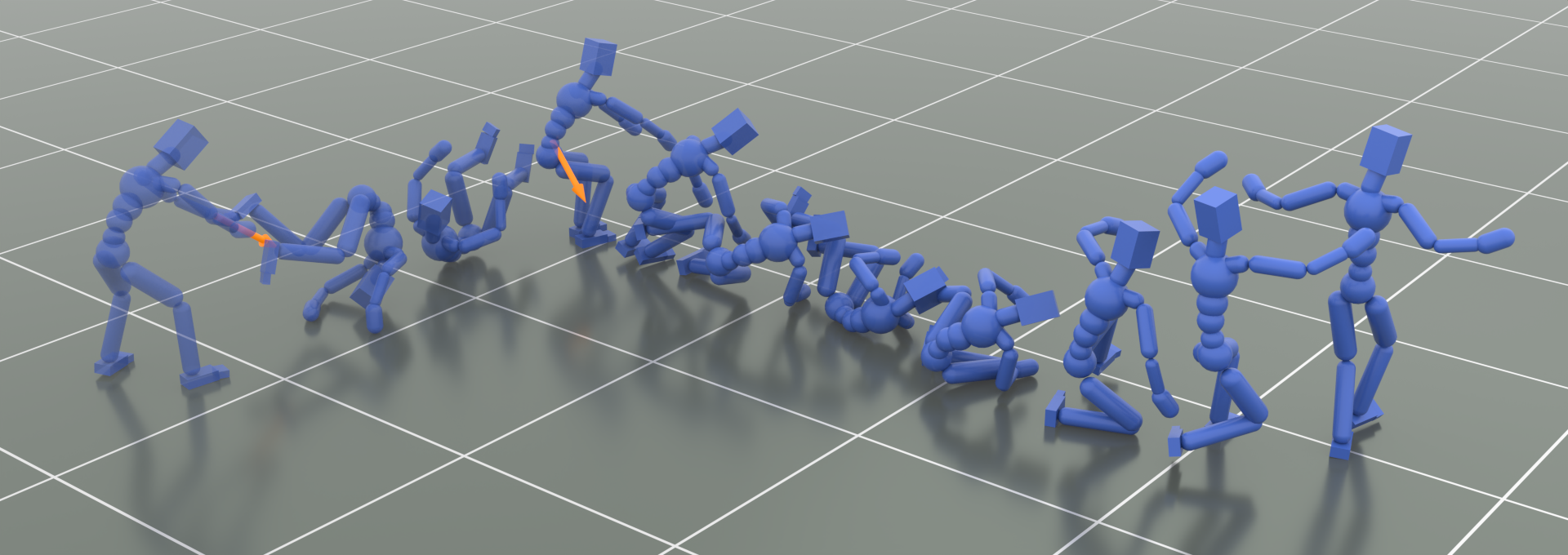}
        \small (b) The character stabilizes with a front roll and gets up when encountering two subsequent forces.
    \end{minipage}
    \caption{GPC exhibits diverse and versatile responses to force perturbations with a generative controller trained with Bones.}
    \label{fig:inhouse_force}
\end{figure*}
In this work, we present a framework for training generative controllers on large-scale motion datasets. These general-purpose controllers are capable of modeling large repertoires of motor skills for physics-based character animation. As illustrated in Fig.~\ref{fig:pipeline2}, the framework consists of three stages. 

In the first stage, Skill Quantization (Section~\ref{method:tracking}), we construct a discrete latent representation of skills using Finite Scalar Quantization (FSQ). The latent codes are optimized directly with end-to-end RL to model a wide range of motor skills from a large motion dataset. This end-to-end RL training helps to ensure that the learned discrete codes correspond to skills that can be faithfully executed by a physically simulated controller. This provides a robust foundation for subsequent generative controller training and downstream task adaptation. 

In the second stage, Generative Controller Training (Section~\ref{method:gc}), we train a generative controller that models the distribution of discrete skill tokens using a transformer decoder. The generative controller autoregressively generates sequences of tokens that drive the character to produce naturalistic behaviors. Given the character's state and a selected latent token, the pretrained decoder outputs actions that drive the movement of the character's body. This method leads to the emergence of rich, reusable behaviors such as life-like responses and recoveries from perturbations. These behaviors emerge without requiring explicit rewards or specialized training. Finally, in Section~\ref{method:peft}, we introduce a parameter-efficient fine-tuning (PEFT) procedure that can adapt the pretrained generative controller to a wide range of downstream control tasks using only a small number of additional parameters, while preserving the diversity and naturalness of the pretrained controller.

\begin{figure*}[t]
    \centering
    \begin{minipage}{0.485\textwidth}
        \centering
        \includegraphics[width=\linewidth]{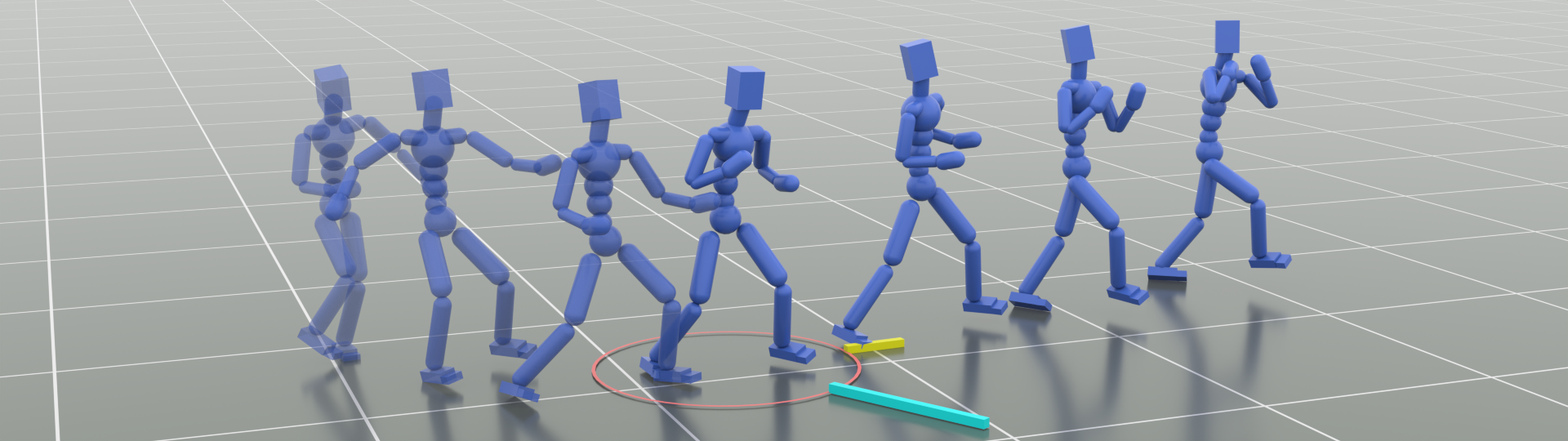}
        \small (a) Steering.
    \end{minipage}\hfill
    \begin{minipage}{0.485\textwidth}
        \centering
        \includegraphics[width=\linewidth]{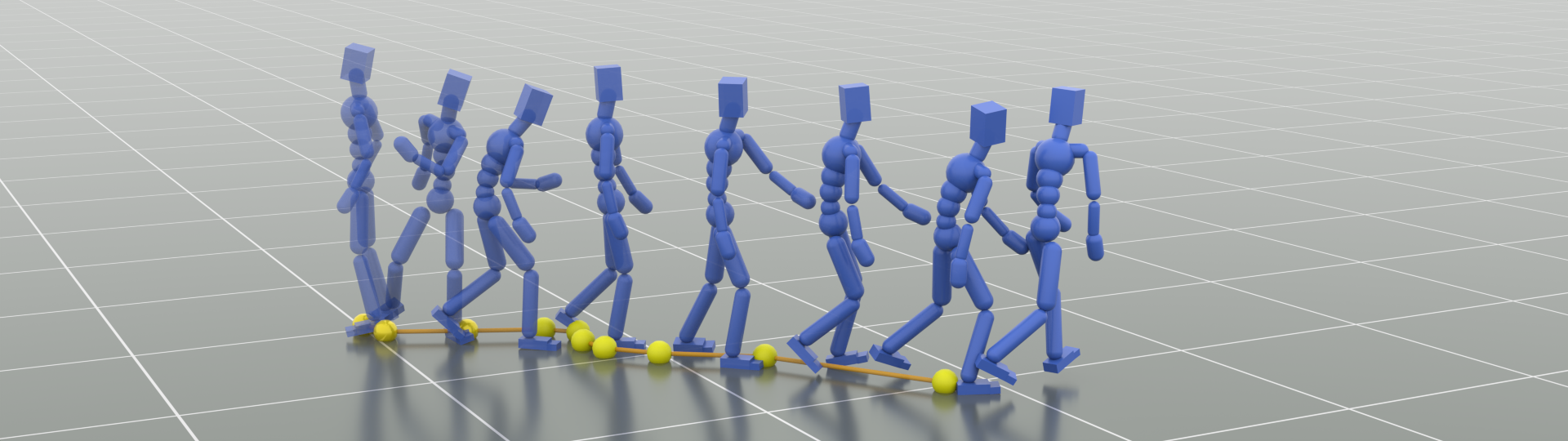}
        \small (b) Trajectory.
    \end{minipage}
    
    \begin{minipage}{0.485\textwidth}
        \centering
        \includegraphics[width=\linewidth]{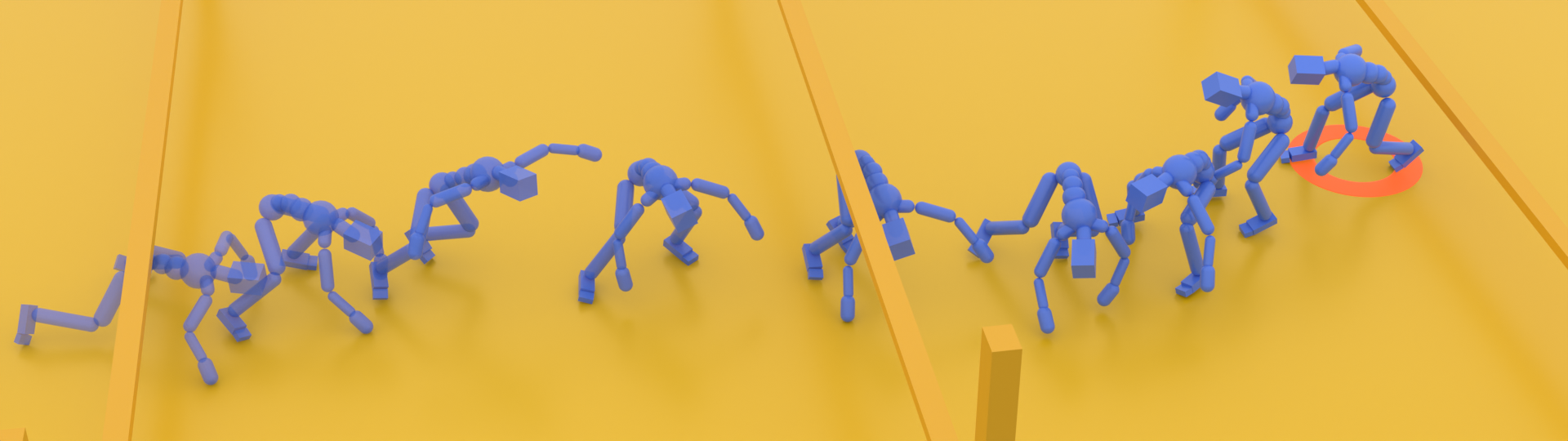}
        \small (c) Barrier.
    \end{minipage}\hfill
    \begin{minipage}{0.485\textwidth}
        \centering
        \includegraphics[width=\linewidth]{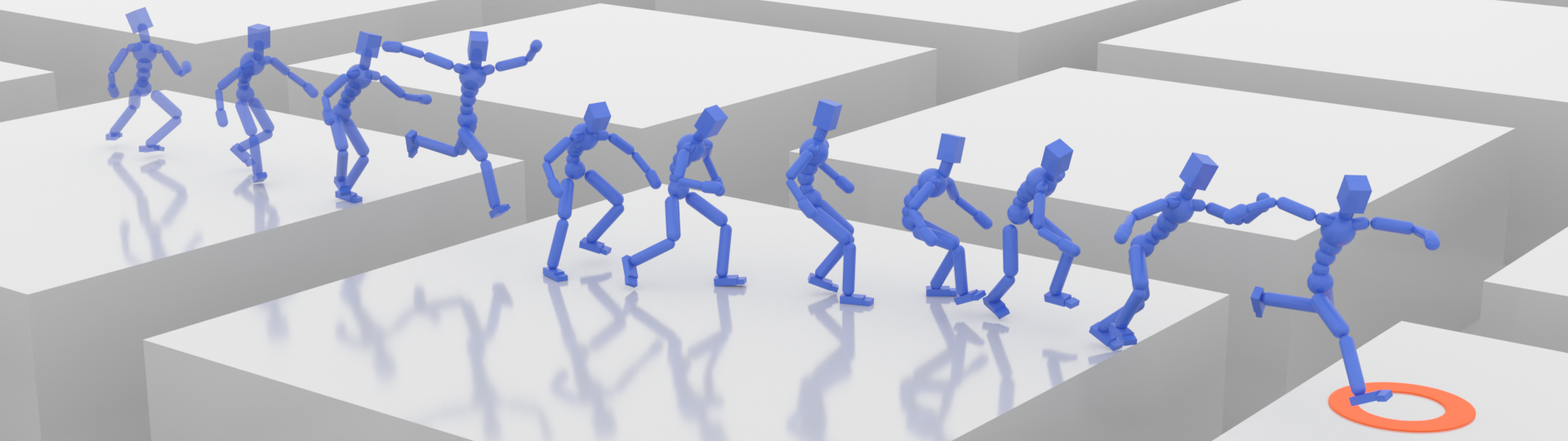}
        \small (d) Platform.
    \end{minipage}
    \caption{CoLA can be used to efficient finetune GPC to new tasks. By only training adaptation layers during finetuning, the pretrained generative controller can effectively complete downstream tasks while preserving natural behaviors encoded in the original base model.}
    \label{fig:inhouse_task}
\end{figure*}

\section{Skill Quantization}

\label{method:tracking}
Training GPC requires a discrete latent representation of motor skills. This is obtained by training an FSQ motion-tracking controller, where the discrete codes are optimized via end-to-end RL to model different skills. We follow a standard motion-tracking framework as described in DeepMimic~\citep{peng2018deepmimic}. 
Given a reference motion sequence which specifies target states $\hat{\mathbf{s}}_{t:T}$ at each timestep $t$, a policy $\pi_\theta(\mathbf{a}_t \mid \mathbf{s}_t, \hat{\mathbf{s}}_{t:t+h})$ is trained to select the appropriate sequence of actions that enable the simulated character to imitate the reference motion.  The policy is conditioned on the character’s proprioceptive state $\mathbf{s}_t$ at time step $t$ and a sequence of $h$ future target states from the reference motion. The training objective is defined via a tracking reward $\mathbf{r}_t$. The reward measures the discrepancy between the simulated character state $\mathbf{s}_{t+1}$ and the reference motion. While this formulation is sufficient for accurately reproducing a wide range of behaviors~\citep{luo2023perpetual,tessler2024maskedmimic}, our goal is to learn a structured and reusable \emph{discrete} representation of skills that supports generative modeling and downstream control. We incorporate FSQ-based quantization into the policy through an encoder–decoder design architecture. The encoder maps a sequence of target states $\hat{\mathbf{s}}_{t:t+h}$ to a continuous latent vector $\mathbf{z_t} = \mathcal{E}(\hat{\mathbf{s}}_{t:t+h}) \in \mathbb{R}^d$. Each dimension of this latent is then independently quantized using FSQ into $L$ fixed scalar levels,
\begin{equation}
\hat{\mathbf{z}}_t = \left\lfloor \left\lfloor \tfrac{L}{2} \right\rfloor \tanh(\mathbf{z}_t) \right\rceil ,
\end{equation}
where $\lfloor \cdot \rceil$ denotes element-wise rounding. This look-up-free quantization defines an implicit discrete codebook of size $L^d$, alleviating the need to learn an explicit codebook, as done in VQ-VAEs \citep{oord2018vqvae}, and thereby mitigating common pitfalls of VQ-VAEs such as codebook collapse~\citep{mentzer2023fsq}. The resulting discrete latent code $\hat{\mathbf{z}}_t \in \mathbb{R}^d$ serves as a compact representation of the target motion $\hat{\mathbf{s}}_{t:t+h}$.
The decoder $\mathcal{D}$ receives the character's current state and discrete code as input, and outputs the actions, $
\mathbf{a}_t \sim \mathcal{D}(\mathbf{a}_t \mid \mathbf{s}_t, \hat{\mathbf{z}}_t) $. The actions represent target joint rotations for proportional-derivative (PD) controllers positioned at each of the character's joints. The PD controller converts these targets into joint torques that are applied to drive the movement of the simulated character.
The encoder–decoder policy is trained end-to-end with Proximal Policy Optimization (PPO) ~\citep{schulman2017proximal}. Gradients are propagated through the FSQ quantization operation using a straight-through estimator (STE) ~\citep{bengio2013ste}.

\section{Generative Controller}
\label{method:gc}
The encoder from the previous quantization stage can produce a discrete latent code. This representation yields a sequence of \(d\) discrete tokens at each time step. However, directly modeling all \(d\) \(L\)-ary tokens results in long token sequences, which increases the computational and memory cost of autoregressive modeling. To reduce sequence length, a fixed grouping scheme is applied that packs every \(G\) consecutive \(L\)-ary tokens into a token with a larger vocabulary. In our experiments, we find this simple grouping scheme for creating grouped token strikes an effective balance between vocabulary size and sequence length. This method reduces both the context length and the cost of self-attention as shown in Table \ref{tab:grouping_ablation}.

The grouped token sequence \(\tilde{\mathbf{\rvz}}_t = (\tilde{\rvz}_t^0, \ldots, \tilde{\rvz}_t^{d'-1})\) can be treated as a skill representation, 
and its joint conditional distribution is modeled autoregressively such that each token is generated based on the character's current state and all previously generated tokens:
\begin{equation}
p_\theta(\tilde{\mathbf{z}}_t \mid \mathbf{s}_t)
= p_\theta(\tilde{\rvz}_{t}^0 \mid \mathbf{s}_t)
\prod_{j=1}^{d'-1} p_\theta(\tilde{\rvz}_t^j \mid \mathbf{s}_t, \tilde{\mathbf{z}}_t^{<j}).
\label{eq:ar_factorization}
\end{equation}
This factorization is modeled using a GPT-style transformer decoder with causal self-attention, where each token $j$ attends only to previously generated tokens $\tilde{\mathbf{z}}_t^{<j}$, ensuring consistency between training and inference. At each step, the transformer predicts a conditional categorical distribution over the next latent code $\tilde{\rvz}_t^j$.
The model is trained using a cross-entropy objective,
\begin{equation}
\mathcal{L}_{\text{CE}} = - \sum_{j=0}^{d'-1} \log p_\theta\!\left(\tilde{\rvz}_t^j \mid \mathbf{s}_t, \tilde{\mathbf{z}}_t^{<j}\right).
\label{eq:ce_loss}
\end{equation}
This training objective encourages the transformer to capture the temporal dependencies among sequences of grouped tokens.

During inference time, at each timestep $t$, we autoregressively sample latent tokens $\tilde{z}_t^j \sim p(\tilde{z}_t^j \mid \mathbf{s}_t, \tilde{\mathbf{z}}_t^{<j})$ to determine the action that should be executed by the controller. At each decoding step, the transformer produces a categorical distribution over \(L^{G}\) discrete codes, from which the next token is sampled using nucleus (top-\(p\)) sampling applied to the softmax-normalized logits~\citep{holtzman2020neucls}. Nucleus sampling restricts sampling to a subset of the most likely tokens, which mitigates the chances of sampling low-probability outliers while still preserving diversity in the generated behaviors. The sampled token is then provided as input at the subsequent step to generate the next token. The final sequence of tokens \(\tilde{\mathbf{z}}_t\) is decoded by the frozen FSQ decoder to generate an action given the character's current state. 

\section{Task Adaptation}
\label{method:peft}
Once trained, GPC can be adapted to task-specific objectives while preserving pretrained behaviors. However, finetuning the full model for each downstream task can be computationally and data intensive. To address this, we introduce Conditional Low-rank Adaptation (CoLA), which enables parameter-efficient fine-tuning (PEFT) through lightweight adaptation layers that introduce less than $1\%$ additional parameters to the model.

\subsection{Conditional Low-rank Adaptation}
CoLA serves as a lightweight adaptation framework that enables specialized learning without the overhead of updating the entire model architecture. CoLA extends the DoRA strategy by incorporating task-specific modulation within a low-rank space \citep{liu2024dora}. Given a pretrained weight matrix $\mathbf{W}_0 \in \mathbb{R}^{d_{\text{out}} \times d_{\text{in}}}$, the adaptation decomposes weight updates into separate magnitude and direction components. To condition the model on task-specific observations $\mathbf{c}$, CoLA applies Feature-wise Linear Modulation (FiLM) to the low-rank components \citep{perez2018film},
\begin{equation}
\mathbf{W x} =
\mathbf{W}_0 \mathbf{x} + \mathbf{m}
\frac{\mathbf{B}\left(\text{diag}(\boldsymbol{\gamma}(\mathbf{c})) \, \mathbf{A} \mathbf{x}
+ \boldsymbol{\beta}(\mathbf{c})\right)}
{\left\lVert\mathbf{B}\left(\text{diag}(\boldsymbol{\gamma}(\mathbf{c})) \, \mathbf{A} \mathbf{x}
+ \boldsymbol{\beta}(\mathbf{c})\right)\right\rVert_F} ,
\end{equation}
where  $\mathbf{x}$ is the input to a particular layer, $m$ is a learned vector controlling the overall magnitude. $\mathbf{A} \in \mathbb{R}^{r \times d_{\text{in}}}$ and $\mathbf{B} \in \mathbb{R}^{d_{\text{out}} \times r}$ are trainable low-rank matrices with $r \ll \min(d_{\text{in}}, d_{\text{out}})$. $\boldsymbol{\gamma}(\mathbf{c})$ and $\boldsymbol{\beta}(\mathbf{c})$ are produced by lightweight MLPs, and apply task-conditioned modulation in the low-rank space $\mathbb{R}^{r}$. This design ensures stability through decoupled updates, while providing flexible task modulation with minimal additional parameters.

\subsection{Fine-Tuning via Reinforcement Learning}
Task adaptation is performed via reinforcement learning fine-tuning (RLFT) by optimizing task-specific reward functions. Training is carried out using Proximal Policy Optimization (PPO) \citep{schulman2017proximal}. The action space consists of sequences of discrete skill tokens with length $d'$, where each token is drawn from an $L^{G}$-ary categorical distribution corresponding to grouped latent skill representations $\tilde{\mathbf{z}}$. At each timestep, the controller autoregressively predicts a token sequence $\tilde{\mathbf{z}}$ conditioned on the current character state and task observations, which is decoded into actions by the frozen FSQ decoder. The generative adapted controller outputs logits over an $L^{G}$-ary categorical distribution for each token, and per-token log-probabilities are aggregated to form the PPO objective, with advantages shared across tokens within the same decision step. During rollouts, exploration is regularized using Nucleus Sampling guided by the unconditional generative model. Nucleus sampling only samples tokens from high-probability regions of the predicted distribution, thereby helping to preserve natural behaviors when fine-tuning GPC on downstream tasks.

\subsection{Supervised Fine-Tuning}
In addition to reinforcement learning fine-tuning, supervised fine-tuning (SFT) provides an effective mechanism for guiding task adaptation when example motions are available. Given a large pretrained generative controller that models a diverse set of skills, exploration during task training can become inefficient. SFT mitigates this challenge by leveraging example motions to bias adaptation toward a subset of skills that are appropriate for the target task. The generative controller is adapted by minimizing a cross-entropy loss over the discrete latent codes, effectively increasing the likelihood of dataset-specific skill tokens conditioned on the character state. This fine-tuning process biases the controller toward selecting desired skills when performing exploration during RL finetuning. The generative controller is first adapted via supervised fine-tuning (SFT) by minimizing a cross-entropy loss over the discrete latent codes, which increases the likelihood of selecting task-relevant skill tokens during exploration in subsequent RL fine-tuning. Starting from the SFT-adapted initialization, the model can then be further refined through reinforcement learning fine-tuning (RLFT) to reliably compose and execute these skills for task completion.

\begin{table}[t]
\centering
\caption{
Skill quantization performances. We compare our method trained end-to-end against an MLP baseline and a VQ-VAE model.
Success rate (\textbf{Succ.}) measures the fraction of evaluation episodes that successfully complete the tracking task. An episode is deemed successful if the average joint position error, computed over all frames, is below 0.5 m.
\textbf{MPJPE} stands for Mean Per-Joint Position Error, measured in millimeters (mm). }

\label{tab:skill_quant_results}
\begin{tabular}{l|cc|cc}
\toprule
& \multicolumn{2}{c}{\textbf{Bones [680 hr]}}
& \multicolumn{2}{c}{\textbf{AMASS [40 hr]}} \\

\textbf{Method} 
& \textbf{Succ. (\%)} $\uparrow$ 
& \textbf{MPJPE} $\downarrow$
& \textbf{Succ. (\%)} $\uparrow$ 
& \textbf{MPJPE} $\downarrow$\\
\midrule

MLP     
& \textbf{99.98} & \textbf{25.56}   
& \textbf{99.59} & \textbf{30.26}\\

VQ-VAE  
& 99.94 & 37.92 
& 99.30 & 59.28\\

\textbf{FSQ}     
& 99.98 & 34.90 
& 99.51 & 44.43 \\

\bottomrule
\end{tabular}
\end{table}

\begin{table}[t]
\centering
\caption{
Comparison between end-to-end trained FSQ and a variant (denoted as FSQ-K) in which the encoder is first trained on kinematic reference data using supervised learning and then kept frozen during subsequent policy training with reinforcement learning.
\textbf{Util.} denotes the codebook utilization rate that is averaged over grouped tokens.}
\label{tab:skill_quant_ablation}
\begin{tabular}{l|ccc}
\toprule
\textbf{Method} 
& \textbf{Succ. (\%)} $\uparrow$ 
& \textbf{MPJPE} $\downarrow$
& \textbf{Util. (\%)} $\uparrow$\\
\midrule

FSQ-K
& 99.03 & 78.26  & 76.34 \\

\textbf{FSQ}     
& \textbf{99.98} & \textbf{34.90} & \textbf{82.15} \\

\bottomrule
\end{tabular}
\end{table}
\section{Experiment}

\begin{table*}[t]
\centering
\caption{
Ablation of the grouping factor $G$ for a generative controller ($L=9$, $d=40$). \textbf{$G$} denotes the number of tokens per group and \textbf{$N_{vocab}$} the resulting vocabulary size. We report motion quality metrics including average pairwise distance (APD, m) \citep{rempe2021humor, apd2020}, average displacement error (ADE, m), and acceleration error (Accel., m/s$^2$), as well as computational cost. We find that for $G \ge 8$, the computational cost becomes prohibitive in terms of memory usage, and the resulting vocabulary size far exceeds that of modern LLMs \citep{bai2023qwen, radford2019gpt2}.
}

\begin{tabular}{ccc|ccc|ccc}
\toprule
\textbf{$G$} &
\textbf{$N_{vocab}$} &
\textbf{$N_{token}$} &
\textbf{APD (m)}~$\uparrow$ &
\textbf{ADE (m)}~$\downarrow$ &
\textbf{Accel. (m/$s^2$)}&
\textbf{$N_{param}$}&
\textbf{Mem (GB)}&
\textbf{FPS}\\
\midrule
 
8 & $4.3 \times 10^{7}$ & 5 & - & - & - & $4.4 \times 10^{10}$ & $1.7 \times 10^{2}$ & - \\
5  & 59,049 & 8 & 0.34  & 0.30 & 3.67 &$6.0 \times 10^{7}$  & 0.27 & 92.85 \\
4  & 6,561 & 10 & 0.29 & 0.29 & 3.21 &$6.7 \times 10^{5}$ & 0.03 & 115.47 \\
2  & 81  & 20  & 0.26 & 0.27 & 3.04 & $8.3 \times 10^{4}$& <0.01 & 56.43 \\
1  & 9 &  40  & 0.27 & 0.24 & 2.88 & $9 \times 10^{3}$ & <0.01 & 25.15\\

\bottomrule
\end{tabular}
\label{tab:grouping_ablation}
\end{table*}

\begin{table}[t]
    \centering
    \caption{We compare the performance of a controller fine-tuned sequentially via SFT and RL against a baseline optimized only using RL without SFT. Incorporating SFT biases the controller toward a more restricted set of skills, yielding less diverse behaviors, with decreased APD, ADE, and entropy. This reduction in diversity does not compromise task performance, as shown by comparable returns and robust performance under external perturbations.}
   
    \label{tab:task_sft}
    \begin{tabular}{l c c c c c}
        \toprule
        \textbf{Method} & \textbf{Return} &  \textbf{Pert. Ret.}  & \textbf{Entropy} & \textbf{APD} & \textbf{ADE} \\
        \midrule
        SFT & 143.36 & 125.44 & 2.57 & 0.24	& 0.15  \\
        w/o SFT &  230.42 & 176.35 & 5.08 & 0.26 & 0.27\\
        \bottomrule
    \end{tabular}
\end{table}

We evaluate our proposed framework through experiments that examine the impact of end-to-end RL training and parameter-efficient fine-tuning on downstream task performance (Section~\ref{ex:task}). Next, we evaluate the performance of different skill quantization methods (Section~\ref{ex:tracking}), and analyze the effects of different token grouping strategies on the performance of the resulting models (Section~\ref{ex:group}).

\subsection{Experimental Setup}
\label{ex:exsetup}
All experiments are conducted in a physically simulated environment implemented with Isaac Gym \citep{makoviychuk2021isaacgym}. Our training and evaluation pipeline is built on top of the ProtoMotions framework \citep{tessler2025protomotions}, which provides scalable infrastructure for large-scale motion tracking and reinforcement learning with simulated characters. For the Bones dataset \citep{BonesStudio2026}, both the FSQ-based tracking controller and the generative controller are trained using 24 NVIDIA A100 GPUs. The rest of the experiments are conducted on a single A100 GPU. Once trained, the models are capable of running on consumer-grade hardware with NVIDIA RTX 4090 GPU. Our FSQ model uses 40 discrete latent tokens with 9 quantization levels per token, resulting in an implicit codebook of size $9^{40}$.

\subsection{Skill Quantization}
\label{ex:tracking}
To evaluate the effectiveness of different quantization methods when applied to motion tracking, we compare our FSQ-based tracking controller against a VQ-VAE-based controller and an MLP baseline without a quantization layer on Bones~\citep{BonesStudio2026} and AMASS~\citep{mahmood2019amass}. All models are trained end-to-end under the same training protocol. The results are summarized in Table~\ref{tab:skill_quant_results}. The MLP baseline achieves the best tracking accuracy across the various metrics, and its strong performance can likely be attributed to the absence of a quantization bottleneck. When comparing different quantization methods, our FSQ model with reinforcement learning consistently outperforms the VQ-VAE–based approach. Our method attains higher success rates and lower MPJPE, indicating improved overall tracking performance. Furthermore, FSQ does not require any of the additional stabilization techniques commonly needed for VQ-VAE methods, such as codebook heuristics or auxiliary losses. Qualitative examples of the motions produced by our controller are shown in Figure~\ref{fig:inhouse_track}.

A key characteristic of our approach is that all components of the model, including the encoder and decoder, are trained through end-to-end RL, which enables the latent representations to be directly optimized to be amenable to physics-based control. In contrast, several prior works with discrete latent representations train VQ-VAE models via supervised distillation from pretrained tracking controllers, rather than optimizing the representations in an end-to-end manner~\citep{bae2025hybridlatent}. To evaluate the impact of end-to-end RL training, we compare our model against a variant that uses a pretrained FSQ encoder that was trained strictly on kinematic motion data with supervised learning. As shown in Table~\ref{tab:skill_quant_ablation}, the model trained with end-to-end RL consistently outperforms the variant with a pretrained kinematic encoder, demonstrating the importance of end-to-end optimization for learning representations that are well suited for physics-based control.

\begin{figure*}[h] 
    \centering
    \begin{minipage}{0.33\textwidth}
        \centering
        \includegraphics[width=\linewidth]{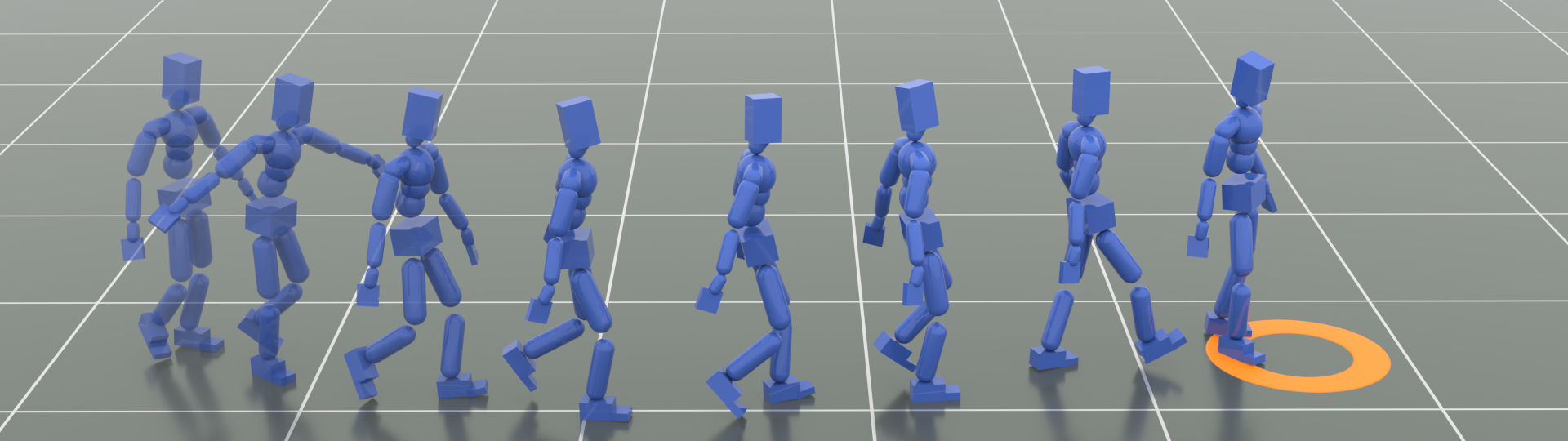}
        \small (a) CVAE episode 1
    \end{minipage}
    \hfill
    \begin{minipage}{0.33\textwidth}
        \centering
        \includegraphics[width=\linewidth]{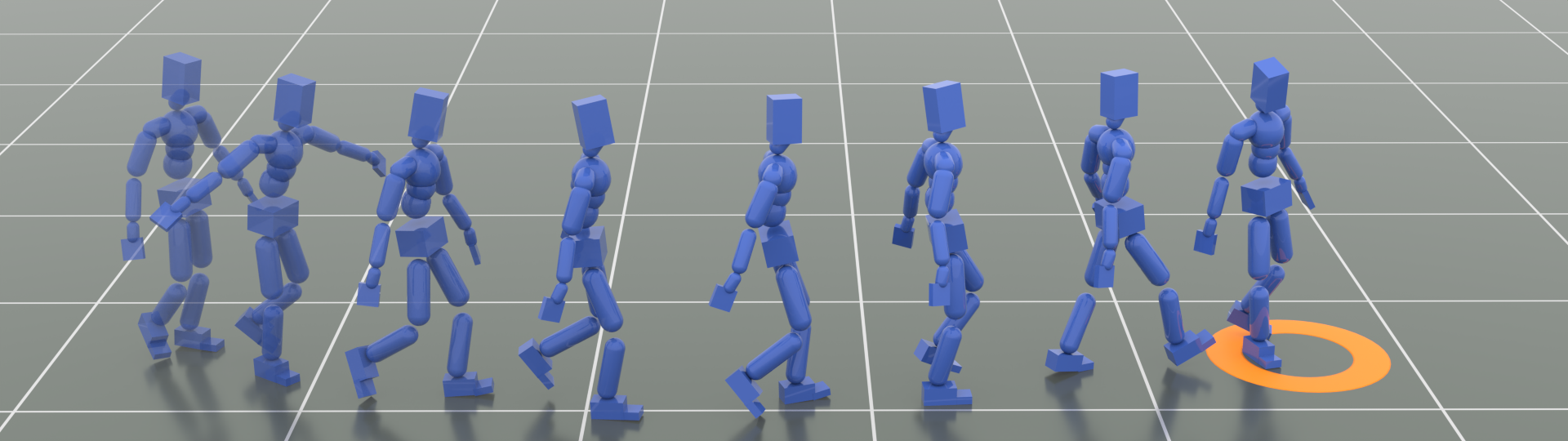}
        \small (b) CVAE episode 2
    \end{minipage}
    \hfill
    \begin{minipage}{0.33\textwidth}
        \centering
        \includegraphics[width=\linewidth]{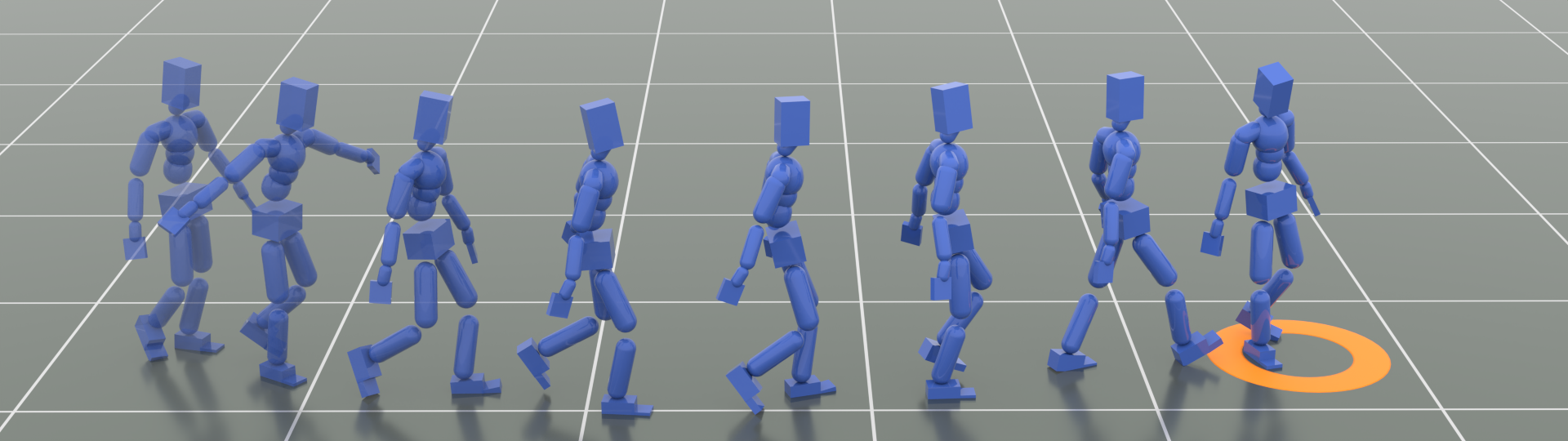}
        \small (c) CVAE episode 3
    \end{minipage}

    \begin{minipage}{0.33\textwidth}
        \centering
        \includegraphics[width=\linewidth]{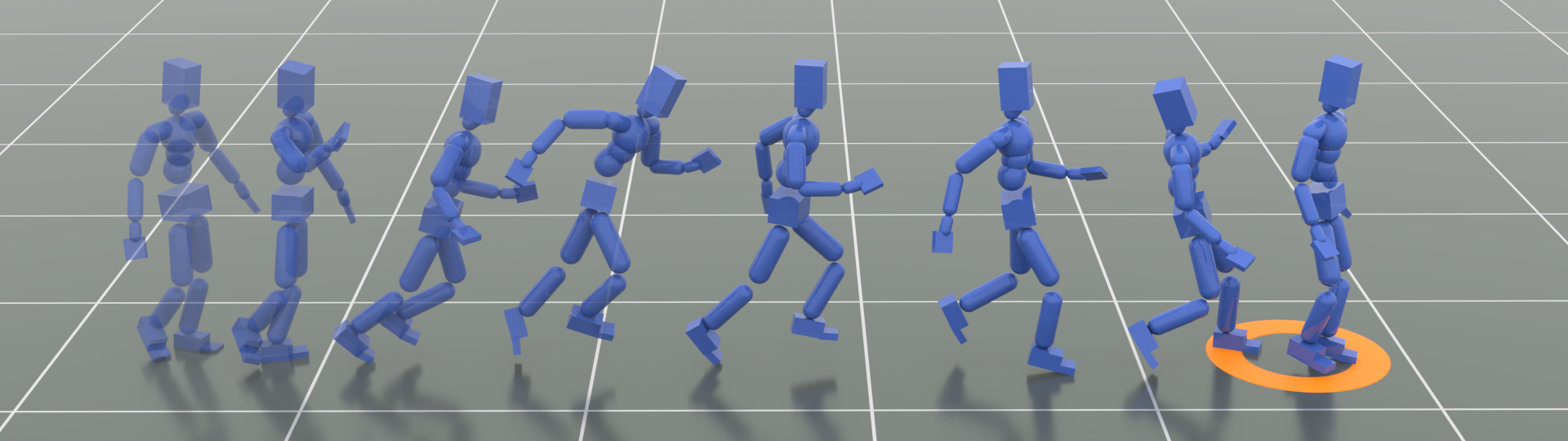}
        \small (d) GPC episode 1
    \end{minipage}
    \hfill
    \begin{minipage}{0.33\textwidth}
        \centering
        \includegraphics[width=\linewidth]{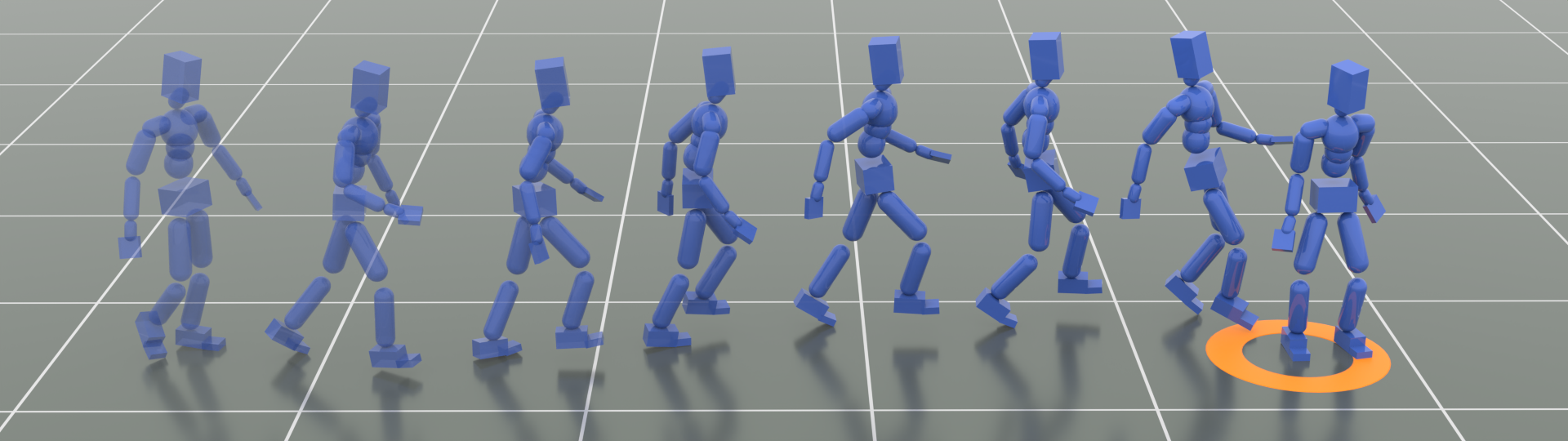}
        \small (e) GPC episode 2
    \end{minipage}
    \hfill
    \begin{minipage}{0.33\textwidth}
        \centering
        \includegraphics[width=\linewidth]{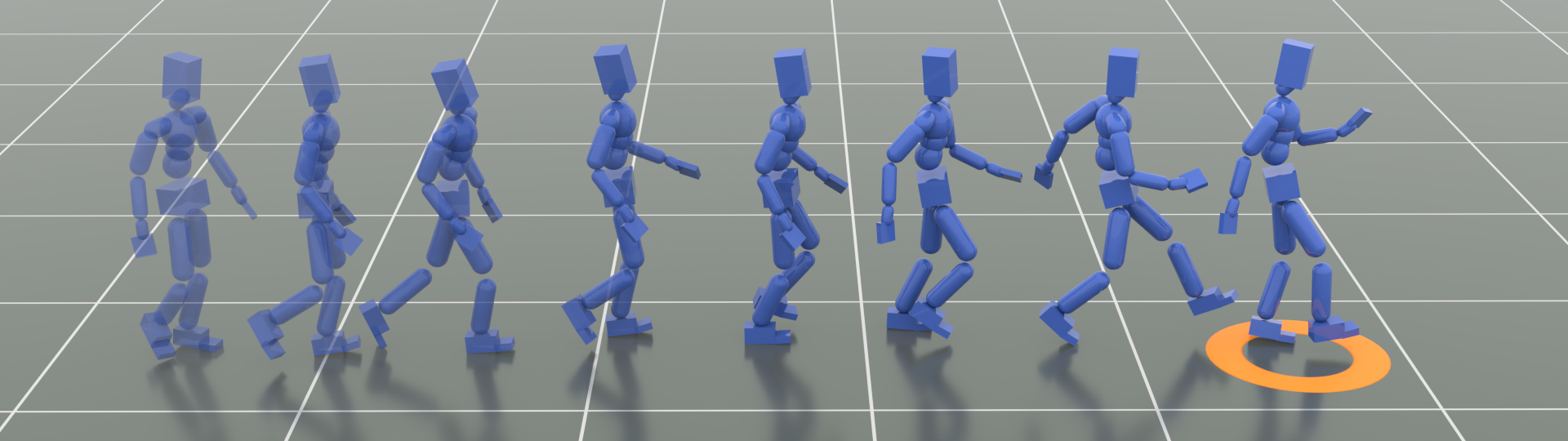}
        \small (f) GPC episode 3
    \end{minipage}

    \caption{Comparison between downstream task behavior of GPC and CVAE under identical task conditions. GPC produces more diverse behaviors than CVAE.}
    
    \label{fig:motion_comparison}
\end{figure*}

\begin{figure}[t]
    \centering
    \includegraphics[width=0.49\linewidth]{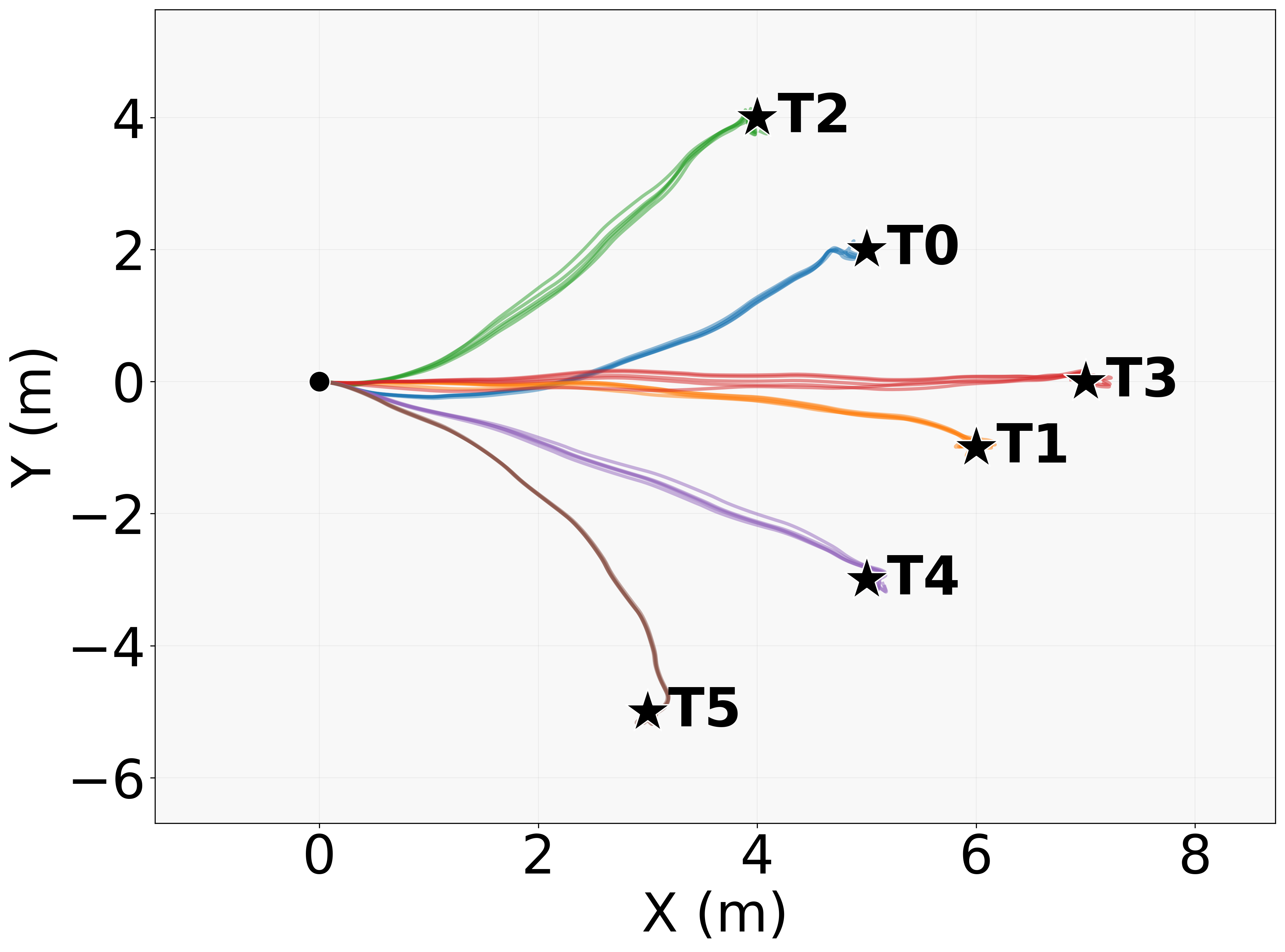}%
    \hfill
    \includegraphics[width=0.49\linewidth]{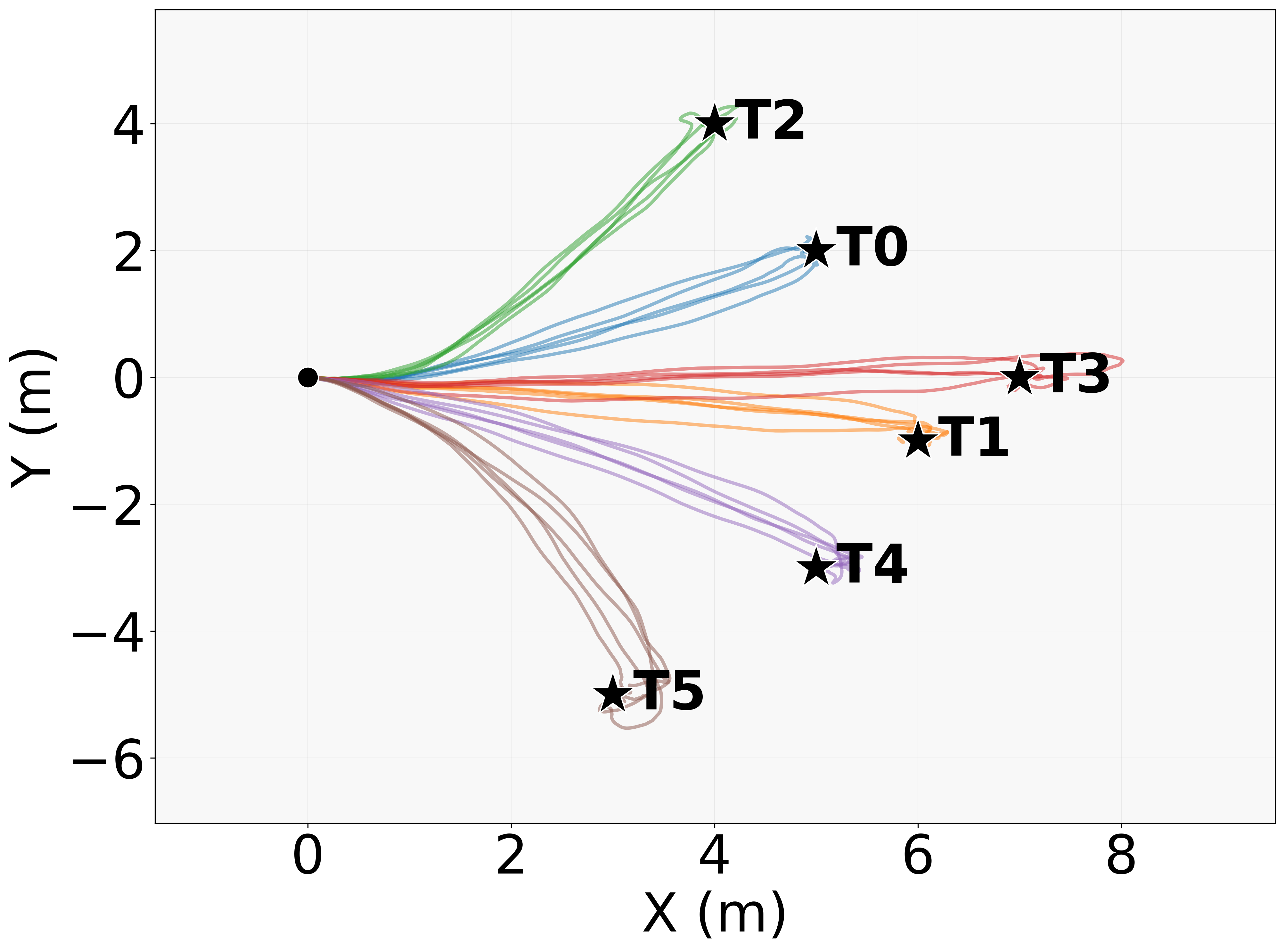}
    \caption{Trajectory comparison. The CVAE baseline (left) collapses to similar behaviors, while GPC (right) produces diverse trajectories.}
    \label{fig:trajectory_comparison}
\end{figure}

\subsection{Generative Controller}
\label{ex:gc}
By training GPC on large-scale motion datasets, the learned generative controller can produce a wide range of life-like skills and natural transitions between diverse behaviors. Figure~\ref{fig:uncond} illustrates examples of the diverse behaviors produced through unconditional sampling from GPC starting at the same initial state. GPC is able to produce a wide range of behaviors, including rolling, jumping, dancing, and acrobatics. Beyond generating diverse behaviors, the controller also exhibits emergent human-like responses to external perturbations. As shown in Fig.~\ref{fig:inhouse_force}, when the character falls down, GPC automatically transitions to a fall-recovery behavior to get back up. 

\subsection{Token Grouping}
\label{ex:group}
Token grouping is an important design choice for improving GPC's performance while reducing the inference cost associated with long context lengths. As shown in Table~\ref{tab:grouping_ablation}, we evaluate several grouping strategies that yield grouped tokens with a vocabulary size of \(L^{G}\). A grouping factor of \(G=5\) achieves the highest APD, indicating increased behavioral diversity, at the expense of reduced tracking accuracy and smoothness, as reflected by higher ADE and joint accelerations. Despite this trade-off, this setting produces the best motion quality in qualitative evaluations. This improvement may be attributed to the reduced sequence length, thereby lowering the challenges of modeling the relationship across long token sequences. Token grouping may also allow each token to model higher-level semantic information of different behaviors. A similar effect has been observed in language modeling, where moving from character-level to subword-level representations improves both efficiency and expressiveness by embedding semantic structure directly into the tokens~\citep{kim2015character,sennrich2016bpe}.

\subsection{Tasks}
\label{ex:task}
To validate the effectiveness of our pipeline, we apply our proposed adaptation approach to fine-tune a pretrained generative controller for a variety of downstream tasks. We evaluate our framework across locomotion tasks, including target reaching, trajectory following, and joystick steering, alongside scene interaction tasks that require robust behavior and the mastery of dynamic skills like jumping and crawling. Figures~\ref{fig:inhouse_task} demonstrate the successful completion of these tasks. During task execution, the adapted controller retains the emergent behaviors of the original generative controller while remaining highly robust to external perturbations.

\subsection{Supervised Fine-Tuning}
\label{ex:sft}
\begin{figure}[t]
    \centering
    \includegraphics[width=\linewidth]{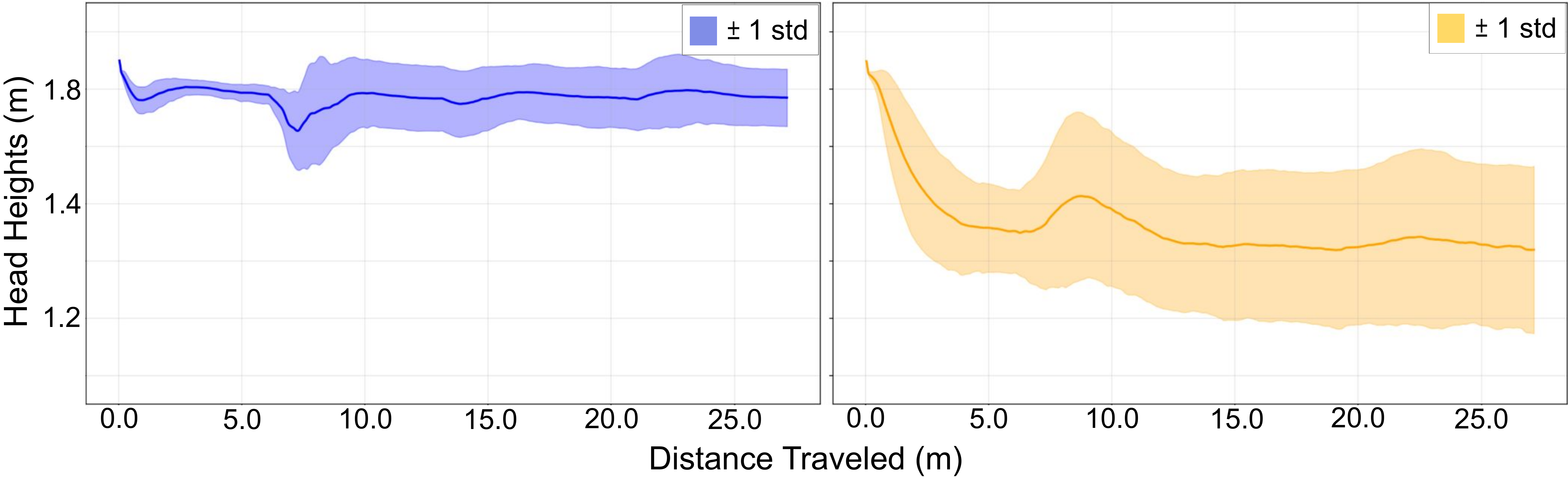}
    \caption{Trajectories of head joint height during task completion for controllers adapted (left) without SFT and (right) with SFT on a crouch motion.}
    \label{fig:sft_z}
\end{figure}

To validate the effect of SFT on downstream task adaptation, we compare task performance and behavioral characteristics of controllers adapted with SFT first against controllers finetuned purely via RL. In both settings, models are initialized with the same GPC model. The SFT variant is fine-tuned on a 20-second crouch walk motion. We evaluate each model over 512 episodes and report policy entropy and the averaged test return. Table~\ref{tab:task_sft} indicates that the SFT-trained controller leads to lower entropy in the predicted logits of the adapted generative controller, indicating more consistent stylized behaviors. This observation is qualitatively supported by Figure ~\ref{fig:sft_z}, which illustrates the trajectories of the head height. The SFT model exhibits a clear preference for maintaining a crouched posture, as indicated by the averaged head height being within the range 0.8-1.3m.

\subsection{Diverse Behaviors on Downstream Tasks}
As illustrated in Figures~\ref{fig:motion_comparison} and ~\ref{fig:trajectory_comparison}, the adapted GPC retains the stochasticity from the base generative controller when solving downstream tasks, as skill codes are sampled from the distribution over latent codes. This stochasticity allows the policy to preserve behavioral diversity, exhibiting diverse behaviors when performing identical tasks. In contrast, the CVAE-based task controller behaves deterministically, as the task controller directly selects the latent codes. Note that the slight trajectory variations observed for the CVAE baseline in Figure~\ref{fig:trajectory_comparison} (left) stem from simulator stochasticity rather than from the model. The CVAE policy lacks behavioral diversity, repeatedly executing similar motions as shown in Figure~\ref{fig:motion_comparison}.

\begin{figure}[t]
    \centering
    \includegraphics[width=0.49\linewidth]{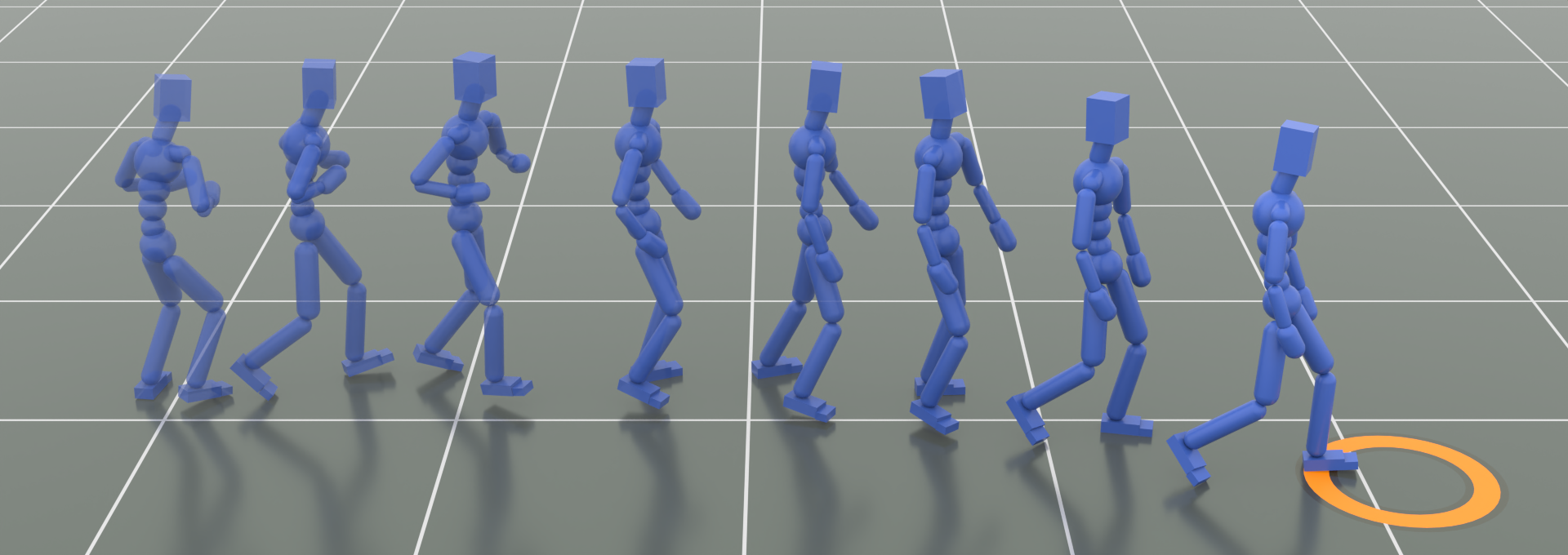}%
    \hfill
    \includegraphics[width=0.49\linewidth]{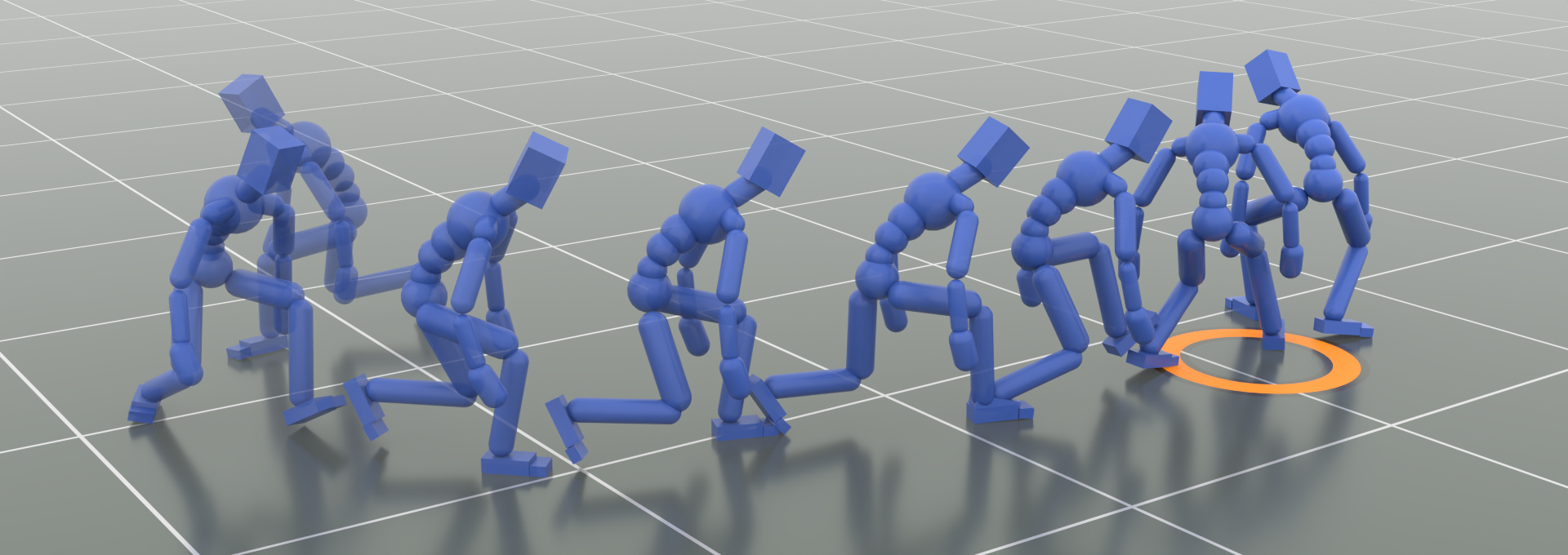}
    \caption{
    Qualitative comparison of task completion with and without SFT (left). SFT enables stylistic control by biasing the controller towards a desired behavioral style, such as crouched walking (right).
    }
    \label{fig:inhouse_sft}
\end{figure}

\section{Discussion and Limitations}
In this work, we present GPC, a generative controller trained on a large-scale motion dataset with over 600 hours of motion clips. While GPC has been effective for modeling a wide range of motor skills and adapting the learned skills to new tasks, our framework has several limitations. First, we primarily focus on locomotion-based tasks. Exploring additional multimodal extensions could enable more flexible and intuitive control, such as incorporating higher-level text-based conditioning to guide task completion. Furthermore, extending the framework to human–object interaction tasks would further broaden its potential applications. We believe these directions represent promising avenues for future work toward more general-purpose and scalable generative controllers for physics-based character animation.

\section{Acknowledgments}
We thank Yuxuan Mu, Ziyu Zhang, Dun Yang, Kaifeng Zhao, Sunmin Lee, Haotian Zhang, and Davis Rempe for their support and insightful discussions.

\bibliographystyle{ACM-Reference-Format}
\bibliography{sample-bibliography}

@String{Computing = "Computing" }

@String{Computer = "{IEEE} Computer" }

@article{peng2018deepmimic,
  title={Deepmimic: Example-guided deep reinforcement learning of physics-based character skills},
  author={Peng, Xue Bin and Abbeel, Pieter and Levine, Sergey and Van de Panne, Michiel},
  journal={ACM Transactions On Graphics (TOG)},
  volume={37},
  number={4},
  pages={1--14},
  year={2018},
  publisher={ACM New York, NY, USA}
}

@article{peng2021amp,
  title={Amp: Adversarial motion priors for stylized physics-based character control},
  author={Peng, Xue Bin and Ma, Ze and Abbeel, Pieter and Levine, Sergey and Kanazawa, Angjoo},
  journal={ACM Transactions on Graphics (ToG)},
  volume={40},
  number={4},
  pages={1--20},
  year={2021},
  publisher={ACM New York, NY, USA}
}

@article{peng2022ase,
  title={Ase: Large-scale reusable adversarial skill embeddings for physically simulated characters},
  author={Peng, Xue Bin and Guo, Yunrong and Halper, Lina and Levine, Sergey and Fidler, Sanja},
  journal={ACM Transactions On Graphics (TOG)},
  volume={41},
  number={4},
  pages={1--17},
  year={2022},
  publisher={ACM New York, NY, USA}
}

@inproceedings{mahmood2019amass,
  title={AMASS: Archive of motion capture as surface shapes},
  author={Mahmood, Naureen and Ghorbani, Nima and Troje, Nikolaus F and Pons-Moll, Gerard and Black, Michael J},
  booktitle={Proceedings of the IEEE/CVF international conference on computer vision},
  pages={5442--5451},
  year={2019}
}

@inproceedings{
	tessler2023calm,
	author={Tessler, Chen and Kasten, Yoni and Guo, Yunrong and Mannor, Shie and Chechik, Gal and Peng, Xue Bin},
	title = {CALM: Conditional Adversarial Latent Models for Directable Virtual Characters},
	year = {2023},
	isbn = {9798400701597},
	publisher = {Association for Computing Machinery},
	address = {New York, NY, USA},
	booktitle = {ACM SIGGRAPH 2023 Conference Proceedings},
	location = {Los Angeles, CA, USA},
	series = {SIGGRAPH '23}
}

@inproceedings{guo2024momask,
  title={Momask: Generative masked modeling of 3d human motions},
  author={Guo, Chuan and Mu, Yuxuan and Javed, Muhammad Gohar and Wang, Sen and Cheng, Li},
  booktitle={Proceedings of the IEEE/CVF Conference on Computer Vision and Pattern Recognition},
  pages={1900--1910},
  year={2024}
}

@inproceedings{rempe2021humor,
  title={Humor: 3d human motion model for robust pose estimation},
  author={Rempe, Davis and Birdal, Tolga and Hertzmann, Aaron and Yang, Jimei and Sridhar, Srinath and Guibas, Leonidas J},
  booktitle={Proceedings of the IEEE/CVF international conference on computer vision},
  pages={11488--11499},
  year={2021}
}

@inproceedings{luo2023perpetual,
  title={Perpetual humanoid control for real-time simulated avatars},
  author={Luo, Zhengyi and Cao, Jinkun and Kitani, Kris and Xu, Weipeng and others},
  booktitle={Proceedings of the IEEE/CVF International Conference on Computer Vision},
  pages={10895--10904},
  year={2023}
}

@article{yao2024moconvq,
  title={Moconvq: Unified physics-based motion control via scalable discrete representations},
  author={Yao, Heyuan and Song, Zhenhua and Zhou, Yuyang and Ao, Tenglong and Chen, Baoquan and Liu, Libin},
  journal={ACM Transactions on Graphics (TOG)},
  volume={43},
  number={4},
  pages={1--21},
  year={2024},
  publisher={ACM New York, NY, USA}
}

@article{tessler2024maskedmimic,
    author = {Tessler, Chen and Guo, Yunrong and Nabati, Ofir and Chechik, Gal and Peng, Xue Bin},
    title = {MaskedMimic: Unified Physics-Based Character Control Through Masked Motion Inpainting},
    year = {2024},
    journal={ACM Transactions on Graphics (TOG)},
    publisher={ACM New York, NY, USA}
}

@article{ho2020denoising,
  title={Denoising diffusion probabilistic models},
  author={Ho, Jonathan and Jain, Ajay and Abbeel, Pieter},
  journal={Advances in neural information processing systems},
  volume={33},
  pages={6840--6851},
  year={2020}
}

@article{jiang2023motiongpt,
  title={Motiongpt: Human motion as a foreign language},
  author={Jiang, Biao and Chen, Xin and Liu, Wen and Yu, Jingyi and Yu, Gang and Chen, Tao},
  journal={Advances in Neural Information Processing Systems},
  volume={36},
  pages={20067--20079},
  year={2023}
}

@article{harvey2020lafan,
  title={Robust motion in-betweening},
  author={Harvey, F{\'e}lix G and Yurick, Mike and Nowrouzezahrai, Derek and Pal, Christopher},
  journal={ACM Transactions on Graphics (TOG)},
  volume={39},
  number={4},
  pages={60--1},
  year={2020},
  publisher={ACM New York, NY, USA}
}

@misc{loshchilov2018decoupled,
      title={Decoupled Weight Decay Regularization}, 
      author={Ilya Loshchilov and Frank Hutter},
      year={2019},
      eprint={1711.05101},
      archivePrefix={arXiv},
      primaryClass={cs.LG}
}

@inbook{pytorch,
author = {Paszke, Adam and Gross, Sam and Massa, Francisco and Lerer, Adam and Bradbury, James and Chanan, Gregory and Killeen, Trevor and Lin, Zeming and Gimelshein, Natalia and Antiga, Luca and Desmaison, Alban and K\"{o}pf, Andreas and Yang, Edward and DeVito, Zach and Raison, Martin and Tejani, Alykhan and Chilamkurthy, Sasank and Steiner, Benoit and Fang, Lu and Bai, Junjie and Chintala, Soumith},
title = {PyTorch: an imperative style, high-performance deep learning library},
year = {2019},
publisher = {Curran Associates Inc.},
address = {Red Hook, NY, USA},
booktitle = {Proceedings of the 33rd International Conference on Neural Information Processing Systems},
articleno = {721},
numpages = {12}
}

@inproceedings{perez2018film,
  title={Film: Visual reasoning with a general conditioning layer},
  author={Perez, Ethan and Strub, Florian and De Vries, Harm and Dumoulin, Vincent and Courville, Aaron},
  booktitle={Proceedings of the AAAI conference on artificial intelligence},
  volume={32},
  number={1},
  year={2018}
}

@article{huang2025diffusecloc,
  title={Diffuse-CLoC: Guided Diffusion for Physics-based Character Look-ahead Control},
  author={Huang, Xiaoyu and Truong, Takara and Zhang, Yunbo and Yu, Fangzhou and Sleiman, Jean Pierre and Hodgins, Jessica and Sreenath, Koushil and Farshidian, Farbod},
  journal={ACM Transactions on Graphics (TOG)},
  volume={44},
  number={4},
  pages={1--13},
  year={2025},
  publisher={ACM},
  note={SIGGRAPH 2025}
}

@inproceedings{
        zhang2025ADD,
        author={Zhang, Ziyu and Bashkirov, Sergey and Yang, Dun and Shi, Yi and Taylor, Michael and Peng, Xue Bin},
        title = {Physics-Based Motion Imitation with Adversarial Differential Discriminators},
        year = {2025},
        booktitle = {SIGGRAPH Asia 2025 Conference Papers (SIGGRAPH Asia '25 Conference Papers)}
    }

@article{won2019learning,
  title   = {Learning body shape variation in physics-based characters},
  author  = {Won, Jungdam and Lee, Jehee},
  journal = {ACM Transactions on Graphics (TOG)},
  volume  = {38},
  number  = {6},
  pages   = {1--12},
  year    = {2019},
  doi     = {10.1145/3355089.3356499},
  publisher = {ACM}
}

@inproceedings{chentanez2018physics,
  title        = {Physics-based Motion Capture Imitation with Deep Reinforcement Learning},
  author       = {Chentanez, Nuttapong and Müller, Matthias and Macklin, Miles and Makoviychuk, Viktor and Jeschke, Stefan},
  booktitle    = {Proceedings of the 11th International Conference on Motion, Interaction and Games (MIG ’18)},
  year         = {2018},
  address      = {Limassol, Cyprus},
  doi          = {10.1145/3274247.3274506},
  publisher    = {ACM},
  pages        = {Article 4, 10 pages}
}

@inproceedings{truong2024pdp,
  title={Pdp: Physics-based character animation via diffusion policy},
  author={Truong, Takara Everest and Piseno, Michael and Xie, Zhaoming and Liu, Karen},
  booktitle={SIGGRAPH Asia 2024 Conference Papers},
  pages={1--10},
  year={2024}
}

@book{sutton1998reinforcement,
  title={Reinforcement learning: An introduction},
  author={Sutton, Richard S and Barto, Andrew G and others},
  volume={1},
  number={1},
  year={1998},
  publisher={MIT press Cambridge}
}

@article{starke2024categorical,
  title={Categorical codebook matching for embodied character controllers},
  author={Starke, Sebastian and Starke, Paul and He, Nicky and Komura, Taku and Ye, Yuting},
  journal={ACM Transactions on Graphics (TOG)},
  volume={43},
  number={4},
  pages={1--14},
  year={2024},
  publisher={ACM New York, NY, USA}
}

@article{yin2007simbicon,
  title={Simbicon: Simple biped locomotion control},
  author={Yin, KangKang and Loken, Kevin and Van de Panne, Michiel},
  journal={ACM Transactions on Graphics (TOG)},
  volume={26},
  number={3},
  pages={105--es},
  year={2007},
  publisher={ACM New York, NY, USA}
}

@article{peng2017deeploco,
  title={Deeploco: Dynamic locomotion skills using hierarchical deep reinforcement learning},
  author={Peng, Xue Bin and Berseth, Glen and Yin, KangKang and Van De Panne, Michiel},
  journal={Acm transactions on graphics (tog)},
  volume={36},
  number={4},
  pages={1--13},
  year={2017},
  publisher={ACM New York, NY, USA}
}

@article{zhu2023neural,
  title={Neural categorical priors for physics-based character control},
  author={Zhu, Qingxu and Zhang, He and Lan, Mengting and Han, Lei},
  journal={ACM Transactions on Graphics (TOG)},
  volume={42},
  number={6},
  pages={1--16},
  year={2023},
  publisher={ACM New York, NY, USA}
}

@article{tevet2024closd,
  title={Closd: Closing the loop between simulation and diffusion for multi-task character control},
  author={Tevet, Guy and Raab, Sigal and Cohan, Setareh and Reda, Daniele and Luo, Zhengyi and Peng, Xue Bin and Bermano, Amit H and van de Panne, Michiel},
  journal={arXiv preprint arXiv:2410.03441},
  year={2024}
}

@inproceedings{xu2025parc,
  title={Parc: Physics-based augmentation with reinforcement learning for character controllers},
  author={Xu, Michael and Shi, Yi and Yin, KangKang and Peng, Xue Bin},
  booktitle={Proceedings of the Special Interest Group on Computer Graphics and Interactive Techniques Conference Conference Papers},
  pages={1--11},
  year={2025}
}

@inproceedings{ye2023physdiff,
  title={PhysDiff: Physics-Guided Human Motion Diffusion Model},
  author={Ye, Yuan and Song, Jiaming and Iqbal, Umar and Vahdat, Arash and Kautz, Jan},
  booktitle={Proceedings of the IEEE/CVF International Conference on Computer Vision (ICCV)},
  year={2023}
}

@InProceedings{wang2024pacer2,
    author    = {Wang, Jingbo and Luo, Zhengyi and Yuan, Ye and Li, Yixuan and Dai, Bo},
    title     = {PACER+: On-Demand Pedestrian Animation Controller in Driving Scenarios},
    booktitle = {Proceedings of the IEEE/CVF Conference on Computer Vision and Pattern Recognition (CVPR)},
    month     = {June},
    year      = {2024},
    pages     = {718-728}
}

@inproceedings{zhou2019continuity,
  title={On the continuity of rotation representations in neural networks},
  author={Zhou, Yi and Barnes, Connelly and Lu, Jingwan and Yang, Jimei and Li, Hao},
  booktitle={Proceedings of the IEEE/CVF conference on computer vision and pattern recognition},
  pages={5745--5753},
  year={2019}
}

@inproceedings{ross2011dagger,
  title     = {A Reduction of Imitation Learning and Structured Prediction to No-Regret Online Learning},
  author    = {Ross, St{\'e}phane and Gordon, Geoffrey and Bagnell, J. Andrew},
  booktitle = {Proceedings of the 14th International Conference on Artificial Intelligence and Statistics (AISTATS)},
  year      = {2011}
}

@inproceedings{sennrich2016bpe,
    title = "Neural Machine Translation of Rare Words with Subword Units",
    author = "Sennrich, Rico and Haddow, Barry  and Birch, Alexandra",
    booktitle = "Proceedings of the 54th Annual Meeting of the Association for Computational Linguistics (Volume 1: Long Papers)",
    month = aug,
    year = "2016",
    address = "Berlin, Germany",
    publisher = "Association for Computational Linguistics",
    doi = "10.18653/v1/P16-1162",
    pages = "1715--1725"
}

@article{schulman2017proximal,
  title={Proximal policy optimization algorithms},
  author={Schulman, John and Wolski, Filip and Dhariwal, Prafulla and Radford, Alec and Klimov, Oleg},
  journal={arXiv preprint arXiv:1707.06347},
  year={2017}
}

@article{
    won2022physvae,
    author = {Won, Jungdam and Gopinath, Deepak and Hodgins, Jessica},
    title = {Physics-based Character Controllers Using Conditional VAEs},
    year = {2022},
    issue_date = {Aug 2022},
    volume = {41},
    number = {4},
    journal = {ACM Trans. Graph.},
    articleno = {96},
}

@misc{mentzer2023fsq,
      title={Finite Scalar Quantization: VQ-VAE Made Simple}, 
      author={Fabian Mentzer and David Minnen and Eirikur Agustsson and Michael Tschannen},
      year={2023},
      eprint={2309.15505},
      archivePrefix={arXiv},
      primaryClass={cs.CV},
}

@article{dou2022case,
  title={C·ASE: Learning Conditional Adversarial Skill Embeddings for Physics-based Characters},
  author={Zhiyang Dou and Xuelin Chen and Qingnan Fan and Taku Komura and Wenping Wang},
  eprint={2309.11351},
  archivePrefix={arXiv},
  year={2023}
}

@misc{oord2018vqvae,
      title={Neural Discrete Representation Learning}, 
      author={Aaron van den Oord and Oriol Vinyals and Koray Kavukcuoglu},
      year={2018},
      eprint={1711.00937},
      archivePrefix={arXiv},
      primaryClass={cs.LG},
}

@misc{goodfellow2014gan,
      title={Generative Adversarial Networks}, 
      author={Ian J. Goodfellow and Jean Pouget-Abadie and Mehdi Mirza and Bing Xu and David Warde-Farley and Sherjil Ozair and Aaron Courville and Yoshua Bengio},
      year={2014},
      eprint={1406.2661},
      archivePrefix={arXiv},
      primaryClass={stat.ML},
}

@misc{kingma2022vae,
      title={Auto-Encoding Variational Bayes}, 
      author={Diederik P Kingma and Max Welling},
      year={2022},
      eprint={1312.6114},
      archivePrefix={arXiv},
      primaryClass={stat.ML},
}

@inproceedings{coros2008stepping_walking,
author = {Coros, Stelian and Beaudoin, Philippe and Yin, Kang Kang and van de Panne, Michiel},
title = {Synthesis of constrained walking skills},
year = {2008},
isbn = {9781450318310},
publisher = {Association for Computing Machinery},
address = {New York, NY, USA},
booktitle = {ACM SIGGRAPH Asia 2008 Papers},
articleno = {113},
numpages = {9},
location = {Singapore},
series = {SIGGRAPH Asia '08}
}

@article{ye2010abstract,
author = {Ye, Yuting and Liu, C. Karen},
title = {Optimal feedback control for character animation using an abstract model},
year = {2010},
issue_date = {July 2010},
publisher = {Association for Computing Machinery},
address = {New York, NY, USA},
volume = {29},
number = {4},
issn = {0730-0301},
journal = {ACM Trans. Graph.},
month = jul,
articleno = {74},
numpages = {9}
}

@inproceedings{hodgins1995athletes,
author = {Hodgins, Jessica K. and Wooten, Wayne L. and Brogan, David C. and O'Brien, James F.},
title = {Animating human athletics},
year = {1995},
isbn = {0897917014},
publisher = {Association for Computing Machinery},
address = {New York, NY, USA},
booktitle = {Proceedings of the 22nd Annual Conference on Computer Graphics and Interactive Techniques},
pages = {71–78},
numpages = {8},
series = {SIGGRAPH '95}
}

@article{dasilva2008shorthorizon,
  title={Simulation of Human Motion Data Using Short-Horizon Model-Predictive Control},
  author={da Silva, Marco and Abe, Yeuhi and Popovi{\'c}, Jovan},
  journal={Computer Graphics Forum},
  volume={27},
  number={2},
  pages={371--380},
  year={2008},
}

@misc{baevski2020vqwav,
      title={vq-wav2vec: Self-Supervised Learning of Discrete Speech Representations}, 
      author={Alexei Baevski and Steffen Schneider and Michael Auli},
      year={2020},
      eprint={1910.05453},
      archivePrefix={arXiv},
      primaryClass={cs.CL}
}

@misc{esser2021vqvgan,
      title={Taming Transformers for High-Resolution Image Synthesis}, 
      author={Patrick Esser and Robin Rombach and Björn Ommer},
      year={2021},
      eprint={2012.09841},
      archivePrefix={arXiv},
      primaryClass={cs.CV}
}

@misc{merel2019neuralprob,
      title={Neural probabilistic motor primitives for humanoid control}, 
      author={Josh Merel and Leonard Hasenclever and Alexandre Galashov and Arun Ahuja and Vu Pham and Greg Wayne and Yee Whye Teh and Nicolas Heess},
      year={2019},
      eprint={1811.11711},
      archivePrefix={arXiv},
      primaryClass={cs.LG}
}

@article{bae2025hybridlatent,
  title={Versatile Physics-based Character Control with Hybrid Latent Representation},
  author={Bae, Jinseok and Won, Jungdam and Lim, Donggeun and Hwang, Inwoo and Kim, Young Min},
  journal={Computer Graphics Forum},
  year={2025},
  publisher={The Eurographics Association and John Wiley \& Sons, Ltd},
  doi={10.1111/cgf.70018}
}

@inproceedings{plt2024bae,
author = {Bae, Jinseok and Lee, Younghwan and Lim, Donggeun and Kim, Young Min},
title = {PLT: Part-Wise Latent Tokens as Adaptable Motion Priors for Physically Simulated Characters},
year = {2025},
publisher = {Association for Computing Machinery},
address = {New York, NY, USA},
doi = {10.1145/3721238.3730637},
booktitle = {Proceedings of the Special Interest Group on Computer Graphics and Interactive Techniques Conference Conference Papers},
articleno = {132},
numpages = {10},
series = {SIGGRAPH Conference Papers '25}
}

@misc{hu2021lora,
      title={LoRA: Low-Rank Adaptation of Large Language Models}, 
      author={Edward J. Hu and Yelong Shen and Phillip Wallis and Zeyuan Allen-Zhu and Yuanzhi Li and Shean Wang and Lu Wang and Weizhu Chen},
      year={2021},
      eprint={2106.09685},
      archivePrefix={arXiv},
      primaryClass={cs.CL}
}

@misc{liu2024dora,
      title={DoRA: Weight-Decomposed Low-Rank Adaptation}, 
      author={Shih-Yang Liu and Chien-Yi Wang and Hongxu Yin and Pavlo Molchanov and Yu-Chiang Frank Wang and Kwang-Ting Cheng and Min-Hung Chen},
      year={2024},
      eprint={2402.09353},
      archivePrefix={arXiv},
      primaryClass={cs.CL}
}

@misc{zhang2023contrlnet,
  title={Adding Conditional Control to Text-to-Image Diffusion Models}, 
  author={Lvmin Zhang and Anyi Rao and Maneesh Agrawala},
  booktitle={IEEE International Conference on Computer Vision (ICCV)},
  year={2023},
}

@inproceedings{houlsby2019parameter,
  title     = {Parameter-Efficient Transfer Learning for NLP},
  author    = {Houlsby, Neil and Giurgiu, Andrei and Jastrzebski, Stanislaw and Morrone, Bruna and de Laroussilhe, Quentin and Gesmundo, Andrea and Attariyan, Mona and Gelly, Sylvain},
  booktitle = {Proceedings of the 36th International Conference on Machine Learning (ICML)},
  year      = {2019}
}

@misc{pfeiffer2021adapterfusion,
      title={AdapterFusion: Non-Destructive Task Composition for Transfer Learning}, 
      author={Jonas Pfeiffer and Aishwarya Kamath and Andreas Rücklé and Kyunghyun Cho and Iryna Gurevych},
      year={2021},
      eprint={2005.00247},
      archivePrefix={arXiv},
      primaryClass={cs.CL}
}

@inproceedings{pan2025tokenhsi,
      title={Tokenhsi: Unified synthesis of physical human-scene interactions through task tokenization},
      author={Pan, Liang and Yang, Zeshi and Dou, Zhiyang and Wang, Wenjia and Huang, Buzhen and Dai, Bo and Komura, Taku and Wang, Jingbo},
      booktitle={Proceedings of the Computer Vision and Pattern Recognition Conference},
      pages={5379--5391},
      year={2025}
}

@misc{luo2024pulse,
      title={Universal Humanoid Motion Representations for Physics-Based Control}, 
      author={Zhengyi Luo and Jinkun Cao and Josh Merel and Alexander Winkler and Jing Huang and Kris Kitani and Weipeng Xu},
      year={2024},
      eprint={2310.04582},
      archivePrefix={arXiv},
      primaryClass={cs.CV}
}

@misc{dettmers2023qlora,
      title={QLoRA: Efficient Finetuning of Quantized LLMs}, 
      author={Tim Dettmers and Artidoro Pagnoni and Ari Holtzman and Luke Zettlemoyer},
      year={2023},
      eprint={2305.14314},
      archivePrefix={arXiv},
      primaryClass={cs.LG}
}

@inproceedings{
   zhang2023adalora,
   title={Adaptive Budget Allocation for Parameter-Efficient Fine-Tuning },
   author={Qingru Zhang and Minshuo Chen and Alexander Bukharin and Pengcheng He and Yu Cheng and Weizhu Chen and Tuo Zhao},
   booktitle={The Eleventh International Conference on Learning Representations },
   year={2023}
}

@article{radford2019gpt2,
  title={Language Models are Unsupervised Multitask Learners},
  author={Radford, Alec and Wu, Jeff and Child, Rewon and Luan, David and Amodei, Dario and Sutskever, Ilya},
  year={2019}
}

@misc{radford2018gpt,
  title  = {Improving Language Understanding by Generative Pre-Training},
  author = {Alec Radford and Karthik Narasimhan and Tim Salimans and Ilya Sutskever},
  year   = {2018},
  note   = {OpenAI technical report}
}

@article{bengio2013ste,
  title   = {Estimating or Propagating Gradients Through Stochastic Neurons for Conditional Computation},
  author  = {Yoshua Bengio and Nicolas L{\'e}onard and Aaron Courville},
  journal = {arXiv preprint arXiv:1308.3432},
  year    = {2013}
}

@misc{jordan2024muon,
  author       = {Keller Jordan and Yuchen Jin and Vlado Boza and Jiacheng You and
                  Franz Cesista and Laker Newhouse and Jeremy Bernstein},
  title        = {Muon: An optimizer for hidden layers in neural networks},
  year         = {2024},
  url          = {https://kellerjordan.github.io/posts/muon/}
}

@misc{liu2025muonscalablellmtraining,
      title={Muon is Scalable for LLM Training}, 
      author={Jingyuan Liu and Jianlin Su and Xingcheng Yao and Zhejun Jiang and Guokun Lai and Yulun Du and Yidao Qin and Weixin Xu and Enzhe Lu and Junjie Yan and Yanru Chen and Huabin Zheng and Yibo Liu and Shaowei Liu and Bohong Yin and Weiran He and Han Zhu and Yuzhi Wang and Jianzhou Wang and Mengnan Dong and Zheng Zhang and Yongsheng Kang and Hao Zhang and Xinran Xu and Yutao Zhang and Yuxin Wu and Xinyu Zhou and Zhilin Yang},
      year={2025},
      eprint={2502.16982},
      archivePrefix={arXiv},
      primaryClass={cs.LG}
}

@misc{holtzman2020neucls,
      title={The Curious Case of Neural Text Degeneration}, 
      author={Ari Holtzman and Jan Buys and Li Du and Maxwell Forbes and Yejin Choi},
      year={2020},
      eprint={1904.09751},
      archivePrefix={arXiv},
      primaryClass={cs.CL}
}

@misc{makoviychuk2021isaacgym,
      title={Isaac Gym: High Performance GPU-Based Physics Simulation For Robot Learning}, 
      author={Viktor Makoviychuk and Lukasz Wawrzyniak and Yunrong Guo and Michelle Lu and Kier Storey and Miles Macklin and David Hoeller and Nikita Rudin and Arthur Allshire and Ankur Handa and Gavriel State},
      year={2021},
      eprint={2108.10470},
      archivePrefix={arXiv},
      primaryClass={cs.RO},
}

@misc{tessler2025protomotions,
  title = {ProtoMotions3: An Open-source Framework for Humanoid Simulation and Control},
  author = {Tessler, Chen and Jiang, Yifeng and Peng, Xue Bin and Coumans, Erwin and Shi, Yi and Zhang, Haotian and Rempe, Davis and Chechik†, Gal and Fidler, Sanja},
  year = {2025},
  publisher = {GitHub},
  journal = {GitHub repository},
  howpublished = {\url{https://github.com/NVLabs/ProtoMotions/}},
}

@misc{kim2015character,
      title={Character-Aware Neural Language Models}, 
      author={Yoon Kim and Yacine Jernite and David Sontag and Alexander M. Rush},
      year={2015},
      eprint={1508.06615},
      archivePrefix={arXiv},
      primaryClass={cs.CL}
}

@misc{bai2023qwen,
      title={Qwen Technical Report}, 
      author={Jinze Bai and Shuai Bai and Yunfei Chu and Zeyu Cui and Kai Dang and Xiaodong Deng and Yang Fan and Wenbin Ge and Yu Han and Fei Huang and Binyuan Hui and Luo Ji and Mei Li and Junyang Lin and Runji Lin and Dayiheng Liu and Gao Liu and Chengqiang Lu and Keming Lu and Jianxin Ma and Rui Men and Xingzhang Ren and Xuancheng Ren and Chuanqi Tan and Sinan Tan and Jianhong Tu and Peng Wang and Shijie Wang and Wei Wang and Shengguang Wu and Benfeng Xu and Jin Xu and An Yang and Hao Yang and Jian Yang and Shusheng Yang and Yang Yao and Bowen Yu and Hongyi Yuan and Zheng Yuan and Jianwei Zhang and Xingxuan Zhang and Yichang Zhang and Zhenru Zhang and Chang Zhou and Jingren Zhou and Xiaohuan Zhou and Tianhang Zhu},
      year={2023},
      eprint={2309.16609},
      archivePrefix={arXiv},
      primaryClass={cs.CL},
      url={https://arxiv.org/abs/2309.16609}, 
}

@misc{lipman2023fm,
      title={Flow Matching for Generative Modeling}, 
      author={Yaron Lipman and Ricky T. Q. Chen and Heli Ben-Hamu and Maximilian Nickel and Matt Le},
      year={2023},
      eprint={2210.02747},
      archivePrefix={arXiv},
      primaryClass={cs.LG},
      url={https://arxiv.org/abs/2210.02747}, 
}

@misc{muller2020labelsmoothing,
      title={When Does Label Smoothing Help?}, 
      author={Rafael Müller and Simon Kornblith and Geoffrey Hinton},
      year={2020},
      eprint={1906.02629},
      archivePrefix={arXiv},
      primaryClass={cs.LG},
      url={https://arxiv.org/abs/1906.02629}, 
}

@misc{hendrycks2023gaussianerrorlinearunits,
      title={Gaussian Error Linear Units (GELUs)}, 
      author={Dan Hendrycks and Kevin Gimpel},
      year={2023},
      eprint={1606.08415},
      archivePrefix={arXiv},
      primaryClass={cs.LG},
      url={https://arxiv.org/abs/1606.08415}, 
}

@inproceedings{muennighoff-etal-2023-mteb,
    title = "{MTEB}: Massive Text Embedding Benchmark",
    author = "Muennighoff, Niklas  and
      Tazi, Nouamane  and
      Magne, Loic  and
      Reimers, Nils",
    editor = "Vlachos, Andreas  and
      Augenstein, Isabelle",
    booktitle = "Proceedings of the 17th Conference of the European Chapter of the Association for Computational Linguistics",
    month = may,
    year = "2023",
    address = "Dubrovnik, Croatia",
    publisher = "Association for Computational Linguistics",
    url = "https://aclanthology.org/2023.eacl-main.148/",
    doi = "10.18653/v1/2023.eacl-main.148",
    pages = "2014--2037",

}

@article{TSNE,
  author  = {Laurens van der Maaten and Geoffrey Hinton},
  title   = {Visualizing Data using t-SNE},
  journal = {Journal of Machine Learning Research},
  year    = {2008},
  volume  = {9},
  number  = {86},
  pages   = {2579--2605},
  url     = {http://jmlr.org/papers/v9/vandermaaten08a.html}
}

@misc{reimers2019sentencebert,
      title={Sentence-BERT: Sentence Embeddings using Siamese BERT-Networks}, 
      author={Nils Reimers and Iryna Gurevych},
      year={2019},
      eprint={1908.10084},
      archivePrefix={arXiv},
      primaryClass={cs.CL},
      url={https://arxiv.org/abs/1908.10084}, 
}

@misc{janner2021trajformer,
      title={Offline Reinforcement Learning as One Big Sequence Modeling Problem}, 
      author={Michael Janner and Qiyang Li and Sergey Levine},
      year={2021},
      eprint={2106.02039},
      archivePrefix={arXiv},
      primaryClass={cs.LG},
      url={https://arxiv.org/abs/2106.02039}, 
}

@misc{chen2021decisiontransformer,
      title={Decision Transformer: Reinforcement Learning via Sequence Modeling}, 
      author={Lili Chen and Kevin Lu and Aravind Rajeswaran and Kimin Lee and Aditya Grover and Michael Laskin and Pieter Abbeel and Aravind Srinivas and Igor Mordatch},
      year={2021},
      eprint={2106.01345},
      archivePrefix={arXiv},
      primaryClass={cs.LG},
      url={https://arxiv.org/abs/2106.01345}, 
}

@article{Hogan01012009,
author = {Neville Hogan and Dagmar Sternad},
title = {Sensitivity of Smoothness Measures to Movement Duration, Amplitude, and Arrests},
journal = {Journal of Motor Behavior},
volume = {41},
number = {6},
pages = {529--534},
year = {2009},
publisher = {Routledge},
doi = {10.3200/35-09-004-RC}
}

@INPROCEEDINGS{apd2020,
  author={Aliakbarian, Sadegh and Sadat Saleh, Fatemeh and Salzmann, Mathieu and Petersson, Lars and Gould, Stephen},
  booktitle={2020 IEEE/CVF Conference on Computer Vision and Pattern Recognition (CVPR)}, 
  title={A Stochastic Conditioning Scheme for Diverse Human Motion Prediction}, 
  year={2020},
  volume={},
  number={},
  pages={5222-5231},
  }

@misc{BonesStudio2026,
  author       = {{Bones Studio}},
  title        = {{BONES-SEED: Skeletal Everyday Embodied Dataset}},
  howpublished = {\url{https://bones.studio/datasets}},
  year         = {2026}
}
\clearpage

\appendix
\section*{Appendix}

\begin{figure}[h]
    \centering
    \includegraphics[width=\linewidth]{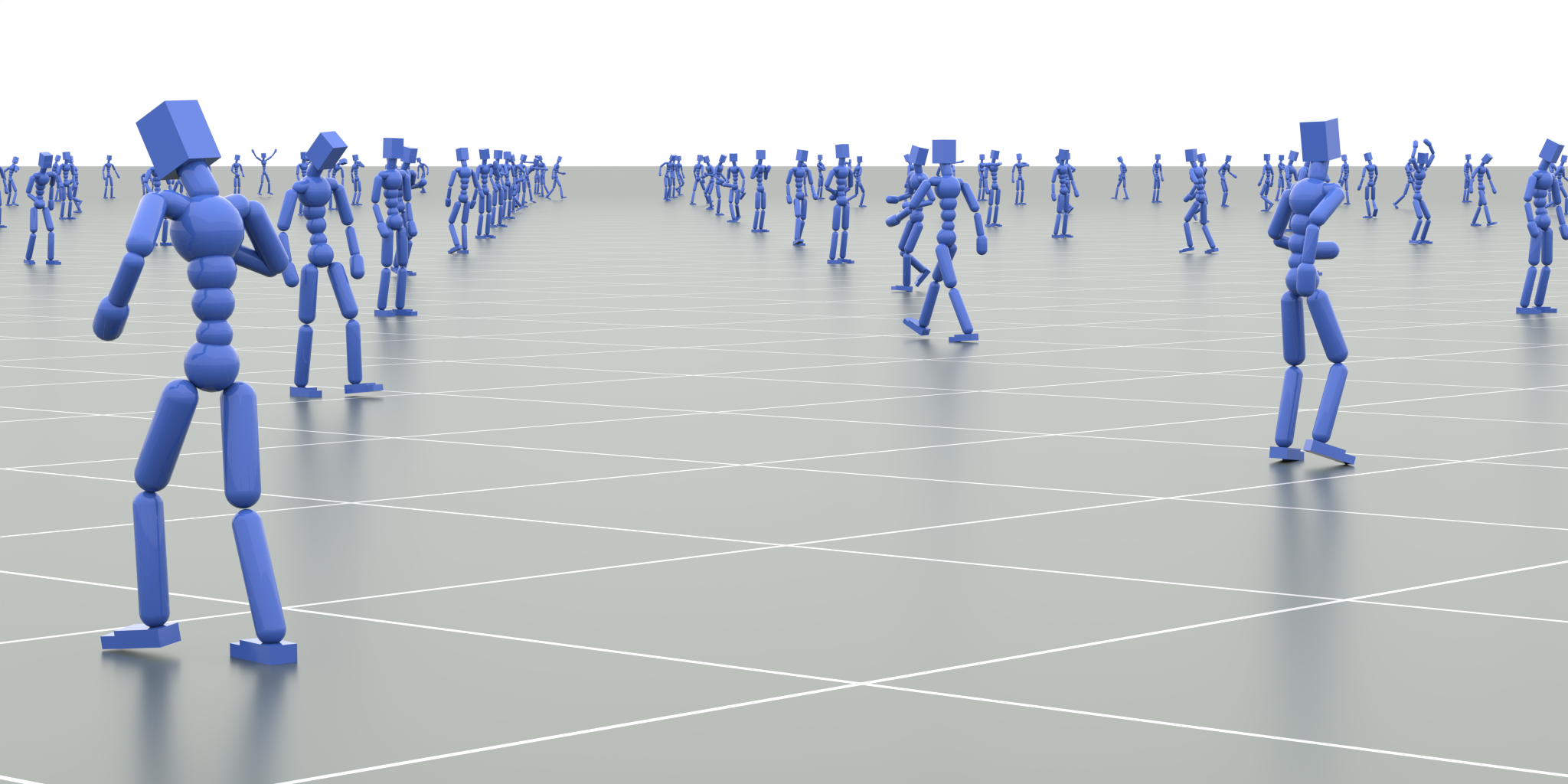}
    
    \caption{Sampled motions from Bones.}
    \label{fig:dataset}
\end{figure}

\section*{A. Dataset and Simulated Humanoid Configurations }
\label{ap:dataset}
We now introduce the datasets used in our experiments. 

\textbf{AMASS} is an aggregation of multiple motion capture datasets, covering a broad spectrum of human behaviors~\citep{mahmood2019amass}. We follow the dataset segmentation and humanoid configuration from prior studies ~\citep{luo2023perpetual,tessler2024maskedmimic}. The dataset comprises approximately 11{,}300 motion clips with a total duration of 40 hours. Owing to its moderate scale, we use it for our ablation experiments. 

\begin{figure}[t]
    \centering
    \begin{minipage}[t]{0.50\columnwidth}
        \centering
        \includegraphics[width=\linewidth]{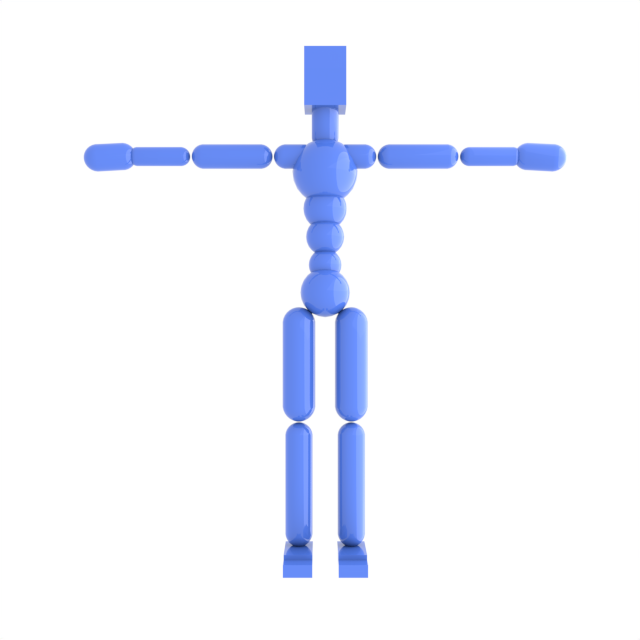}
    \end{minipage}\hfill
    \begin{minipage}[t]{0.25\columnwidth}
        \centering
        \includegraphics[width=\linewidth]{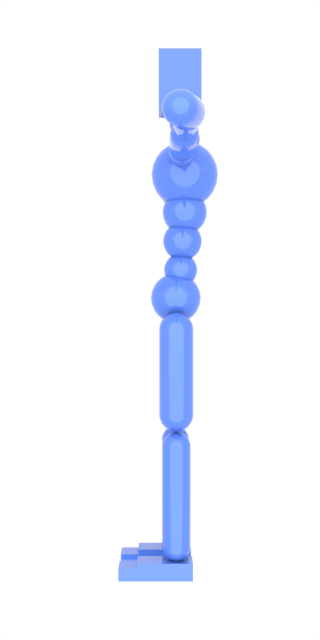}
    \end{minipage}\hfill
    \begin{minipage}[t]{0.25\columnwidth}
        \centering
        \includegraphics[width=\linewidth]{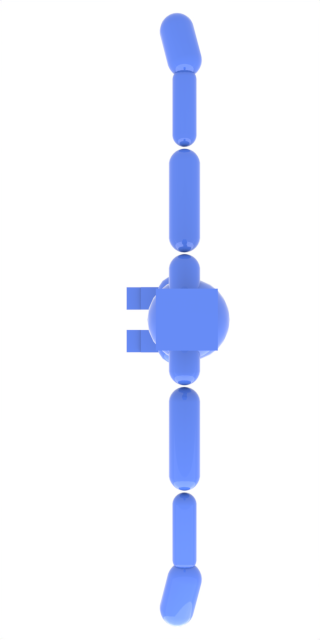}
    \end{minipage}

    \caption{The simulated character used in experiments conducted on Bones.}
    \label{fig:three_one_column}
\end{figure}

\textbf{Bones} is a large-scale motion capture dataset comprising approximately 343{,}000 clips totaling over 680 hours in length \citep{BonesStudio2026}. The collection includes a broad repertoire of daily motions from stylized walking and in-place motions to dynamic acrobatics (Figure~\ref{fig:dataset}). These sequences feature a mean length of 200 frames and a median duration of 160 frames. Our experiments on Bones utilize a customized humanoid character modeled with 23 rigid bodies and 66 actuated degrees of freedom (DoFs). Each non-root joint is modeled as a 3-DoF rotational joint, parameterized by three orthogonal hinge axes, providing full rotational freedom at each articulation. The humanoid has a standing height of 1.85 m and an arm span of 1.75 m. A visualization of the humanoid is shown in Figure~\ref{fig:three_one_column}.

\textbf{LAFAN1} (Ubisoft La Forge Animation Dataset) is a high-quality dataset comprising motion capture sequences with a duration of 4.6 hours~\citep{harvey2020lafan}. Captured from five distinct subjects, the dataset encompasses 77 sequences across 15 categories, including standard locomotion, stylized dynamic maneuvers, and recovery from falls.

\textbf{Beyond} is a motion dataset first introduced in PARC~\citep{xu2025parc}, providing parkour-style locomotions. We use 14 unique motion capture clips, with a total length of 18,600 frames and 10.3 minutes, covering diverse athletic and dynamic locomotion behaviors, including crawling and long strides, parkour rolls and hops, sharp-turn running with obstacle avoidance, and transitional actions such as getting up from the ground.

\section*{B. Skill Quantization}
\subsection*{B.1. Observation}
The observation of the FSQ-based motion tracking controller is composed of the character state $\mathbf{s}_t$ and a window of reference motion states $\hat{\mathbf{s}}_{t:t+h}$, where $h$ stands for set of target frame indices. Following prior work~\citep{tessler2024maskedmimic, peng2018deepmimic}, the character state $\mathbf{s}_t$ consists of proprioceptive information: joint positions relative to the root, root height relative to the terrain, joint orientations in the 6D rotation representation~\citep{zhou2019continuity}, and joint linear and angular velocities. All features are expressed in a canonicalized local coordinate frame aligned with the root joint to promote translational and rotational invariance. The reference motion states use the same feature representation. We use a fixed window of target motions consisting of the next $(1,2,5,7,12, 18, 25)$ frames from the reference motion. This dilated window design enables the representation to capture longer information while remaining compact and easy to model during training. Our framework decouples skill representation learning from action generation across the encoder and decoder. The encoder takes only the reference motion states $\hat{\mathbf{s}}_{t:t+h}$ as input and produces a discrete skill latent, while the decoder conditions on the current character state $\mathbf{s}_t$ and the latent to generate actions. This separation prevents a potential failure mode in which the encoder bypasses the decoder, and discourages the decoder from ignoring the character state and over-relying on the latent code produced by the encoder. As a result, the discrete latent space is encouraged to capture state-independent, reusable motion skills that can be effectively leveraged for training the generative controller.

\subsection*{B.2. Reward}
The tracking reward $r$ in our framework is designed to encourage the simulated character to reproduce the kinematic reference motions from the dataset.
\begin{equation}
r
= w_{\mathrm{gp}}\, r_{\mathrm{gp}}
+ w_{\mathrm{gr}}\, r_{\mathrm{gr}}
+ w_{\mathrm{rh}}\, r_{\mathrm{rh}}
+ w_{\mathrm{jv}}\, r_{\mathrm{jv}}
+ w_{\mathrm{jav}}\, r_{\mathrm{jav}}
\end{equation}
$r_{(\cdot)}$ denote the individual reward terms and $w_{(\cdot)}$ their corresponding weights. Each term measures the discrepancy between the reference motion and the simulated character in terms of global joint positions (gp), global joint rotations (gr), root height (rh), joint velocities (jv), and joint angular velocities (jav).

\subsection*{B.3. Optimizer}
When training FSQ tracking controller, we optimize the actor with Muon~\citep{jordan2024muon}, and train the critic with AdamW ~\citep{loshchilov2018decoupled}. Muon is a geometry-aware optimizer tailored to matrix-shaped parameters, which can greatly speed up the training. We follow a procedure where it is only applied to the hidden weight matrices of the encoder-decoder, an auxiliary AdamW optimizer is used for low-dimensional and interface parameters (e.g., biases, the encoder and decoder boundary linear layers), which improves stability while retaining Muon's favorable scaling behavior on the internal representation-learning weights \citep{liu2025muonscalablellmtraining}.

\begin{table}[t!]
\centering
\caption{Composite Reward Components for our tracking controller}
\label{tab:reward_function}
\begin{tabular}{lcc}
\toprule
\textbf{Name} & \textbf{Weight} $w_k$ & \textbf{Coeff.} $c_k$ \\ 
\midrule
$r_{\text{gp}}$ & $0.5$ & $-100$ \\
$r_{\text{gr}}$ & $0.3$ & $-5$  \\ 
$r_{\text{jv}}$ & $0.1$ & $-0.5$  \\ 
$r_{\text{jav}}$ & $0.1$ & $-0.1$ \\
$r_{\text{rh}}$ & $0.2$ & $-100$ \\
\bottomrule
\end{tabular}
\end{table}

\begin{table}[t!]
\centering
\caption{Hyperparameter configurations of Skill Quantization.}
\label{tab:hyperparameters}
\begin{tabular}{lll}
\toprule
\textbf{Encoder} & Layer & $1024, 1024, 1024, 512, 256$ \\
& Activation & ReLU \\
& Output dim & $40$ \\
\midrule
\textbf{Decoder} & Layer & $1024, 1024, 1024, 512, 256$ \\
& Activation & ReLU \\
& Output dim & $N_{DoF}$ \\
\midrule
\textbf{Critic} & Layer & $1024, 1024, 1024, 1024$ \\
& Activation & ReLU \\
& Output dim & 1 \\ 
\midrule
\textbf{FSQ} & $N_{level}$ & $9$ \\
& $N_{token}$ & $40$ \\
\midrule
\textbf{PPO} & Actor optimizer & Muon \\
& Critic optimizer & AdamW\\
& Clip ratio $\epsilon$ & $0.2$ \\
& GAE $\lambda$ & $0.95$ \\
& Discount factor $\gamma$ & $0.99$ \\
\bottomrule
\end{tabular}
\end{table}

\subsection*{B.4. Prioritized Sampling}
During skill quantization, our objective is to train a single tracking policy capable of imitating hundreds of hours of motion clips. When learning from large motion datasets, a common strategy to improve sample efficiency and success rate is to apply prioritized sampling, which allocates more training updates to difficult or underperforming motions \citep{won2019learning, zhu2023neural}. We adopt a success-rate-based sampling weight update strategy from the tracking controller training setting in ProtoMotions ~\citep{tessler2025protomotions}. 
Tracking a motion is considered successful if its maximum per-joint tracking error over the episode falls below a threshold $\tau$, where we use $\tau=0.5\,\mathrm{m}$. We assign each motion clip a unique index $m$ and associate it with a sampling weight $w_m$ that determines its probability of being selected for rollouts during training. During evaluation, we partition motions into a success set $\mathcal{S}$ and a failure set $\mathcal{F}$ according to this criterion, and update weights by down-weighting successful motions and up-weighting failed motions:

\begin{equation}
w_m \leftarrow 
\begin{cases}
w_m \cdot \gamma^{K}, & m \in \mathcal{S},\\[4pt]
w_m / \gamma^{K}, & m \in \mathcal{F},\\
\end{cases}
\label{eq:success_weight_update}
\end{equation}
where $\gamma \in (0,1)$ controls the adjustment magnitude and $K$ is the number of training epochs between consecutive evaluations, so that the cumulative weight adjustment $\gamma^{K}$ scales with the length of the update interval.

\subsection*{B.5. Comparison vs Diffusion-based Tracking Policy}
Diffusion and flow matching models have demonstrated success in high-dimensional generative modeling \citep{ho2020denoising, lipman2023fm}, prompting recent efforts to leverage their expressiveness for tracking-based control. PDP uses distillation with an MLP expert  policy~\citep{ross2011dagger}, to a DDPM \citep{ho2020denoising}, whereas FPO trains a flow matching model end-to-end from scratch via PPO.

\begin{table}[t!]
\centering
\caption{Comparison of tracking controllers on the AMASS PHC train subset. PDP and FPO are diffusion-based tracking policies. PDP is trained via DAgger distillation from an MLP tracking policy~\citep{ross2011dagger}, whereas all other methods are trained end-to-end from scratch with model-free RL. The MLP baseline remains the strongest overall, achieving the best tracking success rate and the lowest tracking error. Our FSQ-based tracking policy attains a comparable success rate to the continuous trackers but exhibits higher MPJPE.}
\begin{tabular}{lcccc}
\toprule
\textbf{Method} & \textbf{Type} & \textbf{End2end} & \textbf{Succ. (\%)} & \textbf{MPJPE (mm)}\\
\midrule
PDP & Continuous & No & 98.90 & 37.32 \\
FPO & Continuous & Yes & 96.40 & 41.98 \\
GPC & Discrete & Yes & 99.51 & 44.43 \\
MLP & Continuous & Yes & \textbf{99.59}  & \textbf{30.26} \\
\bottomrule
\end{tabular}
\label{tab:comparison_diffusion}
\end{table}

As shown in Table \ref{tab:comparison_diffusion}, we benchmark our FSQ-based tracking controller against these continuous baselines, alongside a standard MLP policy. The MLP baseline remains the strongest overall, attaining both the highest success rate (99.59\%) and the lowest tracking error (30.26 mm), which suggests that the added expressiveness of generative trackers does not directly translate into tighter tracking under a pure imitation objective. Among the generative approaches, our FSQ controller achieves the highest success rate (99.51\%), surpassing PDP and the end-to-end FPO baseline, while incurring a modest increase in MPJPE relative to the continuous trackers. We attribute this gap to the inherent quantization error of discrete action representations, which trades off fine-grained precision for a structured action space. However, our discrete formulation provides a natural interface to the next-token prediction paradigm, enabling seamless integration with downstream autoregressive models in a way that continuous policies cannot easily support.

\subsection*{B.6. Probing FSQ Skill Latent Space}
Since FSQ rounds each latent channel directly to a small set of fixed scalar levels during quantization, we hypothesize that the FSQ encoder can map discrete skill latents with similar skill patterns to nearby regions on the manifold. To test this property, we design a skill retrieval experiment on a subset extracted from Bones dataset, with motion categories defined at two granularities derived from file names: (1) a coarse level with 8 classes (\textit{walk, jog, jump, kick, punch, crawl, idle, dance}), covering 6{,}469 motions; and (2) a fine level with 20 classes that further distinguishes locomotion styles (e.g., \textit{tired\_walk, angry\_walk, zombie\_walk, happy\_jog}), covering 3{,}498 motions. The two-level granularity allows us to assess whether the skill latent captures the details of both broad action categories and subtle stylistic variation. To obtain a global representation for each motion clip, we adopt a mean-pooling strategy following Sentence-BERT \citep{reimers2019sentencebert}. We pass every frame of a motion sequence through the encoder of a pretrained FSQ tracking policy and average the resulting quantized codes across all timesteps, yielding a single mean-pooled FSQ embedding that summarizes the entire clip. We use $L_2$ distance as the retrieval metric, as cosine similarity would discard the magnitude information carried by the discrete FSQ skill latents. We evaluate with two metrics following the evaluation protocol used for text embeddings in the Massive Text Embedding Benchmark (MTEB) \citep{muennighoff-etal-2023-mteb}:
\begin{itemize}
  \item \textbf{Nearest-Neighbor Retrieval.} For each motion, we retrieve the
    top-$K$ nearest neighbors and report Precision@$K$, Mean Reciprocal Rank (MRR), and Mean Average Precision at depth 10 (MAP@10).  
  \item \textbf{Pair Discrimination.} We sample 10{,}000 motion pairs, balanced 50/50 between same-category and cross-category pairs, and evaluate whether the similarity metric can distinguish the two groups. We report Spearman correlation, AUC-ROC, and accuracy at the optimal threshold.
\end{itemize}

Table~\ref{tab:retrieval} summarizes the results. At the coarse level, the skill latents achieve a Precision@1 of 0.72 and MRR of 0.81, meaning the nearest neighbor in FSQ latent space is the correct motion type 72\% of the time, and the first correct match appears near the top of the ranked list on average.  At the fine-grained level (20 classes), retrieval performance naturally decreases (P@1$=$0.66, MRR$=$0.76), as the model must distinguish between closely related styles such as \emph{tired\_walk} and \emph{scared\_walk}. Pair discrimination, however, improves (AUC-ROC$=$0.69, Spearman$=$0.33), suggesting that fine-grained classes are more internally homogeneous than coarse ones.

\begin{table}[t!]
\centering
\caption{FSQ latent retrieval evaluation. Higher is better for all metrics.}
\label{tab:retrieval}
\small
\begin{tabular}{lcc}
\toprule
\textbf{Metric} & \textbf{Coarse (8 cls)} & \textbf{Fine (20 cls)} \\
\midrule
\multicolumn{3}{l}{\textit{Retrieval (leave-one-out)}} \\
\midrule
\quad \textbf{Precision@1} & 0.72 & 0.66 \\
\quad \textbf{Precision@5} & 0.63 & 0.53 \\
\quad \textbf{Precision@10} & 0.58 & 0.46 \\
\quad \textbf{MRR} & 0.81 & 0.76 \\
\quad \textbf{MAP@10} & 0.48 & 0.37 \\
\midrule
\multicolumn{3}{l}{\textit{Pair Discrimination}} \\
\midrule
\quad \textbf{Spearman} & 0.26 & 0.33 \\
\quad \textbf{AUC-ROC} & 0.65 & 0.69 \\
\quad \textbf{Accuracy} & 0.62 & 0.65 \\
\bottomrule
\end{tabular}
\end{table}

\begin{figure}[t]
    \centering
    \begin{minipage}{1.0\columnwidth}
        \centering
        \includegraphics[width=\linewidth]{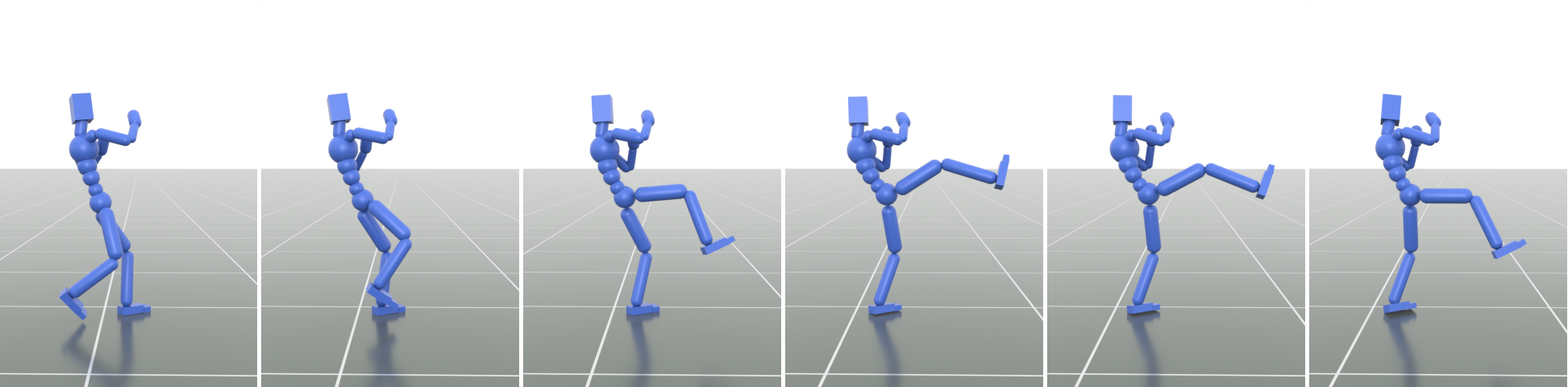}
        \small (a) Kick
    \end{minipage}

    \begin{minipage}{1.0\columnwidth}
        \centering
        \includegraphics[width=\linewidth]{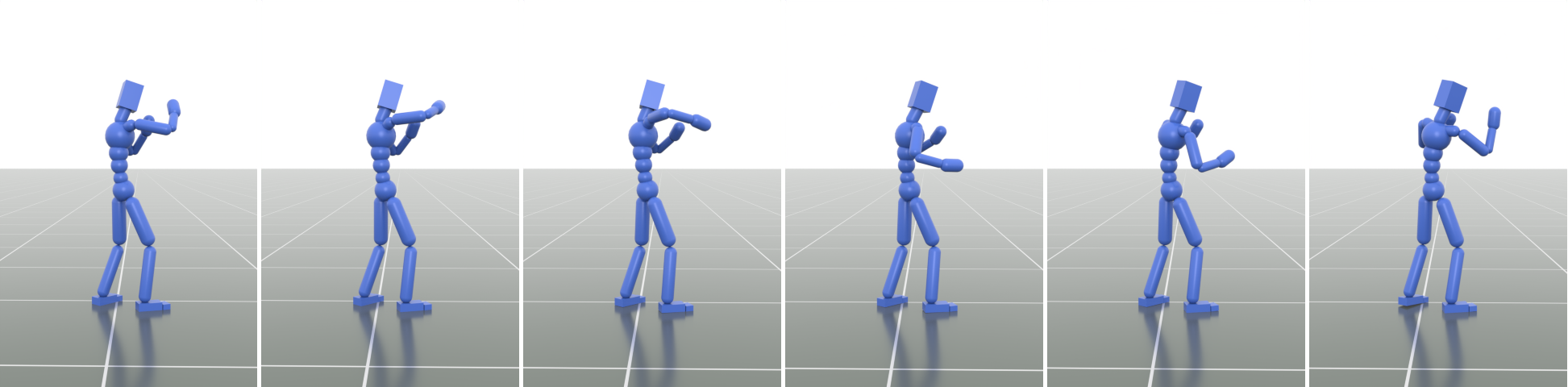}
        \small (b) Punch
    \end{minipage}

    \begin{minipage}{1.0\columnwidth}
        \centering
        \includegraphics[width=\linewidth]{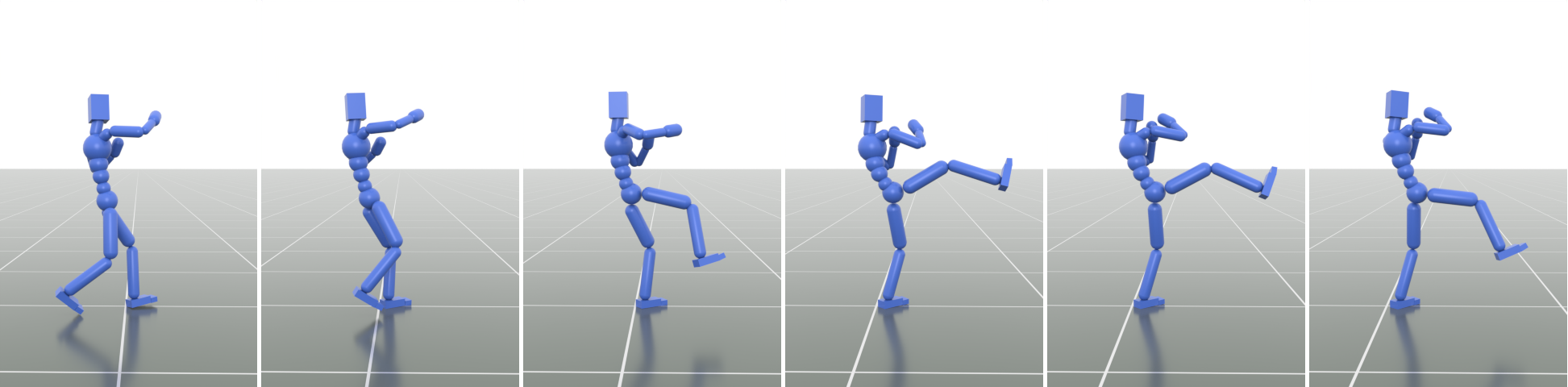}
        \small (c) Punch while kicking  
    \end{minipage}

    \caption{Composited skill achieved by framewise code blending.}
    \label{fig:code_blending}
\end{figure}

\begin{figure*}[t]
    \centering
        \centering
        \includegraphics[width=0.78\linewidth]{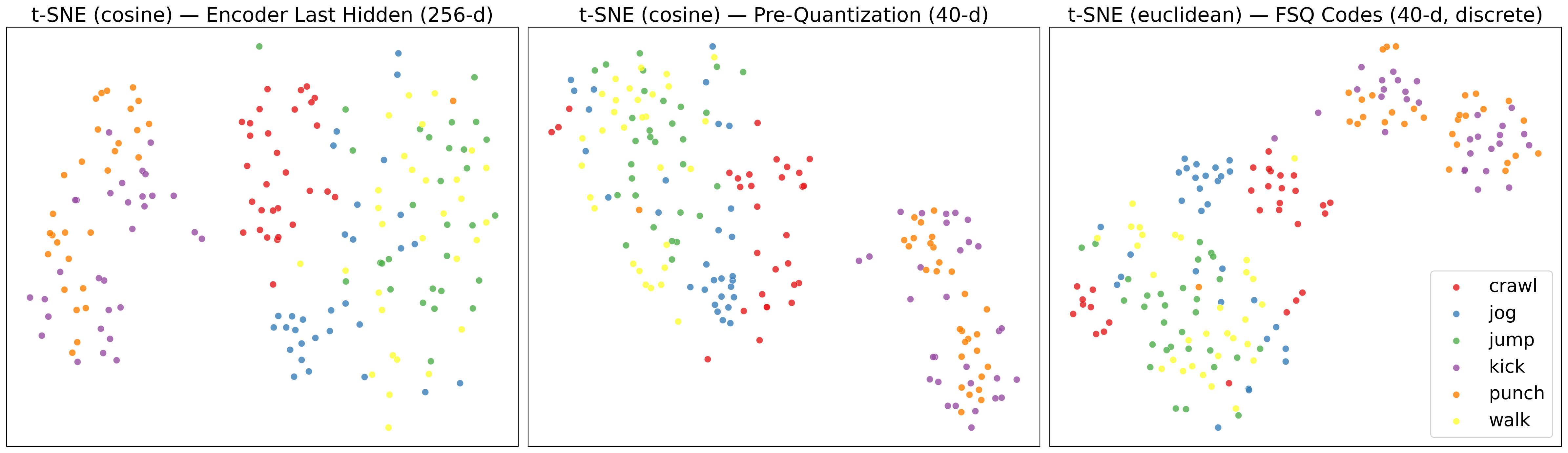}
    
    \caption{t-SNE visualization of FSQ embeddings at various network depths. Note that the pre-quantization FSQ codes and the output embeddings of the last hidden layer are clustered using cosine distance, while the discretized FSQ codes are clustered with a Euclidean distance metric.}
    \label{fig:fsq_emb}
\end{figure*}    

We analyze the learned representation structure by visualizing t-SNE projections of the FSQ encoder's embeddings (Fig.~\ref{fig:fsq_emb}) \citep{TSNE}. Cosine similarity is calculated as a distance metric for the final hidden-layer embeddings and pre-discretization FSQ latents, as shown in Figure~\ref{fig:fsq_emb} (a) and (b). Euclidean distance is utilized for post-discretization FSQ latents, shown in Figure~\ref{fig:fsq_emb} (c). Across all three plots, the representations exhibit clustering effect based on distinct characteristics of motions. The proximity of similar locomotion styles within the latent space suggests that the FSQ bottleneck produces a structured geometry. This shows that the model is learning meaningful relationships between skills, rather than assigning tokens at random. Beyond clustering, FSQ latent codes support skill composition. As shown in Fig.~\ref{fig:code_blending}, interpolating between `punching' and `kicking' produces a motion that performs both strikes.

\section*{C. Generative Controller}
\subsection*{C.1. Architecture and Training Configuration}
 The architecture of our generative controller is built around a causal transformer. It receives the current state as a context token, projected through MLP-based state encoder layers, and autoregressively predicts $N_{token}$ discrete tokens that represent the quantized latent of skills. System architecture and hyperparameters are reported in Table \ref{tab:transformer_arch}. We implement the generative controller using PyTorch's official transformer module \citep{pytorch}, with GELU activations \citep{hendrycks2023gaussianerrorlinearunits}, $N_{heads}$ attention heads, and $N_{layers}$ transformer encoder layers. Causal masking is applied to enable teacher forcing during training. The training objective is a cross-entropy loss with label smoothing \citep{muller2020labelsmoothing}, which discourages overconfident predictions across the $9^5$ possible token combinations. To further stabilize the training process and improve generalization, we maintain an Exponential Moving Average (EMA) of the model weights with a smoothing factor of 0.9. At inference time, we employ autoregressive prediction and use Nucleus Sampling with top-$p=0.9$.

\begin{table*}[t]
\centering
\caption{In a comparison against CVAE across 15 starting poses with 32 rollouts each (20s duration), GPC demonstrates superior performance in both motion quality and diversity in the absence of external perturbations. It achieves lower Jerk and Acceleration values while simultaneously maintaining higher Root APD and Pose APD compared to the baseline. GPC exhibits greater robustness than the baseline, consistently maintaining higher survival rates as external perturbations increase.}
\label{tab:prior_comparison}
\small
\begin{tabular*}{\textwidth}{@{\extracolsep{\fill}}lcccccc@{}}
\toprule
& \multicolumn{2}{c}{\textbf{No Force Perturbation}} & \multicolumn{2}{c}{\textbf{Push Impulse (2.4 m/s)}} & \multicolumn{2}{c}{\textbf{Push Impulse (9.8 m/s)}} \\
\cmidrule(lr){2-3} \cmidrule(lr){4-5} \cmidrule(lr){6-7}
\textbf{Metric} & \textbf{GPC} & \textbf{CVAE} & \textbf{GPC} & \textbf{CVAE} & \textbf{GPC} & \textbf{CVAE} \\
\midrule
\textbf{Norm. Jerk} $\downarrow$       & $\mathbf{657 \pm 147}$  & $1072 \pm 38$          & $\mathbf{1054 \pm 32}$ & $1172 \pm 32$          & $1458 \pm 14$          & $\mathbf{1406 \pm 33}$ \\
\textbf{Norm. Foot Jerk} $\downarrow$  & $1002 \pm 184$          & $\mathbf{989 \pm 89}$  & $1551 \pm 112$         & $\mathbf{1177 \pm 66}$ & $1918 \pm 64$          & $\mathbf{1389 \pm 68}$ \\
\textbf{Mean Accel. (m/s$^2$)} $\downarrow$      & $\mathbf{4.33 \pm 0.46}$ & $6.14 \pm 0.21$        & $\mathbf{5.91 \pm 0.19}$ & $6.23 \pm 0.23$        & $7.72 \pm 0.14$        & $\mathbf{6.51 \pm 0.09}$ \\
\textbf{APD Root (m)} $\uparrow$       & $\mathbf{2.66}$         & $2.40$                 & $3.55$                 & $\mathbf{3.72}$        & $\mathbf{6.89}$        & $6.01$                 \\
\textbf{APD Pose (m)} $\uparrow$       & $\mathbf{13.55}$        & $12.22$                & $17.90$                & $\mathbf{18.72}$       & $\mathbf{34.33}$       & $29.82$                \\
\textbf{Survival Rate (\%) } $\uparrow$ & $\mathbf{82.1\%}$   & $79.4\%$               & $\mathbf{68.1\%}$      & $44.4\%$               & $\mathbf{52.5\%}$      & $3.1\%$                \\
\bottomrule
\end{tabular*}
\end{table*}

\begin{table}[t!]
\centering
\caption{Network architecture and grouping configuration of the generative controller. $d_{\text{model}}$ represents the model dimension, $N_{\text{heads}}$ indicates the number of attention heads, $N_{\text{layers}}$ denotes the total transformer layers, and $d_{\text{ff}}$ refers to the feed-forward network dimension.}
\label{tab:transformer_arch}
\begin{tabular}{llc}
\toprule
\textbf{Component} & \textbf{Parameter} & \textbf{Value} \\ 
\midrule
\textbf{Transformer} & $d_{\text{model}}$ & 1024 \\
& $N_{\text{heads}}$ & 4 \\
& $N_{\text{layers}}$ & 6 \\
& $d_{\text{ff}}$ & 4096 \\
& Activation & GELU \\
\midrule
\textbf{Token Grouping} & $G$ & 5 \\
& $N_{token}$ & 8 \\
& Vocab size ($|\mathcal{V}|$) & $9^5 = 59,049$ \\
\midrule
\textbf{Positional Encoding} & Type & Learned \\
\bottomrule
\end{tabular}
\end{table}

\subsection*{C.2. Connection to Transformer-based RL and Control Sequence Modeling}
Transformer-based offline RL architectures, notably Decision Transformer and Trajectory Transformer~\citep{janner2021trajformer, chen2021decisiontransformer}, circumvent traditional step-wise policy optimization by trajectory modeling. Those approaches treat states, actions, and returns, as a joint sequence prediction task. Decision Transformer operates on continuous embeddings, while Trajectory Transformer discretizes states and actions through per-dimension binning. GPC shares the transformer backbone with those methods but differs from this line of work in three key respects. First, skills are tokenized via FSQ into a discrete vocabulary, rather than through the per-dimension binning of Trajectory Transformer or the continuous embeddings of Decision Transformer. Second, the backbone adopts a GPT-style autoregressive formulation ~\citep{radford2018gpt, radford2019gpt2}, trained purely via next-token prediction over this compact discrete vocabulary. Third, GPC operates in the online reinforcement learning regime and requires a simulator for training, in contrast to the offline setting of Decision Transformer and Trajectory Transformer.

In the online regime, recent studies have utilized transformers as control policies~\citep{pan2025tokenhsi}. TokenHSI unifies multiple skills in a single transformer by tokenizing per-task goals as continuous embeddings, and trains the network as a deterministic policy with an AMP-style adversarial reward~\citep{peng2021amp}. While TokenHSI also supports task adaptation, it transfers from a fixed set of pretrained tasks to new ones. GPC instead trains a generative prior unconditionally on a large-scale motion dataset, and downstream tasks are addressed by selecting skills from this learned latent space.

\subsection*{C.3. GPC vs. Continuous Priors: Unconditional Sampling}
Our study evaluates unconditional generation quality by training GPC and CVAE based on an FSQ tracking controller that is pretrained on the AMASS PHC train subset. We assess their performance across a comprehensive suite of metrics. Normalized Jerk is computed over $0.4\text{s}$ sliding windows of rigid-body positions following the formulation by \citet{Hogan01012009}. We report a whole-body and a foot-only variant for normalized jerk, and the mean acceleration magnitude (m/s$^{2}$) to further evaluate the degree of jitters. Diversity is evaluated using the average pairwise distance (APD), computed as the mean $L_2$ distance across all sample pairs for each starting pose. We compute APD over full episodes on both root xy-trajectories (APD Root) and full-body pose trajectories (APD Pose). Survival rate is a robustness metric representing the proportion of episodes in which the character concludes the simulation with a root height above 0.5 m, demonstrating it is not lying on the ground at the end, even if falls occurred earlier in the duration.

Table~\ref{tab:prior_comparison} compares GPC and CVAE unconditional priors on motion quality, diversity, and robustness. Both models are evaluated from 15 starting poses with 32 parallel rollouts each, running for 20\,s without resets. We test three conditions: no external force, a moderate push impulse (2.4\,m/s), and a strong push impulse (9.8\,m/s), all applied at step~200 with identical, seeded force directions across methods. 

Without perturbation, GPC achieves lower normalized jerk and lower mean acceleration than the CVAE prior, while foot-level jerk remains relatively higher, indicating more foot jitter behavior. GPC exhibits higher trajectory diversity in both root paths and full-body poses. Survival rates are similar, indicating that both priors maintain physically plausible behavior over extended horizons. The robustness gap becomes pronounced under external forces. With a moderate push, GPC retains a 68\% survival rate versus 44\% for CVAE. Under the strong push, the survival rate is still above 50\%, while CVAE collapses to 3.1\%. Characters controlled by GPC frequently recover after being knocked down, whereas those using CVAE rarely do. In the absence of perturbations, CVAE demonstrates better performance with respect to foot jerk. Although GPC also exhibits a sharper increase in foot jerk, this likely reflects the aggressive, corrective foot placements required during recovery. Overall, GPC offers a favorable combination of motion quality, diversity, and robustness compared to the CVAE prior.

\section*{D. Task Adaptation}
\subsection*{D.1. Locomotion Tasks}
In the \textbf{Target Reaching} task, the character is required to navigate toward a specified 2D goal location in the environment. The task observation consists of a 2D target position expressed in the character’s local coordinate frame, enabling the policy to remain invariant to global orientation. The target position is periodically reset to a random location every $\tau_{reset}$ seconds. This setup requires the controller to continuously replan and adapt its motion in response to changing goals, testing both responsiveness and stability under dynamic conditions. We reward proximity to a target location $\mathbf{p}^* \in \mathbb{R}^2$:
\begin{equation}
    r_{\text{target}} = \begin{cases}
        1 & \text{if } \|\mathbf{p} - \mathbf{p}^*\|_2 < \epsilon_{\text{prox}} \\
        \exp\left(-\alpha_{\text{pos}} \|\mathbf{p} - \mathbf{p}^*\|_2\right) & \text{otherwise}
    \end{cases}
\end{equation}
where $\mathbf{p}$ is the root position projected onto the ground plane, $\epsilon_{\text{prox}}$ is the proximity threshold, and $\alpha_{\text{pos}}$ controls the reward sharpness. 

\begin{figure}[h]
    \centering
    \begin{minipage}{0.485\textwidth}
        \centering
        \includegraphics[width=\linewidth]{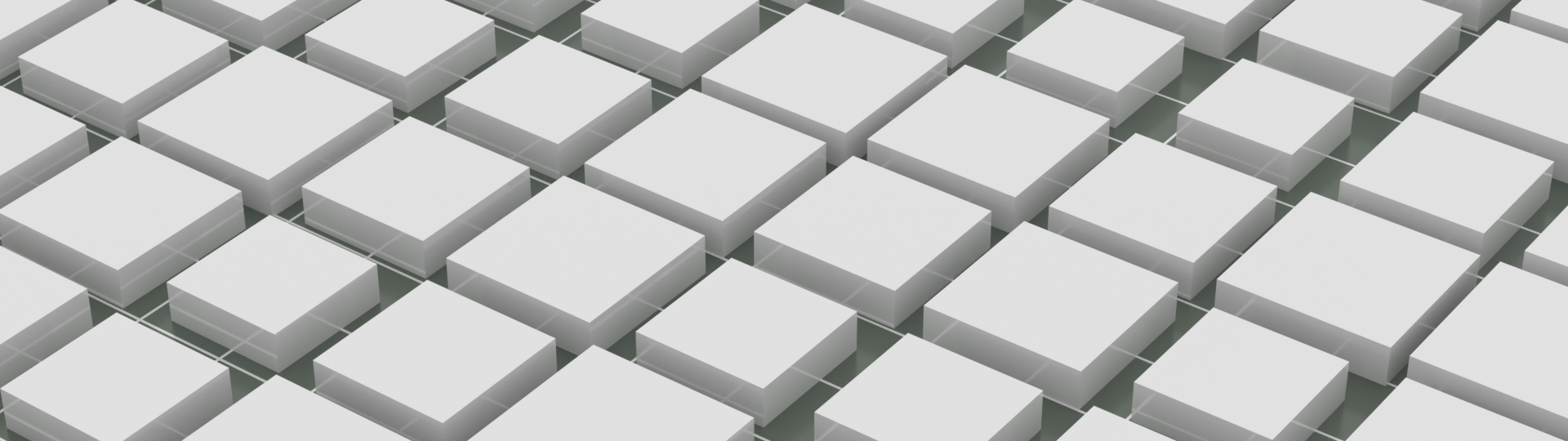}
        \small (a) Platform.
    \end{minipage}\hfill
    \begin{minipage}{0.485\textwidth}
        \centering
        \includegraphics[width=\linewidth]{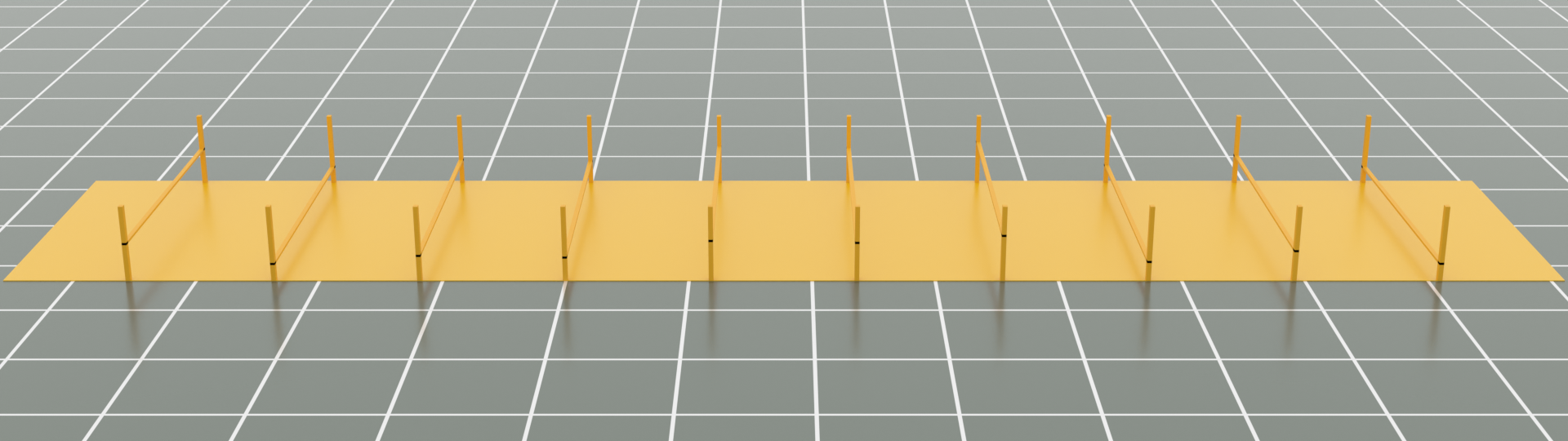}
        \small (b) Barrier.
    \end{minipage}
    
    \caption{Scenes of HSI tasks.}
    
\label{scene}
\end{figure}

In the \textbf{Joystick Steering} task, we evaluate the controller’s ability to simultaneously track a desired velocity and align with a specified facing direction. The observation includes a 2D target velocity vector along with a target heading direction, both defined in the character’s local frame. At each timestep, the controller must generate actions that achieve the desired velocity while also orienting the body toward the commanded direction. This task introduces a coordination challenge between locomotion and orientation control, requiring the policy to balance multiple objectives in a temporally consistent manner.
For velocity-conditioned locomotion, the agent receives a target heading direction $\hat{\mathbf{d}} \in \mathbb{R}^2$, target speed $v^*$, and target facing direction $\hat{\mathbf{f}} \in \mathbb{R}^2$. The reward combines directional velocity matching with heading alignment:
\begin{equation}
    r_{\text{steer}} = w_{\text{dir}} \cdot r_{\text{dir}} + w_{\text{face}} \cdot r_{\text{face}}
\end{equation}
where $w_{\text{dir}} = 0.7$ and $w_{\text{face}} = 0.3$. The directional reward penalizes both speed error and tangential drift:
\begin{equation}
    r_{\text{dir}} = \begin{cases}
        0 & \text{if } v_{\parallel} \leq 0 \\
        \exp\left(-\alpha_v \left[(v^* - v_{\parallel})^2 + \beta \|\mathbf{v}_{\perp}\|_2^2\right]\right) & \text{otherwise}
    \end{cases}
\end{equation}
where $v_{\parallel} = \dot{\mathbf{p}} \cdot \hat{\mathbf{d}}$ is the velocity component along the target direction, $\mathbf{v}_{\perp} = \dot{\mathbf{p}} - v_{\parallel}\hat{\mathbf{d}}$ is the tangential velocity. The facing reward encourages alignment between the robot's heading $\hat{\mathbf{h}}$ and the target facing direction:
\begin{equation}
    r_{\text{face}} = \max\left(0,\, \hat{\mathbf{h}} \cdot \hat{\mathbf{f}}\right)
\end{equation}

In the \textbf{Trajectory Following} task, the controller must follow a predefined 2D path specified by a sequence of future waypoints. The observation provides the next 10 waypoints that span a 5-second horizon, giving the controller a short-term preview of the desired trajectory and requiring it to plan ahead. Success in this task demonstrates the model's ability to generate coherent motion by selecting motor skills, represented as discrete tokens, that align with path constraints. The adapted controller follows the target trajectory accurately while maintaining naturalistic behavior. Given a target position $\mathbf{p}^* \in \mathbb{R}^2$, the reward is as follows,
\begin{equation}
    r_{\text{path}} = 
        \exp\!\left(-\alpha_{\text{traj}} \|\mathbf{p} - \mathbf{p}^*\|_2\right) 
\end{equation}
$\mathbf{p}$ is the root position in the $xy$ plane, $\alpha_{traj}$ is the scale of the proximity reward term. Reward hyperparameters are summarized in Table~\ref{tab:task_rewards}.

\begin{table}[t]
\centering
\caption{Task reward hyperparameters.}
\label{tab:task_rewards}
\begin{tabular}{llcc}
\toprule
\textbf{Task} & \textbf{Parameter} & \textbf{Symbol} & \textbf{Value} \\
\midrule
\multirow{3}{*}{Target}
    & Position scale       & $\alpha_{\text{pos}}$   & 0.42   \\
    & Proximity threshold  & $\epsilon_{\text{prox}}$ & 0.5\,m \\
    & Reset timer          & $\tau_{\text{reset}}$   & 6--8s  \\ 
\midrule
\multirow{4}{*}{Joystick}
    & Velocity scale   & $\alpha_v$        & 0.25 \\
    & Tangent weight   & $\beta$           & 0.1  \\
    & Direction weight & $w_{\text{dir}}$  & 0.7  \\
    & Facing weight    & $w_{\text{face}}$ & 0.3  \\
\midrule
Trajectory & Position Scale & $\alpha_{\text{traj}}$ & 2.0  \\
\bottomrule
\end{tabular}
\end{table}

\subsection*{D.2. Human-Scene Interaction Tasks}
We further introduce two Human-Scene Interaction (HSI) tasks to evaluate the controller in environments that require rich interaction with surrounding geometry. We train our generative controller with a mixed dataset of LAFAN1, Beyond, and Bones. These tasks, including Barrier and Platform, are formulated as target-reaching problems but involve structured scene elements that constrain feasible motion. To represent terrain observations, we use square heightmaps, from which terrain features are extracted via a lightweight 2D CNN encoder. Unlike standard locomotion settings, these tasks require the controller to avoid collisions and adapt its movement dynamically to the surrounding environment. Since the pretrained generative policy already encodes the dataset's skill distribution, a minimal proximity-based reward suffices across all human–scene interaction tasks. The reward is as follows,
\begin{equation}
r_{\mathrm{target}} = \exp\left(-\alpha_{\mathrm{pos}} ||\mathbf{p} - \mathbf{p}^*||_2\right)
\end{equation}
This reduces reward design across the entire task suite to a basic target-reaching objective, eliminating the need for complex task-specific reward shaping.

In the \textbf{Platform} task, the character must traverse a terrain of elevated platforms to reach the target location. We consider two terrain variants: a regular $2\times2$ grid of evenly spaced platforms (\emph{platform-2x2}), and a larger, randomly generated layout (\emph{platform-giant}), shown in Figure~\ref{scene} (a). Platform sizes and inter-platform gaps in \emph{platform-giant} are randomized, requiring the controller to adjust step length, timing, and balance to land safely on discontinuous surfaces. For the platform-giant terrain, gap widths are sampled uniformly between 0.7m and 1.7m. To introduce vertical complexity, neighboring platforms may differ in height by up to 0.3m. To ensure task feasibility and focus on local navigation, the target is spawned on a platform in the immediate vicinity of the character. In the \textbf{Barrier} task, the character encounters obstacles of varying heights along the path to the target. Barrier heights are randomized, and a gap is left beneath each obstacle, so the controller must invoke a distinct set of skills, ducking or crawling under each barrier, to pass through. This task evaluates the model's ability to adapt its motion to strict vertical constraints. Overall, these HSI tasks introduce scene constraints. Despite this increased complexity, our adapted controller retains the diverse and robust skills learned during generative pretraining while effectively handling scene constraints.

\begin{table}[t!]
\centering
\caption{Task Adaptation Configuration}
\label{tab:peft_config}
\begin{tabular}{llc}
\toprule
\textbf{Category} & \textbf{Parameter} & \textbf{Value} \\ 
\midrule
\textbf{Adapter} 
& Rank ($r$) & 64 \\
& Scaling ($\alpha$) & 128 \\
\midrule
\textbf{Rollout} & Top-$p$ & 0.9\\
& Temperature (T) & 1.0 \\
\bottomrule
\end{tabular}
\end{table}

\subsection*{D.3. Task Adaptation Implementation}
CoLA applies FiLM adaptation to DoRA low-rank matrices, making the adapter conditioning-dependent. The low-rank matrices $\mathbf{A}$ and $\mathbf{B}$ have rank $r$ and scaling factor $\alpha$; $\mathbf{B}$ is zero-initialized so that the adapted model coincides with the frozen generative controller at the start of training. At rollout, we restrict the action support to the top-$p$ mass of the frozen controller's output distribution, renormalize the adapted controller's distribution on this support, and draw a sample. Hyperparameters are reported in Table~\ref{tab:peft_config}.

\begin{figure}[h]
    \centering
    \begin{minipage}{0.495\linewidth}
        \centering
        \includegraphics[width=\linewidth]{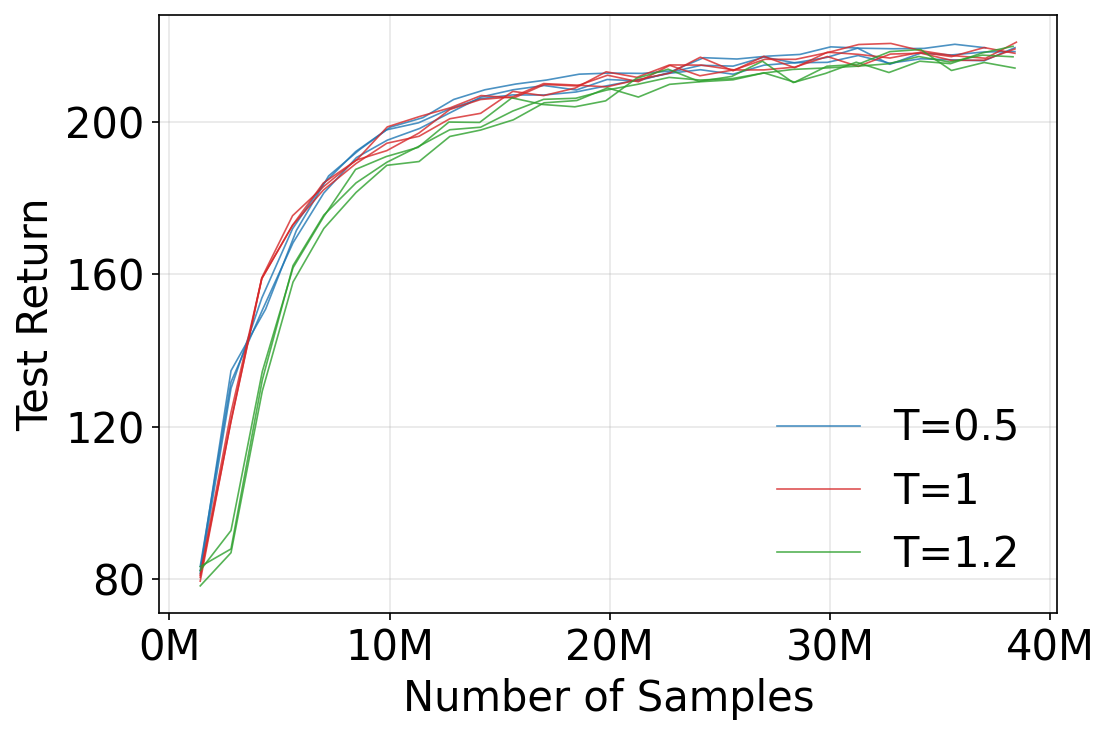}
        \small (a) Temperature sweep
    \end{minipage}
    \hfill 
    \begin{minipage}{0.495\linewidth}
        \centering
        \includegraphics[width=\linewidth]{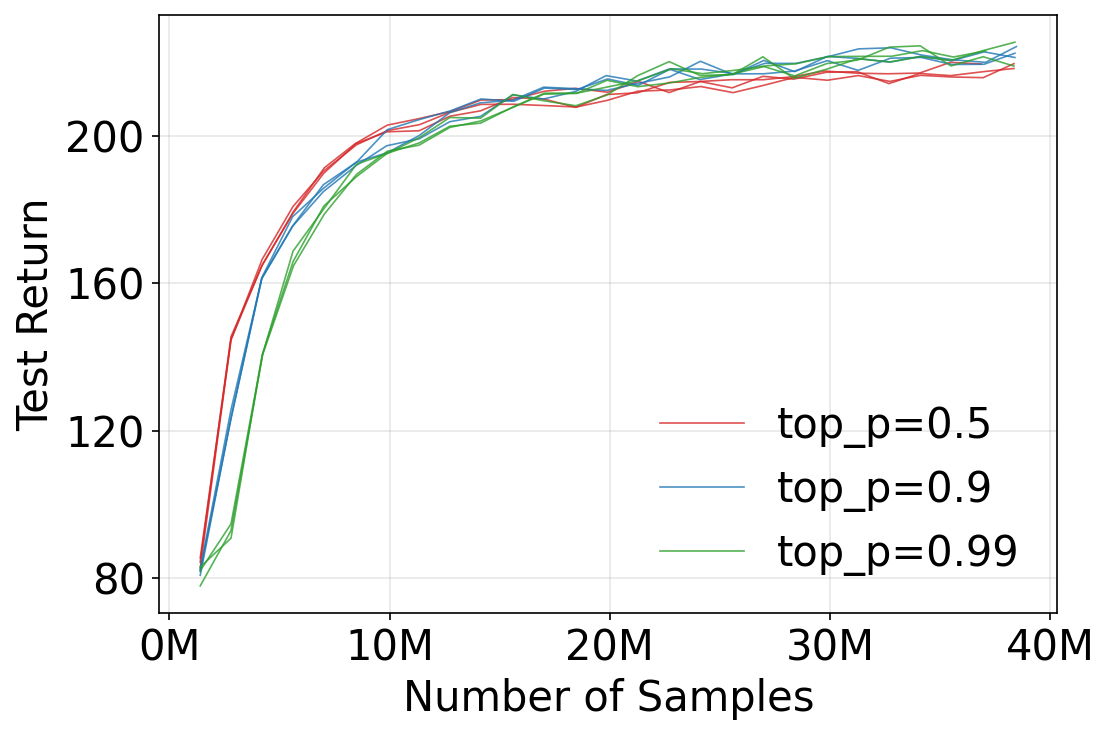}
        \small (b) Nucleus threshold sweep
    \end{minipage}

    \caption{Return Curve under different Nucleus Sampling configurations during task adaptation. (a) Varying the sampling temperature $T$ with a fixed nucleus threshold $p = 0.9$. (b) Varying the nucleus threshold $p$ with a fixed temperature $T = 1.0$.}
    \label{ex:peft_exploration_configuration}
\end{figure}

\subsection*{D.4. Ablation of Nucleus Sampling Configuration}
We conduct an ablation study to evaluate how Nucleus Sampling configurations during task adaptation influence downstream performance. Using a target-reaching task, we compared a pair of default settings ($p = 0.9, T = 1.0$) against variations in temperature ($T \in \{0.5, 1.2\}$) and nucleus thresholds ($p \in \{0.5, 0.99\}$). As illustrated in Figure~\ref{ex:peft_exploration_configuration} (a), variations in temperature at a fixed $p = 0.9$ yielded broadly similar quantitative results. However, higher temperatures qualitatively degraded control by flattening the output distribution. This increased the probability of sampling low-probability codes, resulting in persistent jitter, with $T = 1.2$ performing worse than the default setting in qualitative results. Regarding threshold variations in Figure~\ref{ex:peft_exploration_configuration} (b), settings of $p \geq 0.9$ achieved higher rewards than lower values. While low $p$ configurations converged faster due to a restricted action space, they ultimately limited the agent's repertoire, suggesting that a narrower selection of codes constrains the essential skill sets required for optimal adaptive performance.

\begin{table}[t!]
\centering
\caption{Comparison of GPC, CVAE, and MaskedMimic on success rate and final target distance, evaluated over 256 episodes.}
\label{tab:method_comparison}
\begin{tabular}{lcc}
\toprule
Method & Success Rate (\%) $\uparrow$ & Final Target Dist. (m) $\downarrow$ \\
\midrule
MaskedMimic & 89.92 & 0.44 \\
CVAE        & 93.89 & 0.41 \\
GPC (Ours)  & \textbf{94.20} & \textbf{0.37} \\
\bottomrule
\end{tabular}
\end{table}

\subsection*{D.5. Comparison vs Continuous Priors: Downstream Tasks}
We evaluate GPC against continuous-prior baselines on a goal-reaching downstream task, comparing to the CVAE-based priors of PULSE~\citep{luo2023perpetual} and MaskedMimic~\citep{tessler2024maskedmimic}. GPC and PULSE share the same downstream training schedule, where a task policy is trained with RL on top of a frozen prior. MaskedMimic is trained end-to-end with supervised learning conditioned on randomly masked future joint positions, and at inference is provided with target root positions at a fixed height. As shown in Table~\ref{tab:method_comparison}, GPC and CVAE achieve comparable success rates (0.942 and 0.939, respectively), while MaskedMimic lags behind at 0.899. GPC further achieves the lowest final target distance (0.37m) among the three methods. The performance gap of MaskedMimic stems primarily from a mismatch between the goal-reaching task setup, which conditions on a fixed root height, and its training distribution, in which no real motion exhibits a constant root height. This distributional shift induces unnatural behaviors that hinder task performance. 

\end{document}